%% file: paper.tex
\documentclass[]{memtensor}
\usepackage{enumitem}
\usepackage[utf8]{inputenc}
\usepackage{url}
\usepackage{graphicx}
\usepackage{booktabs}
\usepackage{amsfonts}
\usepackage{nicefrac}
\usepackage{makecell}
\usepackage{microtype}
\usepackage{amsmath}
\usepackage{etoolbox}
\usepackage{lipsum}
\usepackage{minitoc}
\usepackage{tablefootnote}
\usepackage{threeparttable}
\usepackage{wrapfig}
\usepackage{appendix}
\usepackage{multirow}
\usepackage{ulem}
\useunder{\uline}{\ul}{}
\usepackage{colortbl}
\usepackage{adjustbox}
\usepackage{array}
\usepackage{tabularx}
\usepackage{siunitx}
\usepackage{placeins}
\usepackage{float}
\usepackage{xcolor}
\usepackage{mdframed}
\usepackage{listings}
\usepackage{bigstrut}
\usepackage{amsthm}

\usepackage{newtxtext,newtxmath}
\usepackage[most]{tcolorbox}
\usepackage{microtype}
\usepackage{setspace}

\usepackage[table]{xcolor}

\definecolor{InsightBlue}{HTML}{2F5D7C}
\definecolor{InsightBg}{HTML}{F3F6F8}
\definecolor{InsightText}{HTML}{24313A}

\newtcolorbox{keyinsight}[1][Key insight]{
  enhanced,
  breakable,
  colback=InsightBg,
  colframe=InsightBlue,
  coltext=InsightText,
  boxrule=0pt,
  leftrule=2.8pt,
  arc=1.2pt,
  outer arc=1.2pt,
  left=13pt,
  right=13pt,
  top=9pt,
  bottom=10pt,
  before skip=11pt,
  after skip=11pt,
  title={\footnotesize\bfseries\color{InsightBlue}\MakeUppercase{#1}},
  attach title to upper={\par\vspace{4pt}},
  fontupper=\normalsize,
}

\title{
  \titlefont
  \parbox{\textwidth}{
    \centering
    \raisebox{-0.3em}{\includegraphics[height=1.3em]{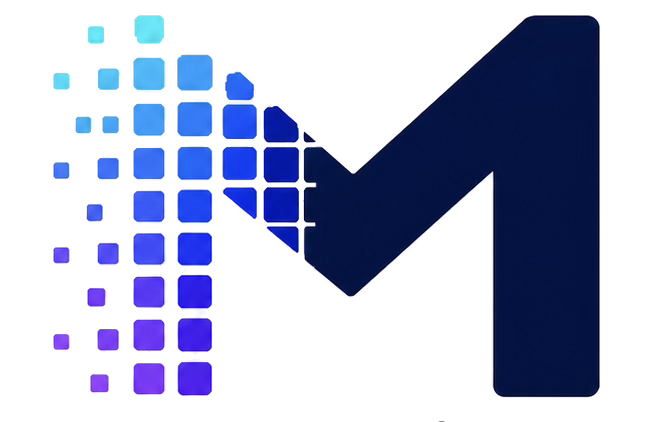}}\hspace{0.3em}
    Metis: Memory Foundation Model
  }
}

\author[]{Zeyu Zhang$^{1,2*}$, Ziliang Guo$^{1*}$, Yihang Sun$^{1,4*}$, Xichong Zhang$^{1*}$, Xixuan Hao$^{1}$, Zehao Lin$^{1}$, Yang Zhang$^{3}$, Xiaoyan Zhao$^{3}$, Tong Shen$^{1,4}$, Bo Tang$^{1}$, Zhi-Qin John Xu$^{4}$, Junchi Yan$^{4}$, Haofen Wang$^{5}$, Xu Chen$^{2\dagger}$, Feiyu Xiong$^{1}$, Zhiyu Li$^{1\dagger}$, Tat-Seng Chua$^{3}$}

\affiliation[1]{MemTensor (Shanghai) Technology Co., Ltd.} 
\affiliation[2]{Renmin University of China}
\affiliation[3]{National University of Singapore}
\affiliation[4]{Shanghai Jiao Tong University}
\affiliation[5]{Tongji University}

\abstract{
Recent advances in AI agents have increasingly internalized native capabilities into their underlying foundation models, giving rise to multimodal foundation models and large reasoning models.
However, agent memory is still primarily implemented through external modules, leaving the native memory capability largely unexplored.
In this paper, we take a first step toward this direction by introducing \textbf{memory foundation models}, which empower foundation models with native memory capabilities.
We formalize \textbf{native memory} from two perspectives: a persistent and dynamically evolving memory state within the backbone, and native memory procedures that autonomously store and utilize information through model computation.
We show that native memory offers advantages in architecture, end-to-end optimization, and efficiency.
Based on this formulation, we propose \textbf{Metis}, the first prototype of memory foundation models.
Metis introduces a new architecture that equips a foundation model with a native memory state, allowing historical information to be compressed into the model and accessed through memory attention.
We construct large-scale memory-specific training data and introduce multiple optimization objectives to acquire these native memory procedures through mid-training.
The online memory maintenance of Metis is gradient-free, and the memory update requires only a forward pass.
At inference time, all learned model weights remain frozen, while the native memory states are autonomously transformed through standard forward computation.
Through extensive experiments, we show that Metis exhibits native memory capabilities and further provide a detailed analysis of its strengths, limitations, and behaviors.
To facilitate future research on memory foundation models, we release our project and model checkpoints.
}

\date{\today}
\correspondence{\email{lizy@memtensor.cn}, \email{xu.chen@ruc.edu.cn}}
\checkdata[Author Legend]{$^*$Co-first authors, $^\dagger$Corresponding authors}

\checkdata[
  \raisebox{-0.7ex}{%
    \includegraphics[height=1.05em]{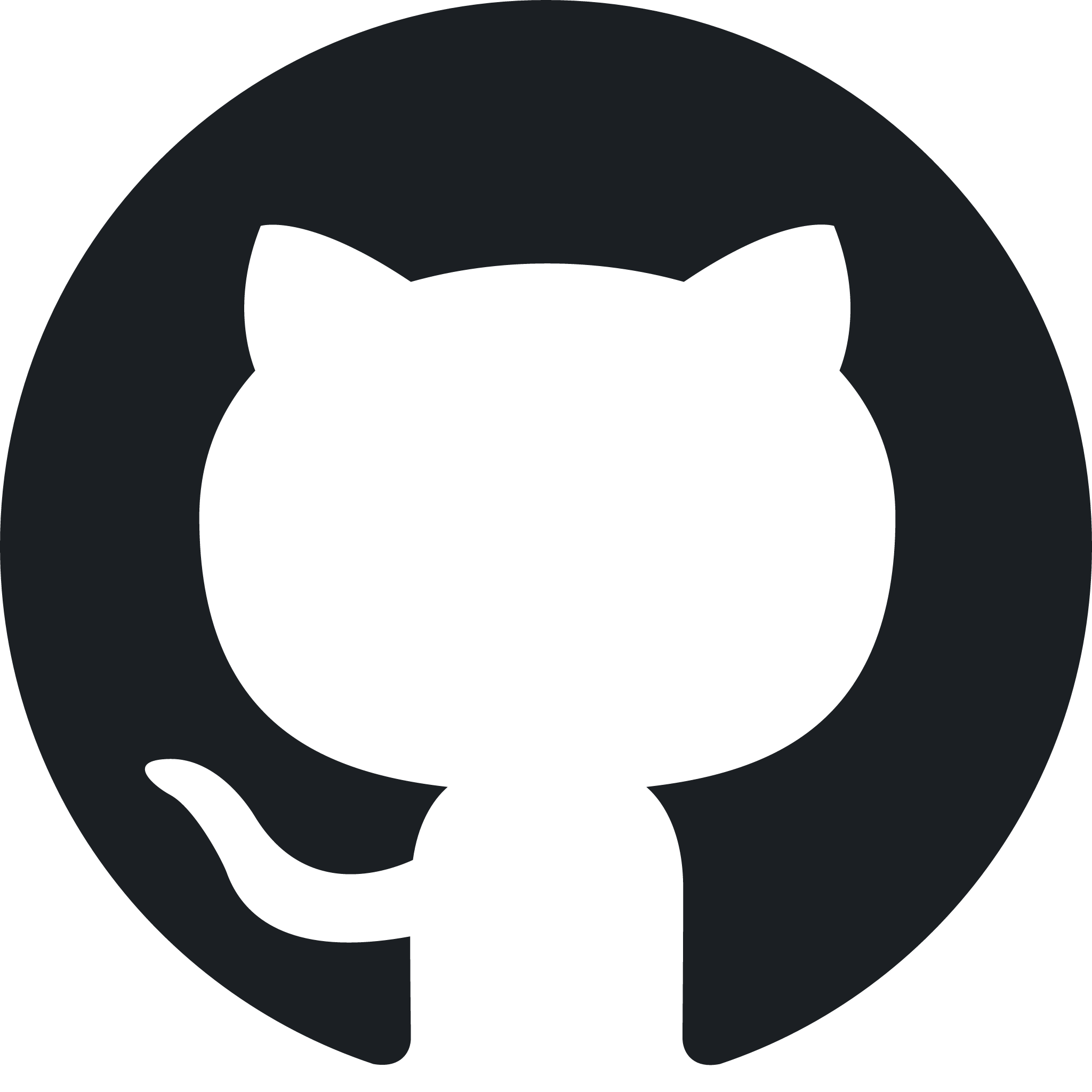}%
  }%
  \hspace{0.3em}Code]{\url{https://github.com/MemTensor/Metis}
}

\checkdata[
  \raisebox{-0.7ex}{%
    \includegraphics[height=1.05em]{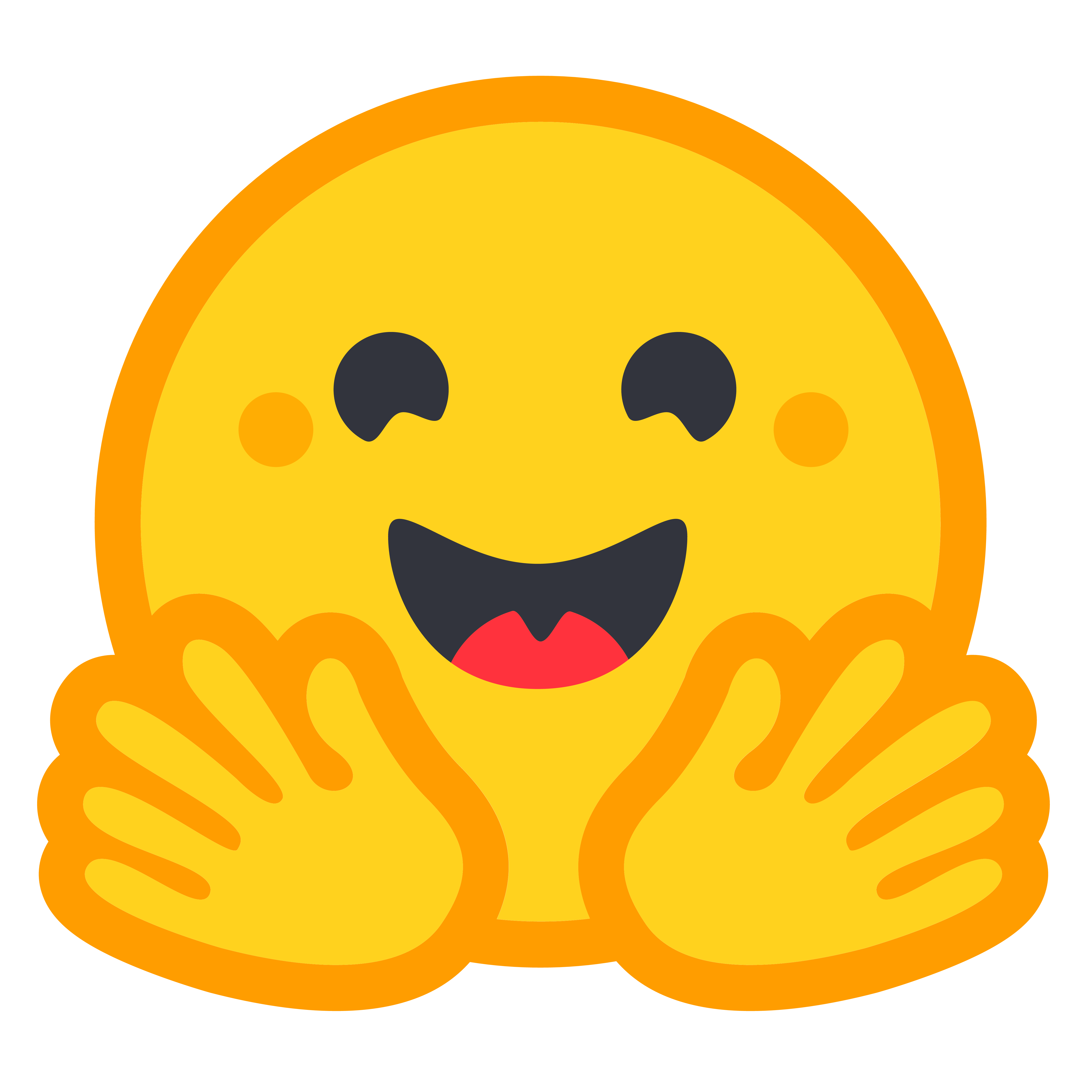}%
  }%
  \hspace{0.3em}Model]{\url{https://huggingface.co/collections/IAAR-Shanghai/metis}%
}

\theoremstyle{definition}
\newtheorem{definition}{Definition}

\begin{document}
\maketitle

\section{Introduction}

In recent years, large foundation models have achieved rapid development, demonstrating significant performance across many aspects, such as language modeling~\cite{zhao2023survey,minaee2024large}, code generation~\cite{jiang2026survey,chen2021evaluating,roziere2023code}, and complex reasoning~\cite{xu2025toward,wei2022chain,guo2025deepseek}.
This provides a solid foundation for constructing AI agents, which enables them to handle more complex tasks.
Beyond the reasoning capabilities of foundation models, memory is another critical capability of AI agents, responsible for retaining past information and leveraging it to support future inference~\cite{zhang2025survey,hu2025memory}.
In most previous works, memory is implemented by a module external to foundation models, rather than being natively integrated into their architectures~\cite{zhong2024memorybank,xu2026mem,packer2024memgptllmsoperatingsystems,li2025memos}.
Representative approaches use Retrieval-Augmented Generation (RAG) to retrieve relevant textual information and incorporate it into the prompt to facilitate inference~\cite{zhong2024memorybank}.

\begin{figure}[!t]
	\centering
    \includegraphics[width=\textwidth]{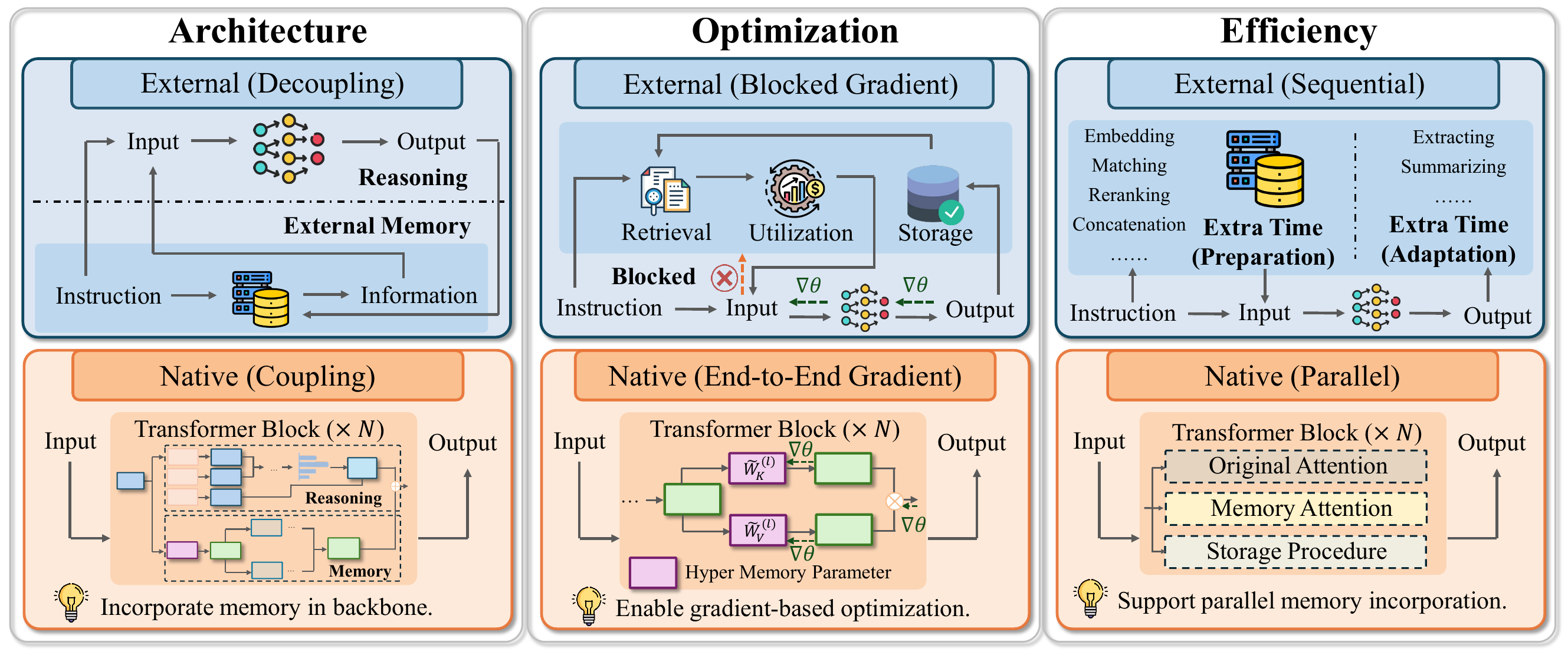}
	\vspace{-0.4cm}
	\caption{From external memory to native memory.}
	\label{fig:intro}
	\vspace{-0.4cm}
\end{figure}

However, external memory suffers from several limitations presented in \textbf{Figure~\ref{fig:intro}}.
First, external memory is decoupled from backbones with separated targets and processing stages~\cite{zhong2024memorybank,packer2024memgptllmsoperatingsystems}.
External memory typically aims to construct an informative context as input, and backbones only perform conditional language modeling over the constructed context.
Therefore, external memory may not provide the most useful information to support the backbone inference, and the backbone may not utilize the memory optimally.
Second, end-to-end optimization is difficult for external memory because gradients cannot be effectively propagated through discrete memory operations~\cite{zhang2025learn}.
As a result, performing domain-specific post-training becomes highly challenging.
Although some RL-based strategies~\cite{zhang2026memrl,guo2026memfactory} can partially alleviate this issue by optimizing memory operations with reward signals, they suffer from efficiency issues.
Finally, external memory requires additional explicit operations over the storage outside backbones, which inevitably increases the online inference latency~\cite{zhang2025survey}.

To address the limitations of external memory, we introduce \textbf{memory foundation models} that empower large foundation models with \textbf{native memory}.
It converts memory from an external module into an internal mechanism of backbones, directly involved in forward computation.
Specifically, memory foundation models can generate responses based on the input instructions and their native memory, with autonomous memory transformation.
We define the native memory from two critical aspects:\\
$\bullet$ \textbf{Native Memory State.}
Unlike traditional large foundation models, memory foundation models are natively stateful across multiple inferences, which can formulate, maintain, and utilize memory states inside backbones from prior inferences.
Their memory states are dynamically represented as part of the parameters of backbones, whose semantic spaces are aligned at the pre-training or mid-training stage.\\
$\bullet$ \textbf{Native Memory Procedure.}
Unlike memory engineering, memory foundation models natively integrate memory procedures within their inferences.
Specific memory operations, such as remembering, forgetting, and updating, are accomplished autonomously alongside the backbone's forward computation, which impacts the native memory state based on input instructions.

Memory foundation models aim to internalize memory capability into the model's forward computation.
The memory state can be represented as dynamic parameters of backbones, and memory procedures are executed through computation.
Therefore, like general foundation models, memory foundation models can be optimized and adapted to specific domains in a data-driven manner through post-training.
This transformation is similar to the evolution from large foundation models to large reasoning models~\cite{guo2025deepseek}, where Chain-of-Thought (CoT)~\cite{wei2022chain} is natively integrated into inference to improve performance and efficiency.
In addition, because native memory procedures can be integrated into the model's computation, they provide a foundation for improving the parallel efficiency of memory processing.

In this paper, we implement the first prototype of memory foundation models, named \textbf{Metis}.
We design a new model architecture that has a native memory state inspired by Fast Weight Programming (FWP)~\cite{ba2016using}, which can be integrated into the backbone computation through memory attention.
Specifically, we propose the Metis blocks as the basic units for native memory. Each of them primarily consists of a hyper memory block and a local memory block.
In addition, we empower Metis with native memory procedures by designing specific optimization objectives, including memory reconstruction and memory operation objectives.
These two objectives correspond to the compression upper limit of memory states and operation targets.
We also design a regularization objective to improve robustness in complex scenarios.
To support this training, we synthesize large-scale memory-specific data from publicly available datasets, enabling Metis to acquire native memory capabilities through mid-training.
Finally, we conduct extensive experiments to demonstrate the effectiveness of our proposed framework, and explore more aspects for analysis. 

From a general perspective, a fundamental problem of memory results from the time-streaming property of online information.
At the storage stage, memory systems cannot determine how the received information will be used in the future.
At the inference stage, the original information is no longer accessible, and only the stored information can be utilized.
Therefore, memory can be considered as a prediction problem, where the model predicts how received information will be utilized in the future.
Like other prediction tasks in machine learning, memory capability can also be acquired at the pre-training stage and generalized to other domains, and memory foundation models can provide the architectural and optimization foundation.

Despite their promising properties, implementing memory foundation models remains highly challenging because their final goal is to completely eliminate the reliance on external memory in contexts.
While Metis achieves great performance in memory-related tasks, it still faces several limitations.
First, its performance degrades on long-term tasks, due to the information loss when compressed into fixed-size parameters.
Second, it exhibits information confusion in some cases, possibly caused by the blending of semantics within the latent space.
Despite these limitations, Metis provides a potential pathway to achieve memory foundation models.
To benefit both the research community and industry, we release our project at \url{https://github.com/MemTensor/Metis}.

Our contributions are summarized as follows: \\
$\bullet$ We introduce memory foundation models and native memory with formal definitions, providing further analysis from the perspective of native memory state and native memory procedure. \\
$\bullet$ We propose the first prototype of memory foundation models, named Metis, which is implemented with novel memory architectures and optimization tasks. \\
$\bullet$ We conduct extensive experiments to verify the effectiveness of our model, followed by detailed studies from multiple perspectives.
We also publicly release our project to benefit the research community and industry.

The rest of our paper is organized as follows. \textbf{Section~\ref{sec:memory_foundation_model}} provides the formal definition of memory foundation models.
\textbf{Section~\ref{sec:metis_architecture}} details the model architecture of Metis.
After that, we introduce our data construction pipeline in \textbf{Section~\ref{sec:data_construction}} and outline the optimization in \textbf{Section~\ref{sec:model_optimization}}.
Extensive experimental results and analysis are presented in \textbf{Section~\ref{sec:experiments}}.
Finally, we review related work in \textbf{Section~\ref{sec:related_work}} and conclude in \textbf{Section~\ref{sec:conclusion}}.

\section{Memory Foundation Model}
\label{sec:memory_foundation_model}
In this section, we provide a formal definition of the memory foundation model.
Then, we introduce native memory from the perspectives of the native memory state and procedure.
After that, we compare memory foundation models with previous works.
Finally, we further discuss memory foundation models from the perspectives of lifelong learning and the evolving trends of AI agents.

\subsection{Definition}
We define the memory foundation model under the multi-step scenario.
Let a continuous interaction process be formulated as a sequence of discrete time steps $t \in \{1, 2, \dots, T\}$.
At each step $t$, the foundation model receives an input instruction sequence denoted as $X_t$ and generates a response sequence $Y_t$.

For traditional foundation models without memory, the generation relies entirely on the current input context.
The autoregressive decoding of the $k$-th token in the response is typically expressed as $y_{t, k} \sim P(y \mid X_t, Y_{t, < k}; \theta)$, where $Y_{t, < k}$ denotes the previously generated tokens at step $t$, and $\theta$ represents the fixed parameters of the backbone.
For foundation models with external memory, the autoregressive decoding process is then conditioned on the context $C_t$ alongside the input instruction.
It can be expressed by
$y_{t, k} \sim P(y \mid C_t, X_t, Y_{t, < k}; \theta)$, where $C_t$ can be obtained from prior information $\{(X_i, Y_i)\}_{i=1}^{t-1}$.
The external memory framework commonly has two explicit procedures, including the storage procedure $\mathbf{C}_t = \mathbf{C}_{t-1} \oplus \{\left(X_t, Y_t \right)\}$, and the retrieval procedure $C_t = \mathbf{C}_{t-1} \otimes X_t$.
Here, $\oplus$ denotes the general writing operation, and $\otimes$ represents the general reading operation with the textual memory storage $\mathbf{C}_{t-1}$.
Both of them are executed outside the model inference process.

\begin{definition}[Memory Foundation Model]
The memory foundation model is defined as an autoregressive foundation model empowered by \textbf{native memory} across multi-step interactions.
At each step $t$, the generation of the $k$-th token is conditioned on the input instruction $X_t$ and the previously generated tokens $Y_{t, < k}$ by
\begin{equation*}
    y_{t, k} \sim P(y \mid X_t, Y_{t, < k}; \theta_t),
\end{equation*}
where the model parameter $\theta_t$ integrates information from previous steps into its native parametric space (\textit{i.e.,} native memory state).
Concurrently, $\theta_{t+1}$ is autonomously transformed during the forward computation inside the model based on the input instruction $X_t$ and output $Y_t$ (\textit{i.e.,} native memory procedure).
\end{definition}

In contrast, the memory foundation model internalizes memory into the backbone's computation, which is empowered with native memory.
Instead of relying on an external explicit storage $\mathbf{C}_t$ and context $C_t$, it maintains a native memory state, which acts as dynamic parameters across multiple steps.
In addition, rather than explicitly executing memory procedures outside backbones, memory procedures in the memory foundation model occur autonomously alongside the model's forward computation, such as operations like remembering, forgetting, and updating.

\subsection{Native Memory State}
In this paper, we adopt a strict definition for the source of memory.
We only consider the information acquired during online interactions as memory, where information available before the interaction starts is excluded.
In fact, such offline information is better viewed as knowledge rather than memory, because it does not contain trajectory-specific information for personalization and does not require real-time adaptation.

Since the stored information varies across different steps, the parameters $\theta_t$ cannot remain completely static. 
Consequently, at least a portion of the parameters must change dynamically according to the input, and we denote this dynamic part as the memory state $\mathbf{M}_t$. 
In the memory foundation model, the representation of stored information is supposed to be coupled with the backbone to participate in forward computation.
Therefore, the native memory state should be represented in parametric form.
Although textual memory offers advantages in interpretability and cross-model compatibility, its discrete representation results in low information density and requires repetitive prefilling.
In contrast, parametric memory represents prior information in a dense form, which increases the efficiency of storage and utilization.

In addition, the semantic spaces of both dynamic parameters $\mathbf{M}_t$ and static parameters $\Phi = \theta_t \setminus \mathbf{M}_t$ must be aligned during the pre-training or mid-training stage before conducting online inference. 
This alignment enables the dynamic parameters at different steps to compute collaboratively with the static parameters.
During online interactions, the native memory state can be updated and utilized through the native memory procedure, which empowers the memory foundation model with statefulness across different steps.

\subsection{Native Memory Procedure}
\label{subsec:mfm_native_procedure}
In terms of memory, storage and utilization are two core procedures to handle online information with the time-streaming property.
Memory storage retains past information, while memory utilization leverages this stored information to support model inference.
They aim to address the temporal mismatch between information supply and usage.

The memory storage procedure typically involves several specific operations, such as remembering, forgetting, and updating.
From the perspective of foundation models, an input instruction contains both the intent and the content of information processing.
For instance, \textit{``Alice is 24 years old''} implies remembering her age, while \textit{``Bob moved from London to Boston''} indicates updating his location.
A native storage procedure should directly map the input instruction to the update value of the memory state.
In contrast, external memory relies on rule-based and predefined operations to handle its intent and content separately.
Although most operations can be categorized into insertion, deletion, and modification, the semantic intent and content cannot be easily decoupled into discrete rules.
In fact, the storage procedure essentially predicts how current information will be used in the future.
Because rule-based procedures operate in a discrete function space, they struggle to achieve optimal prediction performance.

The primary goal of the memory utilization procedure is to assist inference with the stored information.
From the foundation model perspective, it can be considered as letting the required information of the input instruction participate in the forward computation.
For example, answering \textit{``Where does Bob live now?''} requires previously stored living information to facilitate inference.
Thus, a native memory utilization procedure should directly map the input instruction and memory state to the generated output.
External memory designs rules to trigger retrieval, reranking, and concatenation.
However, the information requirement cannot be defined and captured by discrete and finite rules.
For example, an instruction may require information based on semantic similarity, emotion, or even complex combinations of implicit metrics.
The memory utilization procedure predicts the information requirements, which is coupled with the inference process.
Therefore, it should not be divided into discrete stages limited by discrete function spaces.

Consequently, the memory procedure should be modeled within a continuous function space and implemented via numerical computation, which is tightly coupled with the forward computation of the backbone.
In a memory foundation model, the native memory procedure autonomously executes both memory storage and utilization during the forward computation.
This native memory procedure should be established during the pre-training or mid-training stage.
In addition, this memory procedure paradigm has significant advantages in both efficiency and end-to-end optimization.

\subsection{Comparison with Previous Works}

\textbf{Test-time Training.}
Memory foundation models differ from test-time training (TTT) in three key aspects.
First, TTT typically adapts the model within a single sequence, where the dynamic parameters are updated to better fit the current input.
In contrast, memory foundation models are defined under multi-step interactions.
Their dynamic parameters serve as persistent native memory states that store information from previous steps and support future inference.

Second, TTT does not explicitly provide native memory procedures.
Its update is usually driven by self-supervised language modeling, which helps the model absorb prior information within the current sequence.
However, it does not specify how the model should semantically remember, forget, update, or reflect on information according to input instructions.
In contrast, memory foundation models are trained with memory reconstruction and operation objectives, enabling the model to autonomously execute semantic memory operations in the latent parametric space and transform the native memory state accordingly.

Third, many TTT methods are motivated by efficient long-context modeling, and they often introduce recurrent layers to replace full attention.
However, memory foundation models pursue a different goal.
They do not aim to replace the standard full-attention computation within the current step.
Instead, they introduce information from previous interaction steps through native memory states as residuals.

In summary, TTT is primarily a mechanism for inference-time adaptation, while memory foundation models formulate memory as a persistent, instruction-driven, and procedure-aware capability of foundation models.

\textbf{Memory-Augmented Neural Networks.}
Memory-augmented neural networks (MANNs) introduce additional memory modules to neural models, such as differentiable memory slots and learned read-write operations~\cite{graves2014neural}.
These models show that neural networks can store external information and retrieve it for later computation.
Nevertheless, memory foundation models differ in how memory is integrated with the backbone.
In MANNs, the memory module is usually a separate storage component controlled by a neural controller.
Although the operations can be differentiable, the memory is still external to the main model parameters, which are often designed independently from the backbone.

In contrast, memory foundation models internalize memory into the backbone computation.
The memory state is represented in a parametric form and participates directly in forward computation.
The memory procedure is also modeled by the same continuous function space as the backbone, rather than being implemented as a separate controller over explicit slots.
Therefore, memory foundation models can be regarded as a step from externally augmented memory toward native memory inside foundation models.

\textbf{Other Methods.}
Compared with In-place TTT~\cite{feng2026place}, MemGen~\cite{zhang2025memgen} and $\delta$-Mem~\cite{lei2026delta}, which still rely on textual memories in the context and use additional latent summaries to improve inference, memory foundation models remove textual memory from the context entirely.
Compared with MEMO~\cite{quek2026memo} and MemFT~\cite{xu2026lora}, which primarily handle offline documents, memory foundation models focus on test-time information.
Compared with Memory$^3$~\cite{yang2024memory3}, which takes an important step beyond textual RAG by encoding knowledge into retrievable explicit memories, memory foundation models further extend this direction toward native memory.
While Memory$^3$ primarily constructs explicit memories from offline corpora and retrieves them to augment attention computation, memory foundation models internalize information acquired from online interactions as persistent dynamic states within the backbone.
They further enable these states to be autonomously maintained and transformed through native memory procedures across multiple interaction steps.

\subsection{Discussion}
For memory foundation models, the onset of interaction represents a key transition from static to dynamic knowledge acquisition.
Knowledge acquired before interaction originates from offline pre-training and is retained in static parameters, while information received during interaction is acquired at test time and stored in dynamic memory states.
Therefore, $\theta_1$ can also serve as initial supplementary information outside of pre-training.
Because $\theta_1$ captures transferable domain knowledge, models deployed in similar domains can be initialized with the same $\theta_1$ to provide baseline information.

Moreover, native memory aligns with the evolving trend of foundation models.
Inspired by large reasoning models~\cite{guo2025deepseek}, we find that an agent's external capabilities can be expressed natively by the foundation model through optimization.
In other words, supervised data of target behaviors can activate internal capabilities and generalize them to other tasks.
These native capabilities can provide advantages in generalization, efficiency, and optimization properties.
Memory is also a critical agent capability that traditionally relies on external modules.
Therefore, we argue that memory can also be natively triggered through memory-specific tasks.
However, unlike reasoning, which is purely a process, memory also involves a storage entity.
It requires us to modify the model architecture to incorporate a storage entity as the memory state.
Then, both the memory state and procedures are supposed to be modeled under a collaborative function space.
This enables us to empower foundation models with memory capabilities via an optimization-driven approach.

\section{Metis Architecture}
\label{sec:metis_architecture}

\begin{figure}[!t]
	\centering
    \includegraphics[width=\textwidth]{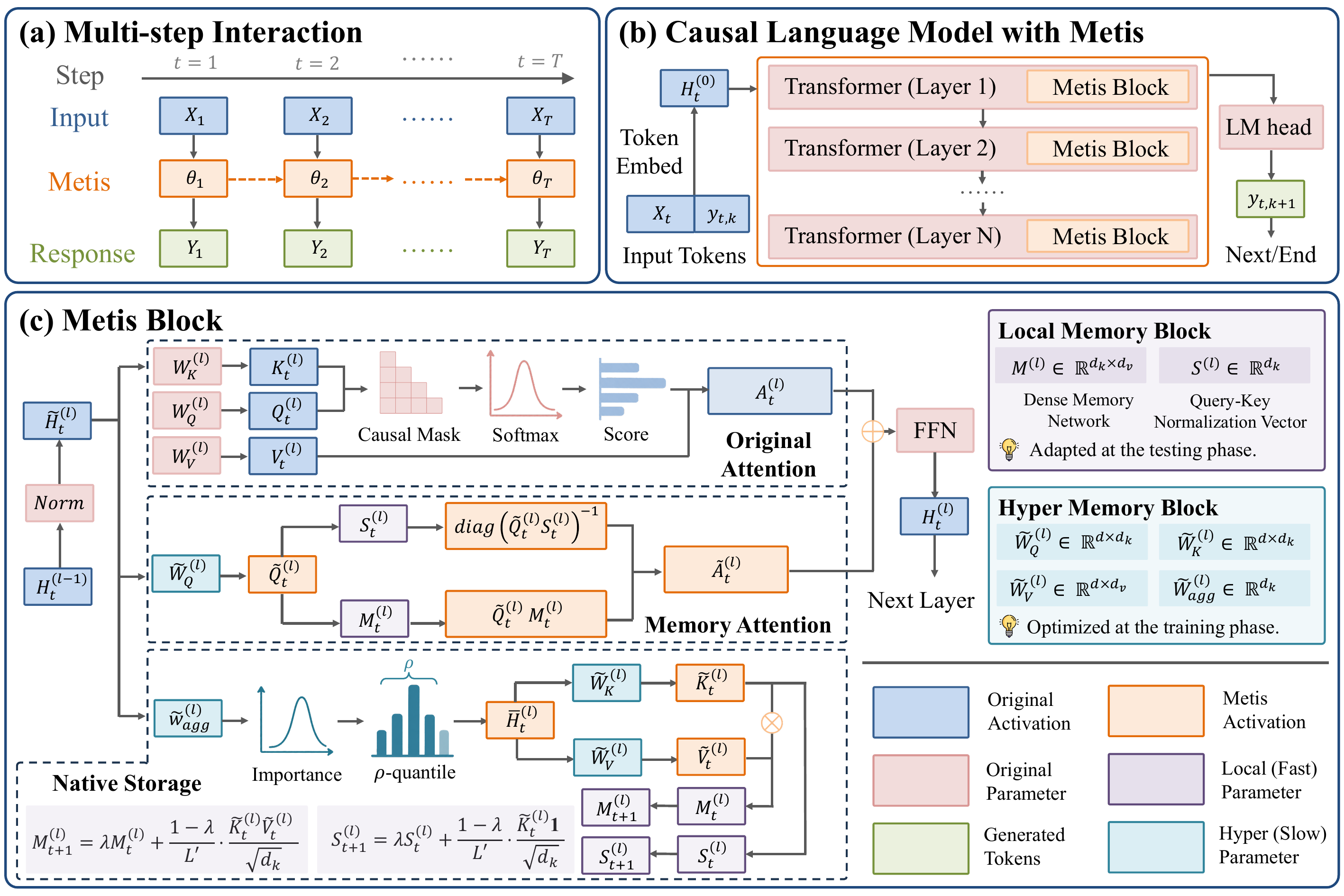}
	\vspace{-0.4cm}
	\caption{Overview of the Metis architecture.}
	\label{fig:method}
	\vspace{-0.4cm}
\end{figure}

In this section, we first present some preliminaries.
Then, we introduce Metis as the prototype of memory foundation models with the Metis block.
After that, we demonstrate how this architecture supports native memory storage and utilization procedures through computation.
Finally, we provide theoretical insights, theoretical error analysis, and further discussions.
The overview of the Metis framework is presented in \textbf{Figure~\ref{fig:method}}.

\subsection{Preliminaries}
We adopt causal language models as the primary implementation of memory foundation models, as they are dominantly used in modern foundation models.
We present the standard architecture of causal language models, which primarily consists of $N$ stacked Transformer blocks followed by a language modeling head.

\textbf{Transformer Block.}
To highlight the core architecture, we focus on causal self-attention and the feed-forward network (FFN), which are major components of modern Transformers.
We omit other details, such as positional embeddings, multi-head attention strategies, and hybrid attention mechanisms, as they can be directly incorporated into our framework.

We denote the input of the $l$-th Transformer block as $\mathbf{H}^{(l-1)} \in \mathbb{R}^{L \times d}$, where $L$ is the sequence length and $d$ is the dimension of hidden states.
After applying the pre-normalization function \text{PreNorm}, we obtain $\tilde{\mathbf{H}}^{(l)} = \text{PreNorm}(\mathbf{H}^{(l-1)}),$ as the input of causal self-attention.
We denote $\mathbf{W}_Q^{(l)}, \mathbf{W}_K^{(l)} \in \mathbb{R}^{d \times d_k}, \mathbf{W}_V^{(l)} \in \mathbb{R}^{d \times d_v}$ as query, key, and value projection matrices of this layer.
Then, we obtain the query state, key state, and value state of the $l$-th layer by
\begin{equation*}
    \mathbf{Q}^{(l)} = \tilde{\mathbf{H}}^{(l)}\mathbf{W}_Q^{(l)}, \quad \mathbf{K}^{(l)} = \tilde{\mathbf{H}}^{(l)}\mathbf{W}_K^{(l)}, \quad \mathbf{V}^{(l)} = \tilde{\mathbf{H}}^{(l)}\mathbf{W}_V^{(l)}.
\end{equation*}
After that, the causal self-attention can be calculated by
\begin{equation}
    \label{equ:original_attention}
    \mathbf{A}^{(l)} = \text{Softmax}\left(\frac{\mathbf{Q}^{(l)}(\mathbf{K}^{(l)})^\top}{\sqrt{d_k}} + \text{Mask}(L)\right)\mathbf{V}^{(l)},
\end{equation}
where $d_k$ is the attention head dimension, and $\text{Mask}(L)$ is the causal mask defined as $\text{Mask}(L)_{i,j} = -\infty$ if $j > i$, and $0$ otherwise.
Then, it adds the projected output attention to the residual after projection by 
\begin{equation*}
    \mathbf{H}^{\prime(l)} = \mathbf{H}^{(l-1)} + \mathbf{A}^{(l)}\mathbf{W}_O^{(l)},
\end{equation*}
where $\mathbf{W}_O^{(l)} \in \mathbb{R}^{d_v \times d}$.
Finally, $\mathbf{H}^{\prime(l)}$ is passed through FFN with a residual connection to get the $l$-th layer output by
\begin{equation*}
    \mathbf{H}^{(l)} = \mathbf{H}^{\prime(l)} + \text{FFN}(\text{Norm}(\mathbf{H}^{\prime(l)})),
\end{equation*}
where the activation $\mathbf{H}^{(l)}$ is also the input of $\left(l+1 \right)$-th layer.

\textbf{Causal Language Model.}
We denote the sequence of input tokens as $X = (x_1, x_2, \dots, x_L)$.
A causal language model first maps these discrete tokens into continuous vector representations.
Let $\mathbf{E} \in \mathbb{R}^{|\mathcal{V}| \times d}$ denote the token embedding matrix, where $|\mathcal{V}|$ is the vocabulary size. The initial hidden state $\mathbf{H}^{(0)} \in \mathbb{R}^{L \times d}$ is obtained by extracting the corresponding embeddings and combining them with positional embeddings.
After that, this initial representation is processed sequentially through the stack of $N$ Transformer blocks by
\begin{equation*}
    \mathbf{H}^{(l)} = \text{TransformerBlock}^{(l)}(\mathbf{H}^{(l-1)}), \quad \text{for } l = 1, 2, \dots, N.
\end{equation*}
Then, the final hidden state $\mathbf{H}^{(N)}$ represents the contextualized input, and the language modeling head maps this final state back to the vocabulary space to predict the probability distribution for the next token.
This process is commonly modeled by a linear projection after normalization, followed by a softmax function
\begin{equation*}
    P(\cdot \mid x_{\le i}) = \text{Softmax}(\text{Norm}(\mathbf{H}_{(i)}^{(N)}) \mathbf{W}_{\text{LM}}),
\end{equation*}
where $\mathbf{H}_{(i)}^{(N)} \in \mathbb{R}^d$ is the final hidden vector at position $i$, and $\mathbf{W}_{\text{LM}} \in \mathbb{R}^{d \times |\mathcal{V}|}$ represents the projection matrix.
After sampling $x_{i+1} \sim P(x_{i+1} \mid x_{\le i})$, this new token is appended to the sequence, and the model repeats the process until it decodes an end-of-sequence token or reaches the maximum length.

During the pre-training phase, the causal language model is optimized using the standard autoregressive next-token prediction objective.
It minimizes the negative log-likelihood of the training sequences by
\begin{equation*}
\theta = \arg\min_{\theta} \sum_{X \in \mathcal{D}} \sum_{i=1}^{|X|-1} -\log P(x_{i+1} \mid x_{\le i}; \theta),
\end{equation*}
where $\mathcal{D}$ represents the pre-training corpus and $\theta$ encompasses all the trainable parameters of the model.

\subsection{Native Memory State}
To implement the native memory state, we propose the Metis blocks inside Transformer blocks in \textbf{Figure~\ref{fig:method}(b)}, where each Metis block consists of a local memory block and a hyper memory block in \textbf{Figure~\ref{fig:method}(c)}.
The local memory blocks are responsible for maintaining the dense representation of prior information, while the hyper memory blocks construct parametric function spaces for native memory procedures to transform memory states.

\textbf{Local Memory Block.}
Local memory blocks maintain the memory state at the current step, so we define a dense memory network $\mathbf{M}^{(l)} \in \mathbb{R}^{d_k \times d_v}$ inside the $l$-th local memory block.
At the step $t$, we denote it as $\mathbf{M}^{(l)}_{t}$.
The model also maintains a query-key normalization vector as $\mathbf{S}^{(l)}_t \in \mathbb{R}^{d_k}$.
Both $\mathbf{M}^{(l)}$ and $\mathbf{S}^{(l)}_t$ are dynamic parameters updated across different steps.
Specifically, we set $\mathbf{M}^{(l)}_1=\mathbf{0}$ and $\mathbf{S}^{(l)}_1=\mathbf{0}$ by default.

\textbf{Hyper Memory Block.}
Hyper memory blocks are responsible for updating the dynamic parameters in local memory blocks based on the intermediate activations of the current input $X_t$ and output $Y_t$.
Each of them consists of static parameters obtained through mid-training, which remain unchanged during interactions.
It serves as the parametric foundation of the native memory storage procedure.
Specifically, each hyper memory block is parameterized by several optimizable parameters.
First, it has a learnable importance vector $\tilde{\mathbf{w}}_{\text{agg}}^{(l)} \in \mathbb{R}^{d}$, which scores the intermediate activations for adaptive aggregation.
In addition, we set the memory key and value projection matrices $\tilde{\mathbf{W}}_K^{(l)}\in \mathbb{R}^{d \times d_k}$ and $\tilde{\mathbf{W}}_V^{(l)} \in \mathbb{R}^{d \times d_v}$, which map the selected hidden states into the memory keys and values for the local memory.
We also set the memory query projection matrix $\tilde{\mathbf{W}}_Q^{(l)}\in \mathbb{R}^{d \times d_k}$ to reduce the error in the native memory utilization procedure.

\subsection{Native Memory Procedure}
The native memory procedure consists of memory storage and utilization procedures, as we discuss in \textbf{Section~\ref{subsec:mfm_native_procedure}}.
In the native memory storage procedure of our framework, hyper memory blocks update local memory blocks as part of the model computation, based on intermediate activations.
In the native memory utilization procedure, local memory blocks incorporate the current memory states into the forward computation.

\textbf{Native Memory Storage Procedure.}
After completing step $t$, we denote the input hidden states at the $l$-th layer as $\mathbf{H}^{(l-1)}_t$.
Then, the hyper memory block aggregates it into a compact representation through a learned adaptive aggregation.
Specifically, we first pre-normalize the hidden states as $\tilde{\mathbf{H}}_t^{(l)} = \text{PreNorm}(\mathbf{H}^{(l-1)}_t)$ and score each of the $L$ tokens with a learnable importance vector $\tilde{\mathbf{w}}_{\text{agg}}^{(l)} \in \mathbb{R}^{d}$, obtaining an importance distribution
\begin{equation*}
    \mathbf{p}_t^{(l)} = \text{Softmax}\left( \frac{\tilde{\mathbf{H}}_t^{(l)} \tilde{\mathbf{w}}_{\text{agg}}^{(l)}}{\tau} \right) \in \mathbb{R}^{L},
\end{equation*}
where $\tau$ is a temperature coefficient.
Then, we obtain a subset of positions based on the top-$\rho$.
We sort these probabilities in descending order as $p_{(1)} \ge p_{(2)} \ge \cdots \ge p_{(L)}$, and keep the smallest prefix whose cumulative value reaches the threshold $\rho$.
The number of selected positions can be expressed by
\begin{equation*}
    L_t' = \text{clip}\left( \min\Big\{ k : \sum\nolimits_{r=1}^{k} p_{(r)} \ge \rho \Big\},\ K_{\min},\ L \right),
\end{equation*}
where $K_{\min}$ denotes the minimum number of selected positions.
These $L_t'$ positions with the highest scores form the selected set $\mathcal{S}_t^{(l)}$, and we gather corresponding hidden states by
\begin{equation*}
    \bar{\mathbf{H}}_t^{(l)} = \mathbf{\Pi}_t^{(l)} \tilde{\mathbf{H}}_t^{(l)} \in \mathbb{R}^{L_t' \times d}, \quad \text{with } L_t' \ll L,
\end{equation*}
where $\mathbf{\Pi}_t^{(l)} \in \{0, 1\}^{L_t' \times L}$, whose rows are the one-hot indicators of $\mathcal{S}_t^{(l)}$.
Since the top-$\rho$ selection is non-differentiable, we adopt a straight-through estimator that routes the gradients through the dense distribution $\mathbf{p}_t^{(l)}$, making the scorer $\tilde{\mathbf{w}}_{\text{agg}}^{(l)}$ end-to-end trainable.
After that, we compute the projected memory key states and memory value states as
$$
\tilde{\mathbf{K}}_t^{(l)} = \bar{\mathbf{H}}_t^{(l)} \tilde{\mathbf{W}}_K^{(l)}, \quad \tilde{\mathbf{V}}_t^{(l)} = \bar{\mathbf{H}}_t^{(l)} \tilde{\mathbf{W}}_V^{(l)}.
$$
Finally, the dense memory network is updated based on $\tilde{\mathbf{K}}_t^{(l)}$ and $\tilde{\mathbf{V}}_t^{(l)}$ by
\begin{equation}
    \mathbf{M}^{(l)}_{t+1} = \lambda \mathbf{M}^{(l)}_{t} + \frac{(1 - \lambda)}{L_t'} \cdot\frac{\tilde{\mathbf{K}}_t^{(l)\top} }{\sqrt{d_k}}\tilde{\mathbf{V}}_t^{(l)},
\end{equation}
where $\lambda$ represents the discount factor.
In addition, the query-key normalization vector can be updated by
\begin{equation}
    \mathbf{S}^{(l)}_{t+1} = \lambda \mathbf{S}^{(l)}_{t} + \frac{(1 - \lambda)}{L_t'} \cdot\frac{\tilde{\mathbf{K}}_t^{(l)\top} \mathbf{1}}{\sqrt{d_k}}.
\end{equation}
This native storage procedure is presented in \textbf{Figure~\ref{fig:method}(c)}.
Based on the constructed function space, we aim to internalize various memory operations into the model's computation through optimization. Specifically, the selection and projection provide the model with compression capabilities.
Meanwhile, semantic-based computation enables memory instructions to be understood and applied within the latent space.
In practice, we find that replacing the linear update with a Gated Delta Network (GDN)-based~\cite{yang2025gated} update obtains better performance, so Metis finally adopts the GDN-based update (GDU) strategy.
\textbf{Section~\ref{subsec:ablation_studies}} and \textbf{Appendix~\ref{app:lu_gdu_scaling}} compares the two implementations through ablation studies.

\textbf{Native Memory Utilization Procedure.}
We define the memory attention as
\begin{equation}
    \tilde{\mathbf{A}}^{(l)}_t = \text{diag}\left( \tilde{\mathbf{Q}}^{(l)}_{t} \mathbf{S}^{(l)}_{t}\right)^{-1}  \tilde{\mathbf{Q}}^{(l)}_t \mathbf{M}^{(l)}_{t},
\end{equation}
where $\tilde{\mathbf{Q}}^{(l)}_{t}=\tilde{\mathbf{H}}_{t}^{(l)}\tilde{\mathbf{W}}_Q^{(l)}$ denotes the memory query states with optimizable parameter $\tilde{\mathbf{W}}_Q^{(l)} \in \mathbb{R}^{d \times d_k}$.
In practice, we add an identity vector to the normalization denominator to prevent numerical overflow and improve numerical stability.
Then, the memory attention is integrated into the main branch of attention, and replaces \textbf{Equation~(\ref{equ:original_attention})} with
\begin{equation}
    \mathbf{A}^{(l)}_t = \gamma \cdot  \text{Softmax}\left(\frac{\mathbf{Q}^{(l)}_t(\mathbf{K}^{(l)}_t)^\top}{\sqrt{d_k}} + \text{Mask}(L)\right)\mathbf{V}_t^{(l)} + (1-\gamma) \cdot \text{Norm}\left(\tilde{\mathbf{A}}^{(l)}_t\right),
\end{equation}
where $\text{Norm}(\cdot)$ is applied to the memory readout to align its scale with the original attention branch, $\mathbf{K}^{(l)}_t, \mathbf{V}^{(l)}_t$ are input key states and value states at the current step, and $\gamma \in [0, 1]$ balances the two branches.

\subsection{Theoretical Insight of Native Memory Procedures}
We provide theoretical insights on how information from previous steps influences subsequent inference through Metis blocks.
At step $t$, we prepend an additional virtual memory prefix $\mathbf{P}^{(l)}_t \in \mathbb{R}^{L_p\times d}$ to the input $\tilde{\mathbf{H}}^{(l)}_t$ of the $l$-th attention layer, resulting in the augmented input
$$\hat{\mathbf{H}}^{(l)}_t = 
\begin{bmatrix}
\mathbf{P}^{(l)}_t \\
\tilde{\mathbf{H}}^{(l)}_t
\end{bmatrix}.
$$
Then, we compute the corresponding query state $\hat{\mathbf{Q}}^{(l)}_t$ as follows
$$
\hat{\mathbf{Q}}^{(l)}_t =
\hat{\mathbf{H}}^{(l)}_t \mathbf{W}_Q^{(l)} =
\begin{bmatrix}
\mathbf{P}^{(l)}_t \mathbf{W}_Q^{(l)} \\
\mathbf{Q}_t^{(l)}
\end{bmatrix}.
$$
Similarly, we have the key states and value states
$$
\hat{\mathbf{K}}^{(l)}_t = \begin{bmatrix}
\mathbf{P}^{(l)}_t \mathbf{W}_K^{(l)} \\
\mathbf{K}_t^{(l)}
\end{bmatrix}, \quad
\hat{\mathbf{V}}^{(l)}_t = 
\begin{bmatrix}
\mathbf{P}^{(l)}_t \mathbf{W}_V^{(l)} \\
\mathbf{V}_t^{(l)}
\end{bmatrix}.
$$
The attention output $\hat{\mathbf{A}}_t$ is then computed using a modified causal mask $\text{Mask}(L_p + L) \in \mathbb{R}^{(L_p+L) \times (L_p+L)}$ by
$$
\hat{\mathbf{A}}_t = \text{Softmax} \left( \frac{\hat{\mathbf{Q}}^{(l)}_t \hat{\mathbf{K}}_t^{(l)\top}}{\sqrt{d_k}} + \text{Mask}(L_p + L) \right) \hat{\mathbf{V}}_t^{(l)}.
$$
Specifically, we divide the causal mask into four parts as follows
$$
\text{Mask}(L_p + L) = 
\left[
\begin{array}{c|c}
\mathbf{0}_{L_p \times L_p} & -\infty_{L_p \times L} \\
\hline
\mathbf{0}_{L \times L_p} & \text{Mask}(L)
\end{array}
\right],
$$
where $\mathbf{0}_{L \times L_p}$ allows the virtual memory prefix tokens to be visible to input tokens.
Then, we decompose the calculation of attention as 
$$
\hat{\mathbf{A}}_t = 
\begin{bmatrix}
    \text{Softmax} \left( \frac{\left( \mathbf{P}^{(l)}_t \mathbf{W}_Q^{(l)} \right) \left( \mathbf{P}^{(l)}_t \mathbf{W}_K^{(l)} \right)^\top}{\sqrt{d_k}} \right) & \mathbf{0}_{L_p \times L} \\
    \text{Softmax}^* \left( \frac{\mathbf{Q}_t^{(l)} \left( \mathbf{P}^{(l)}_t \mathbf{W}_K^{(l)} \right)^\top}{\sqrt{d_k}} \right) & \text{Softmax}^* \left( \frac{\mathbf{Q}_t^{(l)} \mathbf{K}_t^{(l)\top}}{\sqrt{d_k}} + \text{Mask}(L) \right)
\end{bmatrix} 
\begin{bmatrix}
\mathbf{P}^{(l)}_t \mathbf{W}_V^{(l)} \\
\mathbf{V}_t^{(l)}
\end{bmatrix},
$$
where $\text{Softmax}^*\left( \cdot \right)$ denotes the global softmax function applied to the entire row.
Then, we retain the attention outputs corresponding to the non-virtual tokens by
\begin{equation}
\label{eq:original_memory_attention}
\mathbf{A}^{(l)}_t = \underbrace{\text{Softmax}^* \left( \frac{\mathbf{Q}_t^{(l)} \mathbf{K}_t^{(l)\top}}{\sqrt{d_k}} + \text{Mask}(L) \right) \mathbf{V}_t^{(l)}}_{\text{Original Attention}} +
\underbrace{\text{Softmax}^* \left( \frac{\mathbf{Q}_t^{(l)} \left( \mathbf{P}^{(l)}_t \mathbf{W}_K^{(l)} \right)^\top}{\sqrt{d_k}} \right) \left( \mathbf{P}^{(l)}_t \mathbf{W}_V^{(l)} \right)}_{\text{Memory Attention}}.
\end{equation}

Let $\mathbf{z}_{\text{orig}}$ and $\mathbf{z}_{\text{mem}}$ denote the partition items of the original attention and memory attention
\begin{equation}
    \mathbf{z}_{\text{orig}} = \exp \left( \frac{\mathbf{Q}_t^{(l)} \mathbf{K}_t^{(l)\top}}{\sqrt{d_k}} + \text{Mask}(L) \right) \mathbf{1}_{L}, \quad
\mathbf{z}_{\text{mem}} = \exp \left( \frac{\mathbf{Q}_t^{(l)} \left( \mathbf{P}^{(l)}_t \mathbf{W}_K^{(l)} \right)^\top}{\sqrt{d_k}} \right) \mathbf{1}_{L_p}.
\end{equation}
Then, we perform element-wise division by its element-wise sum to get the weighting matrices
$$
\mathbf{\Lambda}_{\text{orig}} = \text{diag}\left( \mathbf{z}_{\text{orig}} \oslash \left( \mathbf{z}_{\text{orig}} + \mathbf{z}_{\text{mem}} \right) \right), \quad
\mathbf{\Lambda}_{\text{mem}} = \text{diag}\left( \mathbf{z}_{\text{mem}} \oslash \left( \mathbf{z}_{\text{orig}} + \mathbf{z}_{\text{mem}} \right) \right) = \mathbf{I} - \mathbf{\Lambda}_{\text{orig}}.
$$

Then, \textbf{Equation~(\ref{eq:original_memory_attention})} is equivalent to the equation with the normal Softmax function for each part:
$$
\mathbf{A}^{(l)}_t = \mathbf{\Lambda}_{\text{orig}} \cdot \text{Softmax} \left( \frac{\mathbf{Q}_t^{(l)} \mathbf{K}_t^{(l)\top}}{\sqrt{d_k}} + \text{Mask}(L) \right) \mathbf{V}_t^{(l)} +
\left( \mathbf{I} - \mathbf{\Lambda}_{\text{orig}} \right) \cdot \text{Softmax} \left( \frac{\mathbf{Q}_t^{(l)} \left( \mathbf{P}^{(l)}_t \mathbf{W}_K^{(l)} \right)^\top}{\sqrt{d_k}} \right) \left( \mathbf{P}^{(l)}_t \mathbf{W}_V^{(l)} \right).
$$
To control the influence of the two attention components, we introduce a global weighting parameter $\gamma \in [0, 1]$ to approximate the original weighting matrices by
$$
\mathbf{A}^{(l)}_t = \gamma \cdot \text{Softmax} \left( \frac{\mathbf{Q}_t^{(l)} \mathbf{K}_t^{(l)\top}}{\sqrt{d_k}} + \text{Mask}(L) \right) \mathbf{V}_t^{(l)} + (1-\gamma) \cdot
\text{Softmax} \left( \frac{\mathbf{Q}_t^{(l)} \left( \mathbf{P}^{(l)}_t \mathbf{W}_K^{(l)} \right)^\top}{\sqrt{d_k}} \right) \left( \mathbf{P}^{(l)}_t \mathbf{W}_V^{(l)} \right).
$$
Then, we denote this specific memory attention part for $\mathbf{P}^{(l)}_t$ as
$$
\check{\mathbf{A}}^{(l)}_t = \text{Softmax} \left( \frac{\mathbf{Q}_t^{(l)} \left( \mathbf{P}^{(l)}_t \mathbf{W}_K^{(l)} \right)^\top}{\sqrt{d_k}} \right) \left( \mathbf{P}^{(l)}_t \mathbf{W}_V^{(l)} \right).
$$
We define the function of similarity between $\mathbf{Q}_t^{(l)}$ and $\mathbf{P}^{(l)}_t \mathbf{W}_K^{(l)}$ as
$$
\text{Sim}\left(\mathbf{Q}_t^{(l)}, \mathbf{P}^{(l)}_t \mathbf{W}_K^{(l)}\right) = \text{exp}\left( \frac{\mathbf{Q}_t^{(l)} \left( \mathbf{P}^{(l)}_t \mathbf{W}_K^{(l)} \right)^\top}{\sqrt{d_k}} \right).
$$
Then, the memory attention can be rewritten as
$$
\check{\mathbf{A}}^{(l)}_t = \text{diag}\left( \text{Sim}\left(\mathbf{Q}_t^{(l)}, \mathbf{P}^{(l)}_t \mathbf{W}_K^{(l)}\right) \cdot \mathbf{1} \right)^{-1} \text{Sim}\left(\mathbf{Q}_t^{(l)}, \mathbf{P}^{(l)}_t \mathbf{W}_K^{(l)}\right) \left( \mathbf{P}^{(l)}_t \mathbf{W}_V^{(l)} \right).
$$
In order to decompose the $\mathbf{Q}_t^{(l)}$ part and $\mathbf{P}^{(l)}_t \mathbf{W}_K^{(l)}$ part, we approximate the similarity with
$$
\text{Sim}\left(\mathbf{Q}_t^{(l)}, \mathbf{P}^{(l)}_t \mathbf{W}_K^{(l)}\right) = \frac{\mathbf{Q}_t^{(l)} \left( \mathbf{P}^{(l)}_t \mathbf{W}_K^{(l)} \right)^\top }{\sqrt{d_k}}.
$$

So the memory attention can be rewritten as
\begin{equation}
    \check{\mathbf{A}}^{(l)}_t = \text{diag}\left(  \mathbf{Q}_t^{(l)} \frac{ \left( \mathbf{P}^{(l)}_t \mathbf{W}_K^{(l)} \right)^\top}{\sqrt{d_k}} \cdot \mathbf{1} \right)^{-1}  \mathbf{Q}_t^{(l)}  \left[ \frac{ \left( \mathbf{P}^{(l)}_t \mathbf{W}_K^{(l)} \right)^\top}{\sqrt{d_k}} \left( \mathbf{P}^{(l)}_t \mathbf{W}_V^{(l)} \right) \right].
\end{equation}
Finally, we consider the prefix tokens $\mathbf{P}^{(l)}_t$ as the $c$-th ($c < t$) step aggregated results $\bar{\mathbf{H}}^{(l)}_c$, so we get
\begin{equation}
    \label{equ:ideal_memory_attention}
    \check{\mathbf{A}}^{(l)}_t = \text{diag}\left( \mathbf{Q}_t^{(l)} \frac{ \left( \bar{\mathbf{H}}^{(l)}_c \tilde{\mathbf{W}}_K^{(l)} \right)^\top}{\sqrt{d_k}} \cdot \mathbf{1} \right)^{-1}  \mathbf{Q}_t^{(l)} \left[ \frac{\left( \bar{\mathbf{H}}^{(l)}_c \tilde{\mathbf{W}}_K^{(l)} \right)^\top}{\sqrt{d_k}} \left( \bar{\mathbf{H}}^{(l)}_c \tilde{\mathbf{W}}_V^{(l)} \right) \right],
\end{equation}
where $\tilde{\mathbf{W}}_K^{(l)}, \tilde{\mathbf{W}}_V^{(l)}$ are parameters of the hyper memory block.
It is worth noting that, in practice, the reference step $c$ is not accessible in advance, and the evidence required at step $t$ may span more than a single step.
Meanwhile, the memory key states and memory value states of different steps are coupled together within the fixed-size memory network $\mathbf{M}^{(l)}_t$ and the normalization vector $\mathbf{S}^{(l)}_t$, so that the non-reference steps ($j \neq c$) inevitably leak into the readout as noise.
To mitigate the impact of such noise, instead of directly reusing the vanilla attention query $\mathbf{Q}_t^{(l)}$ in \textbf{Equation~(\ref{equ:ideal_memory_attention})}, we utilize a trainable memory query $\tilde{\mathbf{Q}}^{(l)}_{t} = \tilde{\mathbf{H}}^{(l)}_{t} \tilde{\mathbf{W}}_Q^{(l)}$ to get
\begin{equation}
    \label{equ:memory_attention}
    \check{\mathbf{A}}^{(l)}_t = \text{diag}\left( \tilde{\mathbf{Q}}_t^{(l)} \frac{ \left( \bar{\mathbf{H}}^{(l)}_c \tilde{\mathbf{W}}_K^{(l)} \right)^\top}{\sqrt{d_k}} \cdot \mathbf{1} \right)^{-1}  \tilde{\mathbf{Q}}_t^{(l)} \left[ \frac{\left( \bar{\mathbf{H}}^{(l)}_c \tilde{\mathbf{W}}_K^{(l)} \right)^\top}{\sqrt{d_k}} \left( \bar{\mathbf{H}}^{(l)}_c \tilde{\mathbf{W}}_V^{(l)} \right) \right],
\end{equation}
where $\tilde{\mathbf{W}}_Q^{(l)} \in \mathbb{R}^{d \times d_k}$ is an optimizable projection.
This decouples the memory query from the original attention and offers the freedom to reshape the cross-step similarities $\tilde{\mathbf{Q}}^{(l)}_{t} \tilde{\mathbf{K}}_j^{(l)\top}$.
Therefore, it can emphasize the relevant steps while suppressing the irrelevant ones, thereby reducing the influence of noise when using the memory query states.
We provide a detailed theoretical error analysis in \textbf{Section~\ref{subsec:theoretical_insight_memory_states}}.

\subsection{Theoretical Error Analysis}
\label{subsec:theoretical_insight_memory_states}

Unlike standard Transformers that store all historical KV pairs in a growing cache, the hyper memory block compresses information into a fixed-size matrix.
Suppose the model requires extracting information from the $c$-th step.
The dense memory network can be expressed as 
$$
\mathbf{M}^{(l)}_{t} = \sum_{j=1}^{t-1} \lambda^{t - (j+1)} \cdot \frac{(1 - \lambda)}{L_j'} \cdot\frac{\tilde{\mathbf{K}}_j^{(l)\top}}{\sqrt{d_k}} \tilde{\mathbf{V}}_j^{(l)}.
$$
Then, we use the memory query $\tilde{\mathbf{Q}}^{(l)}_{t} = \tilde{\mathbf{H}}^{(l)}_{t} \tilde{\mathbf{W}}_Q^{(l)}$ to extract information from the dense memory network and query-key normalization vector by
\begin{equation*}
\begin{aligned}
    \tilde{\mathbf{A}}_{t}^{(l)} = & \text{diag}\left(  \tilde{\mathbf{Q}}^{(l)}_{t}  \mathbf{S}^{(l)}_{t}\right)^{-1} \tilde{\mathbf{Q}}^{(l)}_{t} \mathbf{M}^{(l)}_{t} \\
    = & \text{diag}\left( \sum_{j=1}^{t-1} \lambda^{t - (j+1)} \frac{(1 - \lambda)}{L_j'}  \tilde{\mathbf{Q}}^{(l)}_{t} \frac{ \tilde{\mathbf{K}}_j^{(l)\top} }{\sqrt{d_k}} \mathbf{1} \right)^{-1} \\
    & \cdot \left( \sum_{j=1}^{t-1} \lambda^{t - (j+1)} \cdot \frac{(1 - \lambda)}{L_j'}  \tilde{\mathbf{Q}}^{(l)}_{t} \frac{\tilde{\mathbf{K}}_j^{(l)\top}}{\sqrt{d_k}} \tilde{\mathbf{V}}_j^{(l)} \right). 
\end{aligned}
\end{equation*}
We define the individual terms in the above summation as
$$
\mathbf{U}_{j} = \lambda^{t - (j+1)} \frac{(1 - \lambda)}{L_j'}  \tilde{\mathbf{Q}}^{(l)}_{t}  \frac{ \tilde{\mathbf{K}}_j^{(l)\top} }{\sqrt{d_k}} \mathbf{1},
$$
$$
\mathbf{R}_{j'} = \lambda^{t - (j'+1)} \cdot \frac{(1 - \lambda)}{L_{j'}'}  \tilde{\mathbf{Q}}^{(l)}_{t}  \frac{\tilde{\mathbf{K}}_{j'}^{(l)\top}}{\sqrt{d_k}} \tilde{\mathbf{V}}_{j'}^{(l)},
$$
and $ \tilde{\mathbf{A}}_{t}^{(l)}$ can be rewritten as 
\begin{equation*}
\begin{aligned}
     \tilde{\mathbf{A}}_{t}^{(l)} = & \text{diag}\left( \sum_{j=1}^{t-1} \mathbf{U}_j \right)^{-1} \left( \sum_{j'=1}^{t-1} \mathbf{R}_{j'} \right).
\end{aligned}
\end{equation*}
Assume that our target information is stored at the $c$-step (\textit{i.e.,} the similarity $\tilde{\mathbf{Q}}^{(l)}_{t} \tilde{\mathbf{K}}_{c}^{(l)}$ is significantly higher than others). Then, we can further rewrite the equation as
\begin{equation*}
\begin{aligned}
     \tilde{\mathbf{A}}_{t}^{(l)} = & \text{diag}\left( \mathbf{U}_c + \sum_{j=1, j\neq c}^{t-1} \mathbf{U}_j \right)^{-1} \left( \mathbf{R}_{c} + \sum_{j'=1, j' \neq c}^{t-1} \mathbf{R}_{j'} \right). \\
     = & \left(\text{diag}\left( \mathbf{U}_c \right) \left( \mathbf{I} + \text{diag}\left( \mathbf{U}_c \right)^{-1} \cdot \text{diag}\left( \sum_{j=1, j\neq c}^{t-1} \mathbf{U}_j \right) \right) \right)^{-1} \left( \mathbf{R}_{c} + \sum_{j'=1, j' \neq c}^{t-1} \mathbf{R}_{j'} \right).
\end{aligned}
\end{equation*}
According to the first-order Taylor expansion, we have
\begin{equation*}
\begin{aligned}
     \tilde{\mathbf{A}}_{t}^{(l)}
     \approx & \left( \mathbf{I} - \text{diag}\left( \mathbf{U}_c \right)^{-1} \cdot \text{diag}\left( \sum_{j=1, j\neq c}^{t-1} \mathbf{U}_j \right) \right) \text{diag}\left( \mathbf{U}_c \right)^{-1} \left( \mathbf{R}_{c} + \sum_{j'=1, j' \neq c}^{t-1} \mathbf{R}_{j'} \right), \\
     = & \text{diag}\left( \mathbf{U}_c \right)^{-1} \mathbf{R}_{c} + \underbrace{\text{diag}\left( \mathbf{U}_c \right)^{-1} \sum_{j'=1, j' \neq c}^{t-1} \mathbf{R}_{j'}}_{\boldsymbol{\epsilon}_1} - \underbrace{\text{diag}\left( \mathbf{U}_c \right)^{-1} \cdot \text{diag}\left( \sum_{j=1, j\neq c}^{t-1} \mathbf{U}_j \right) \text{diag}\left( \mathbf{U}_c \right)^{-1} \mathbf{R}_{c}}_{\boldsymbol{\epsilon}_2} \\
     - & \underbrace{\text{diag}\left( \mathbf{U}_c \right)^{-1} \cdot \text{diag}\left( \sum_{j=1, j\neq c}^{t-1} \mathbf{U}_j \right) \text{diag}\left( \mathbf{U}_c \right)^{-1} \sum_{j'=1, j' \neq c}^{t-1} \mathbf{R}_{j'}}_{\boldsymbol{\epsilon}_3},
\end{aligned}
\end{equation*}
where the first term is equivalent to $\check{\mathbf{A}}^{(l)}_{t}$ in \textbf{Equation~(\ref{equ:memory_attention})} with $\tilde{\mathbf{K}}_c^{(l)} = \bar{\mathbf{H}}^{(l)}_c \tilde{\mathbf{W}}_K^{(l)}$ and $\tilde{\mathbf{V}}_c^{(l)} = \bar{\mathbf{H}}^{(l)}_c \tilde{\mathbf{W}}_V^{(l)}$ by
\begin{equation*}
\begin{aligned}
    \text{diag}\left( \mathbf{U}_c \right)^{-1} \mathbf{R}_{c} = & \text{diag}\left( \tilde{\mathbf{Q}}^{(l)}_{t} \frac{ \tilde{\mathbf{K}}_c^{(l)\top} }{\sqrt{d_k}} \mathbf{1} \right)^{-1}  \tilde{\mathbf{Q}}^{(l)}_{t}  \frac{\tilde{\mathbf{K}}_{c}^{(l)\top}}{\sqrt{d_k}} \tilde{\mathbf{V}}_{c}^{(l)} \\
    = & \text{diag}\left(  \tilde{\mathbf{Q}}^{(l)}_{t}  \frac{ \left(\bar{\mathbf{H}}^{(l)}_c \tilde{\mathbf{W}}_K^{(l)}\right)^\top }{\sqrt{d_k}} \mathbf{1} \right)^{-1}  \tilde{\mathbf{Q}}^{(l)}_{t}  \frac{\left(\bar{\mathbf{H}}^{(l)}_c \tilde{\mathbf{W}}_K^{(l)}\right)^\top}{\sqrt{d_k}} \left( \bar{\mathbf{H}}^{(l)}_c \tilde{\mathbf{W}}_V^{(l)} \right)\\
    = & \check{\mathbf{A}}^{(l)}_{t}.
\end{aligned}
\end{equation*}
Therefore, there are three error terms for $||\tilde{\mathbf{A}}_{t}^{(l)} - \check{\mathbf{A}}^{(l)}_{t}||_2$.
It is worth noting that $\boldsymbol{\epsilon}_2$ and $\boldsymbol{\epsilon}_3$ are structural errors caused by global normalization, whereas $\boldsymbol{\epsilon}_1$ is an attention error introduced by irrelevant information.
Across these three terms, there is always at least one factor of $ \tilde{\mathbf{Q}}^{(l)}_{t}  \tilde{\mathbf{K}}_{j}^{(l)\top}$ where $j\neq c$ in the summation. Therefore, when this similarity is low, the resulting error is expected to be small.

\subsection{Efficiency Analysis}

Metis introduces native memory with limited additional inference overhead compared with external memory.
The key reason is that the original attention, memory attention, and memory storage procedure can be largely executed in parallel.
For the $l$-th layer at step $t$, the original attention branch computes token-token attention over the current input, while the memory utilization branch performs memory attention over the native memory state $\mathbf{M}^{(l)}_t$ and $\mathbf{S}^{(l)}_t$.
These two branches depend on the same input hidden states but have no sequential dependency on each other.
Therefore, memory attention does not need to wait for the output of the original attention, and its results can be fused only after both branches finish.

The memory storage procedure can also be decoupled from the current inference path.
It updates the memory state for future steps, while the current step only reads from the existing memory state.
Thus, after the required hidden states are available, the storage branch can be executed in parallel with the original attention and memory utilization, instead of becoming an additional sequential stage.
As a result, the layer-level latency can be expressed as
\begin{equation}
T_{\text{parallel}}^{(l)}
=
\max\left(
T_{\text{orig}}^{(l)},
T_{\text{util}}^{(l)},
T_{\text{store}}^{(l)}
\right)
+
T_{\text{fuse}}^{(l)}.
\end{equation}

Moreover, Metis stores historical information in fixed-size native memory states, rather than appending retrieved textual memories to the input context.
Therefore, its memory utilization cost depends mainly on the memory state size, instead of growing linearly with the number of historical interactions.
This enables Metis to provide native memory capabilities while avoiding the retrieval, concatenation, and prefilling overhead commonly introduced by external memory systems.

\section{Data Construction}
\label{sec:data_construction}

In order to build Metis by mid-training based on general foundation models, we synthesize a comprehensive training dataset based on existing public datasets.
This dataset consists of primary data and auxiliary data, which are used for training native memory procedures and improving generalization in complex scenarios.

\subsection{Primary Data}
\label{subsec:primary_data}
The primary data serves as the core supervision for training native memory procedures.
It is designed to teach memory foundation models to perform different memory operations in the forward computation through optimization, thereby generalizing to various scenarios.
The native memory procedure is acquired through optimization rather than manual rules, so the primary data must provide explicit supervision for the desired memory operations.

\textbf{Data Principles.}
We highlight two data principles.
First, the data should be structured as a temporally ordered sequence of interaction steps, which mirrors the time-streaming nature of online information.
Second, the data should be state-consistent.
The response to a later query must agree with the memory state shaped by earlier operations.
Together, these two properties teach the model to store information and use it at the appropriate later step.

Instead of generating data from scratch, we synthesize the primary data from established public benchmarks.
This choice offers three advantages.
Mature benchmarks provide verified facts and reasoning chains, which reduce hallucination when we extend them into long interaction sequences.
Their broad coverage of fiction, science, news, and logical reasoning enriches the context and improves generalization.
In addition, every synthetic sample is anchored to a source fact, which keeps the corpus traceable and easy to verify.

\textbf{Data Summary.}
We select 27 public benchmarks across four memory operations, as shown in \textbf{Table~\ref{tab:primary_data_summary}}.
We organize the primary data along three orthogonal dimensions: 
(1) \underline{Memory operation} includes \textit{remember}, \textit{forget}, \textit{update}, and \textit{reflect}, which together span the core behaviors of native memory.
For every operation, a structured fact serves as the unit of synthesis, and the final query is answerable only from the information introduced in the preceding turns.
A \textit{remember} sample states a fact and then queries it, whereas a \textit{reflect} sample introduces several single-hop facts and then queries their multi-hop composition.
An \textit{update} sample modifies a previously stated fact before the query, and a \textit{forget} sample revokes a previously stated fact before the query.
(2) \underline{Salience of the instruction} ranges from explicit memory commands to implicit statements that embed information within natural narratives.
(3) \underline{Noise level}, where clean sequences form the basic case and noisy sequences are produced by inserting irrelevant turns.
Jointly, these dimensions encourage the memory procedure to generalize across operations, instruction styles, and noise levels.

\begin{table}[t]
\centering
\small
\renewcommand{\arraystretch}{1.3}
\caption{Summary of the primary data. Samples for each memory operation are synthesized from public benchmarks and follow a distinct multi-turn skeleton, where the answer to the final query stays consistent with the preceding memory operations.}
\label{tab:primary_data_summary}
\vspace{-0.2cm}
\resizebox{1.0\textwidth}{!}{
\begin{tabular}{l l p{9.2cm}}
\hline
\hline
\textbf{Operation} & \textbf{Interaction Streaming} & \textbf{Source Benchmarks} \\
\hline
Remember   & Info($\text{A}_1$) $\to$ Query($\text{A}$)                       & LoCoMo~\cite{maharana2024evaluating}, LongMemEval~\cite{wu2024longmemeval}, NeedleInAHaystack~\cite{kamradt2023needle}, RULER~\cite{hsieh2024ruler}, LongBench~\cite{bai2024longbench}, $\infty$Bench~\cite{zhang2024bench}, L-Eval~\cite{an2024eval}, BABILong~\cite{kuratov2024babilong}, Bamboo~\cite{dong2024bamboo}, NaturalQuestions~\cite{kwiatkowski2019natural}, LongChat-Eval~\cite{li2023long} \\
\hline
Update     & Info($\text{A}_1$) $\to$ Info($\text{A}_2$) $\to$ Query($\text{A}$)          & ZsRE~\cite{levy2017zero}, RippleEdits~\cite{cohen2024evaluating}, KnowEdit~\cite{zhang2024comprehensive}, TemporalWiki~\cite{jang2022temporalwiki} \\
\hline
Forget     & Info($\text{A}_1$) $\to$ Info($\bar{\text{A}}_1$) $\to$ Query($\text{A}$)          & TOFU~\cite{maini2024tofu}, WMDP~\cite{li2024wmdp}, MUSE~\cite{shi2025muse}, RWKU~\cite{cao2024rwku}, WhoIsHarryPotter~\cite{eldan2023s}, BLUR~\cite{hu2026blur}, LKF~\cite{singh2025unlearning}, CLEAR~\cite{dontsov2025clear}, CounterFact~\cite{meng2022locating} \\
\hline
Reflect & Info($\text{A}_1$) $\to$ Info($\text{B}_1$) $\to$ Query($\text{A} \cap \text{B}$)               & MuSiQue~\cite{trivedi2022musique}, StrategyQA~\cite{geva2021did}, Bamboogle~\cite{press2023measuring} \\
\hline
\hline
\end{tabular}}
\vspace{-0.2cm}
\end{table}

\textbf{Construction Pipeline.}
Our data synthesis pipeline comprises three major steps, including seed extraction, static synthesis, and quality verification.

\textit{Step 1: Seed Extraction.}
From each source dataset, we extract the source reference, query, and answer to form a base dialogue.
Then, we summarize the underlying fact into a structured seed, which records a subject, a relation, and a target, together with operation-specific fields such as the updated target or the multi-hop chain.
We also collect a pool of distractor dialogues that are logically orthogonal to each query, which are used to extend the sequence length.

\textit{Step 2: Static Synthesis.}
Guided by the structured seed, a strong instruction-following language model rewrites each base dialogue into two salience styles.
The explicit style phrases the reference as a clear memory instruction, while the implicit style states the same fact as a description without an explicit instruction.
To cover long-range memory, we insert a variable number of distractor turns between the reference and the query, which yields the distract variant of both styles.

\textit{Step 3: Quality Verification.}
A language model acts as an automatic judge and filters samples according to several quality criteria.
The consistency check confirms that the final answer faithfully reflects the intended memory state.
The orthogonality check ensures that inserted distractors do not leak the core fact, and the shortcut check removes any query that can be answered without its reference.
We additionally monitor the semantic diversity of the queries to prevent template collapse.
Samples that fail any check are discarded, so that only reliable samples enter the final corpus.

\textbf{Data Statistics.}
We report the statistics of the synthesized primary data in \textbf{Table~\ref{tab:primary_data_stats}}.
After filtering, the corpus contains 357{,}137 samples and about 406 million tokens, drawn from 27 source benchmarks.
The samples are distributed across explicit, implicit, and distractor styles, which balance instruction salience and noise level.
The token count is dominated by the distractor samples, especially for \textit{remember}, because long irrelevant contexts are inserted to strengthen long-range memory.
This profile indicates that the primary data covers diverse memory operations at varied interaction lengths, which provides a solid basis for training native memory procedures.

\begin{table}[t]
\renewcommand{\arraystretch}{1.3}
\centering
\small
\caption{Statistics of the synthesized primary data. The explicit, implicit, and distract columns report the number of samples of each style, and the last column reports the total token count in millions.}
\label{tab:primary_data_stats}
\vspace{-0.2cm}
\begin{tabular}{c c c c c c c}
\hline
\hline
\textbf{Operation} & \textbf{Sources} & \textbf{Explicit} & \textbf{Implicit} & \textbf{Distract} & \textbf{All Samples} & \textbf{Tokens (M)} \\
\hline
Remember   & 11 & 14{,}682  & 13{,}671 & 28{,}502  & 56{,}855  & 362.0 \\
Forget     & 9  & 59{,}900  & 8{,}251  & 68{,}120  & 136{,}271 & 21.7  \\
Update     & 4  & 33{,}452  & 7{,}300  & 40{,}749  & 81{,}501  & 11.0  \\
Reflect & 3  & 20{,}646  & 20{,}615 & 41{,}249  & 82{,}510  & 11.4  \\
\hline
\textbf{Total} & \textbf{27} & \textbf{128{,}680} & \textbf{49{,}837} & \textbf{178{,}620} & \textbf{357{,}137} & \textbf{406.1} \\
\hline
\hline
\end{tabular}
\vspace{-0.2cm}
\end{table}

\subsection{Auxiliary Data}
\label{subsec:auxiliary_data}
The auxiliary data is used to improve the model's generalizability.
It further enhances the capabilities of memory foundation models for complex scenarios, such as multi-entity tasks and mixed dialogues.

\textbf{Construction Principles.}
The primary data consists of basic memory operations interactions, ranging from single-fact operations to multi-fact reasoning cases.
However, real interactions are more complex.
Multiple similar facts may coexist, some facts may be revoked while others persist, and memory turns are often interleaved with ordinary conversation.

The first is interference, where the model confuses similar facts or allows a forgetting operation to corrupt a retained fact in parametric spaces.
The second is memory pollution, where the model applies stored values to questions that do not need them.
The auxiliary data complements the primary data by targeting exactly these scenarios.
It preserves the same fact-level structure, but composes facts and dialogues into more challenging interaction patterns, which improves the generalization and robustness of native memory.

\textbf{Data Summary.}
We organize the auxiliary data into four subtypes, shown in \textbf{Table~\ref{tab:auxiliary_data_summary}}.
The first two subtypes address multi-fact scenarios.
(1) \underline{Multi-entity binding} jointly stores two confusable facts and queries both, which trains the model to bind each value to its own fact.
(2) \underline{Selective forgetting} revokes one fact while the other persists, which trains the model to forget one fact selectively without collateral loss.
The other two subtypes address memory pollution.
(3) \underline{Post-memory dialogue} continues an ordinary conversation right after a memory query, so the model does not carry stored values into unrelated answers.
(4) \underline{Memory-irrelevant dialogue} answers a question that does not need memory even when a memory state exists, so the model learns when memory should not influence the response.
Therefore, these four subtypes extend the primary data to realistic mixed interactions.

\begin{table}[t]
\centering
\small
\renewcommand{\arraystretch}{1.2}
\caption{Summary of the auxiliary data. Each subtype composes facts or dialogues into a complex interaction pattern, where the final answers remain consistent with the intended memory state.}
\label{tab:auxiliary_data_summary}
\vspace{-0.2cm}
\begin{tabular}{p{4.2cm} p{5.2cm} p{5.2cm}}
\hline
\hline
\textbf{Auxiliary Subtype} & \textbf{Interaction Streaming} & \textbf{Construction Source} \\
\hline
Multi-Entity Binding & Info($\text{A}_1$) $\to$ Info($\text{B}_1$) $\to$ Query($\text{A}$) $\to$ Query($\text{B}$) & Paired facts synthesized from primary source facts \\
\hline
Selective Forgetting & Info($\text{A}_1$) $\to$ Info($\text{B}_1$) $\to$ Info($\bar{\text{B}}_1$) $\to$ Query($\text{B}$) $\to$ Query($\text{A}$) & Paired facts synthesized from primary source facts \\
\hline
Post-Memory Dialogue & Info($\text{A}_1$) $\to$ Query($\text{A}$) $\to$ Chat & Primary memory samples with curated normal dialogues \\
\hline
Memory-Irrelevant Dialogue & Info($\text{A}_1$) $\to$ Chat \; / \; Chat $\to$ Chat & Primary memory samples with curated normal dialogues \\
\hline
\hline
\end{tabular}
\vspace{-0.2cm}
\end{table}

\textbf{Construction Pipeline.}
The auxiliary data is built from two shared ingredients, including synthesized paired facts and prepared normal dialogues, which are then formulated into the four subtypes.

\textit{Paired Fact Synthesis.}
The multi-entity binding and selective forgetting subtypes require pairs of similar facts.
For each source fact, represented by a subject, relation, and value, we synthesize one confusable counterpart fact.
Each counterpart is generated using one of four transformations relative to the source fact.
It keeps the subject but changes the relation, keeps the relation but changes the subject, imitates the value format, or stays semantically adjacent.
A language model generates each counterpart, and a verifier discards any fact that contradicts, restates, or depends on the source.
We then rewrite the verified facts into natural statements, queries, and revocation snippets to ensure fluency and diversity.

\textit{Normal Dialogue Preparation.}
The post-memory dialogue and memory-irrelevant dialogue subtypes require conversations that do not require access to memory.
We prepare a dialogue pool from three sources.
These are general assistant dialogues for everyday requests, open-domain conversations from public corpora, and entity-related dialogues that are topically related to a stored fact yet remain answerable without it.
We filter out turns with memory cues, real-time facts, or unsafe content, and we remove duplicates.

\textit{Subtype Formulation.}
The multi-entity binding subtype states the paired facts in turn and then queries both, which forces the model to bind each value to its correct fact.
The selective forgetting subtype states both facts, revokes one, and then queries both, so the revoked fact becomes unavailable while the retained fact stays correct.
The post-memory dialogue subtype appends an unrelated ordinary turn after a memory query, so the model returns to normal conversation without leaking any stored value.
The memory-irrelevant dialogue subtype keeps the memory state but drops its query before an ordinary turn, and it also includes standalone dialogues that carry no memory at all.

\textbf{Data Statistics.}
We report the statistics of the synthesized auxiliary data in \textbf{Table~\ref{tab:auxiliary_data_stats}}.
The paired-fact synthesis yields 76{,}153 natural snippet sets after quality verification.
These snippets support 76{,}153 multi-entity binding samples and 76{,}153 selective forgetting samples.
The dialogue-based subtypes are larger, because they reuse the full set of primary memory samples.
Post-memory dialogue contributes 357{,}137 samples, and memory-irrelevant dialogue contributes 100{,}000 samples.
In total, the auxiliary data adds 609{,}443 samples that emphasize multi-fact reasoning and pollution-resistant conversation.
Together with the primary data, it provides broad coverage from single-fact operations to complex mixed interactions.

\begin{table}[t]
\renewcommand{\arraystretch}{1.3}
\centering
\small
\caption{Statistics of the synthesized auxiliary data. We report the number of samples for each subtype, and we exclude every sample that fails quality verification.}
\label{tab:auxiliary_data_stats}
\vspace{-0.2cm}
\begin{tabular}{>{\centering\arraybackslash}p{3.0cm}>{\centering\arraybackslash}p{4.5cm}>{\centering\arraybackslash}p{2.5cm}}
\hline
\hline
\textbf{Target} & \textbf{Auxiliary Subtype} & \textbf{Samples}  \\
\hline
\multirow{2}{*}{Multi-fact Scenario} & Multi-Entity Binding & 76{,}153     \\
 & Selective Forgetting & 76{,}153     \\
 \hline
\multirow{2}{*}{Memory Pollution}    & Post-Memory Dialogue & 357{,}137   \\
 & Memory-Irrelevant Dialogue  & 100{,}000    \\
\hline
\textbf{All} & \textbf{Total}      & \textbf{609{,}443}  \\
\hline
\hline
\end{tabular}
\vspace{-0.2cm}
\end{table}

\section{Model Optimization}
\label{sec:model_optimization}
To empower Metis with native memory procedures, we design multiple training objectives for mid-training.
These objectives primarily consist of memory reconstruction, memory operation, and regularization.
The three objectives share a common likelihood form but operate on different data.
They jointly shape the native memory state and procedure.

\subsection{Overview}
We organize every training sample as a multi-step interaction $s = \{(X_t, Y_t)\}_{t=1}^{T_s}$, following the definition in \textbf{Section~\ref{sec:memory_foundation_model}}.
At step $t$, the model reads the input instruction $X_t$ and generates the assistant response $Y_t = (y_{t,1}, \dots, y_{t,|Y_t|})$.
All steps are forwarded sequentially, and the native memory procedure stores information from each step to the memory state before the next step starts.
Therefore, the parameters $\theta_t$ at step $t$ already integrate the memory state from all preceding steps $\{(X_i, Y_i)\}_{i<t}$.
We supervise only a subset of query steps $\mathcal{Q}_s \subseteq \{1, \dots, T_s\}$, where the assistant response is labeled, while the reference and operation steps remain unlabeled.
However, the responses of the reference and operation steps are still generated or provided for memory state updates.
In addition, all three objectives share the per-step loss below, and they differ only in how the supervised target $Y_t$ is constructed from the training data.
For a supervised step $t \in \mathcal{Q}_s$, we define the per-step loss as the token-averaged negative log-likelihood
\begin{equation}
\label{eq:per_step_loss}
\ell(s, t) = -\frac{1}{|Y_t|} \sum_{k=1}^{|Y_t|} \log P(y_{t,k} \mid X_t, Y_{t, <k}; \theta_t),
\end{equation}
where $Y_{t, <k}$ denotes the previously generated tokens at step $t$, and $\theta_t$ is conditioned on the memory state shaped by earlier steps.
The loss of a sample aggregates over its supervised steps as $\sum_{t \in \mathcal{Q}_s} \ell(s, t)$, which allows a single trajectory to supervise multiple responses.
During mid-training, we freeze the backbone parameters and optimize only the native memory parameters.

The three objectives correspond to five data subsets, and we control their contributions through a task-weighted sampler rather than explicit loss coefficients.
At training epoch $e$, the sampling probability of subset $\tau$ is
\begin{equation}
\label{eq:task_weight}
\pi_\tau(e) = \frac{w_\tau(e)}{\sum_{\tau' \in \mathcal{T}} w_{\tau'}(e)}, \qquad
w_\tau(e) = w_\tau^{\text{s}} + \left(w_\tau^{\text{e}} - w_\tau^{\text{s}}\right) \cdot \min\!\left(\frac{e}{E - 1}, 1\right),
\end{equation}
where $\mathcal{T}$ is the set of subsets, $E$ is the total number of epochs, and $w_\tau^{\text{s}}, w_\tau^{\text{e}}$ are the start and end weights of subset $\tau$.
This linear annealing forms a curriculum that gradually shifts the sampling mass from storage-oriented data toward harder long-range and regularization data.
Because the weights only modulate the sampling frequency, the expected mid-training objective can be written as
\begin{equation}
\label{eq:overall_objective}
\mathcal{L} = \sum_{\tau \in \mathcal{T}} \pi_\tau(e) \cdot \mathbb{E}_{s \sim \mathcal{D}_\tau} \!\left[ \sum_{t \in \mathcal{Q}_s} \ell(s, t) \right],
\end{equation}
where $\mathcal{D}_\tau$ is the data of subset $\tau$, and every sampled step contributes an unweighted loss from \textbf{Equation~(\ref{eq:per_step_loss})}.

\subsection{Memory Reconstruction Objective}
The memory reconstruction objective enables Metis to store and reconstruct information.
It provides an important training signal during the model's warm-up phase, as initialized models typically lack such capabilities.
Furthermore, it targets the upper bound of information storage, with completely lossless compression and reconstruction.
However, a trade-off exists between this objective and the native memory procedure.
First, reconstruction and instruction following are contradictory, as they require specificity and generalization, respectively.
Second, from the perspective of prediction tasks, the native memory procedure requires lossy compression guided by input instructions.
In contrast, memory reconstruction opposes lossy compression.

This objective is built on a reconstruction subset derived from the primary data in \textbf{Section~\ref{subsec:primary_data}}, denoted as $\mathcal{D}_{\text{rec}}$.
In each sample, a reference passage is presented and stored into the memory state at an early step, and a later query step requires the model to regenerate its content.
Because the supervised response $Y_t$ reproduces the stored reference, the memory state must retain the source with minimal loss.
We instantiate the per-step loss over this subset as
\begin{equation}
\label{eq:loss_reconstruction}
\mathcal{L}_{\text{rec}} = \pi_{\text{rec}}(e) \mathbb{E}_{s \sim \mathcal{D}_{\text{rec}}} \!\left[ \sum_{t \in \mathcal{Q}_s} \ell(s, t) \right],
\end{equation}
where the expectation averages over samples drawn from $\mathcal{D}_{\text{rec}}$, and $\ell(s, t)$ measures the negative log-likelihood of reconstructing the stored content at the query step $t$.
By step $t$, the reference passage has already been stored in the native memory state represented within $\theta_t$.
Minimizing $\mathcal{L}_{\text{rec}}$ thus drives the hyper memory block to encode the reference into a state from which the memory utilization procedure can recover it.

\subsection{Memory Operation Objective}
While reconstruction establishes lossless storage, native memory must additionally support input-driven operations.
The memory operation objective teaches Metis to remember, forget, update, and reflect, so that the memory state evolves according to the instruction at each step.
It is built on the primary data, which exhibits these operations under controlled instruction salience and noise.

We use two complementary subsets of the primary data.
The first subset, denoted as $\mathcal{D}_{\text{op}}^{\text{e/i}}$, contains the explicit and implicit samples.
Explicit samples phrase the operation as a clear command, whereas implicit samples embed the same information within a natural narrative.
This contrast forces the model to infer the operation from intent rather than from surface keywords.
The second subset, denoted as $\mathcal{D}_{\text{op}}^{\text{d}}$, contains the distractor samples, where irrelevant turns are inserted between the reference and the query.
It promotes long-range retention and robustness against intervening noise.

In all operation samples, the supervised response stays consistent with the information from the earlier steps.
For an \textit{update} sample, the answer reflects the new value rather than the old one.
For a \textit{forget} sample, the answer no longer exposes the forgotten value.
For a \textit{reflect} sample, the answer composes several stored facts into multi-hop reasoning.
Therefore, a single likelihood objective suffices to supervise all operations as
\begin{equation}
\label{eq:loss_operation}
\mathcal{L}_{\text{op}} = \pi_{\text{e/i}}(e) \mathbb{E}_{s \sim \mathcal{D}_{\text{op}}^{\text{e/i}}} \!\left[ \sum_{t \in \mathcal{Q}_s} \ell(s, t) \right] + \pi_{\text{d}}(e) \mathbb{E}_{s \sim \mathcal{D}_{\text{op}}^{\text{d}}} \!\left[ \sum_{t \in \mathcal{Q}_s} \ell(s, t) \right],
\end{equation}
where $\theta_t$ now encodes the net effect of the preceding operation sequence on the memory state.
Unlike reconstruction, the target is no longer a copy of the stored content, so the model learns to transform and read the memory state under the guidance of the instruction.

\subsection{Regularization Objective}
The reconstruction and operation objectives are primarily built on simple interaction patterns, which leave the model vulnerable in complex scenarios.
The regularization objective mitigates two failure modes that arise when memory operates in realistic interactions.
The first is interference, where similar facts are confused or a forgetting operation corrupts a retained fact.
The second is memory pollution, where stored values leak into responses that do not require them~\cite{lin2026surveylongtermmemorysecurity}.
This objective is built on the auxiliary data in \textbf{Section~\ref{subsec:auxiliary_data}}, which composes facts and dialogues into more complex and realistic interaction patterns.

We use two subsets of the auxiliary data.
The multi-fact subset $\mathcal{D}_{\text{mf}}$ targets interference.
Its multi-entity binding samples jointly present two confusable facts and query the model about both, which constrains the memory utilization procedure to bind each value to its own key.
Its selective forgetting samples include an instruction that revokes one fact while preserving the other, which constrains the forget operation to act locally.
The memory pollution subset $\mathcal{D}_{\text{mp}}$ targets leakage.
Its post-memory dialogue samples continue an ordinary conversation right after a memory query, and its memory-irrelevant samples answer a question that needs no memory even when a memory state exists.
In both cases, they discourage the model from injecting memory into unrelated responses.

These subsets act as regularization because they constrain memory behavior under more realistic and diverse interaction scenarios.
The supervised targets penalize cross-fact interference, collateral forgetting, and value leakage, which suppress degenerate solutions that always read or overwrite the memory state.
We define the objective as
\begin{equation}
\label{eq:loss_regularization}
\mathcal{L}_{\text{reg}} = \pi_{\text{mf}}(e) \mathbb{E}_{s \sim \mathcal{D}_{\text{mf}}} \!\left[ \sum_{t \in \mathcal{Q}_s} \ell(s, t) \right] + \pi_{\text{mp}}(e) \mathbb{E}_{s \sim \mathcal{D}_{\text{mp}}} \!\left[ \sum_{t \in \mathcal{Q}_s} \ell(s, t) \right],
\end{equation}
where many samples expose multiple supervised steps, so $|\mathcal{Q}_s| > 1$ jointly constrains the retained and the revoked facts within one interaction.
For the memory-irrelevant case, the supervised step is an ordinary turn whose target is independent of the memory state.

\section{Experiments}
\label{sec:experiments}

\subsection{Experimental Settings}
We evaluate Metis on memory operation tasks and memory-based question-answering (QA) tasks.
The memory operation task evaluates the performance of executing memory operations.
In addition, to verify the effectiveness of the native memory state, we evaluate the performance on the memory-based QA task.
Our major experiments focus on evaluating the native memory state and procedure primarily through relatively short-term tasks.
As for the long-term capability, we explore it from the perspective of memory capability in \textbf{Section~\ref{subsec:memory_capacity}}.

\textbf{Datasets and Metrics.}
For memory operations, we employ MemOps~\cite{hao2026memopsbenchmarkinglifecyclememory}, which is a specific benchmark focusing on memory operations, such as remembering, forgetting, and updating.
In the \emph{Full} setting, the model receives three complete evidence segments, containing 24 utterances.
In the \emph{Gold} setting, it receives only the oracle turns required for the question.
We also present the performance of the \underline{Test} set of our constructed dataset.
For the memory-based QA task, we conduct experiments on the golden-session setting of LoCoMo (\textit{i.e.,} LoCoMo (Gold)), where we provide the gold evidence sessions as input.
We also utilize the contextual generation task dataset from NextMem~\cite{zhang2026nextmem} for further analysis.
This dataset evaluates whether models can utilize the provided information to answer questions correctly, consisting of SQuAD~\cite{rajpurkar2016squad}, HotpotQA~\cite{yang2018hotpotqa}, LoCoMo~\cite{maharana2024evaluating}, and LongMemEval~\cite{wu2024longmemeval}.
In all these settings, we utilize gpt-4.1-mini to judge each prediction against its reference answer in three repeated evaluations.
Then, we report the median LLM-as-a-judge score.
In each dataset, we calculate the average performance (\textit{i.e.,} Avg.) across different types using a micro-average.
It should be noted that, to cover a wide range of entities for memory, we extract seed entities from various public datasets to synthesize our training data, such as LoCoMo and LongMemEval.
However, we do not leak their exact QA behaviors in the training phase.

\textbf{Baselines.}
We comprehensively evaluate our approach against four categories of baselines.
For backbone models evaluated with full information appended to the context, we utilize Qwen3.5~\cite{qwen3_5} across 4B, 9B, and 27B sizes.
For the partial-context baselines, we apply RAG~\cite{lewis2020retrieval} to these backbone models.
It encodes observations and queries into dense representations, and calculates the cosine similarity between queries and all observations. The top-$5$ observations are appended to the context.
For TTT-based models, we evaluate Temp-LoRA~\cite{wang2024greater} as the baseline.
It fuses information into the model by training a temporary LoRA module on previous text chunks during inference, encoding historical context as transient parameter updates.
Specifically, we implement Temp-LoRA with corresponding sizes of Qwen3.5 backbones.
Regarding parametric memory models, we compare with $\delta$-Mem~\cite{lei2026delta}, which steers attention with low-rank corrections. More details are provided in Appendix~\ref{app:evaluation_details}.

\textbf{Training Configuration.}
The reported Metis models are built upon Qwen3.5 backbones and trained on $8\times$H100 GPUs.
The backbone is frozen during training, and the trainable memory parameters are initialized using the key and value projection matrices of the corresponding backbone layers.
We use AdamW with a learning rate of $2\times10^{-4}$, a constant schedule after 200 warmup steps, weight decay 0.01, $\beta=(0.9,0.999)$, $\epsilon=10^{-8}$, and gradient clipping at 1.0.
Training uses BF16 and seed 42, and saves a checkpoint every 2,000 steps.
For Metis-4B, we train our model for 14,000 steps, corresponding to one epoch.
For Metis-27B, we use the same number of training steps, corresponding to approximately 0.4 epochs.
For Metis-9B, we use 8,000 steps (approximately 0.5728 epochs), which is selected by early stopping on validation-set performance.

\textbf{Evaluation Pipeline.}
For the memory operation and memory-based QA tasks, we adopt a static evaluation paradigm.
Each test trajectory is divided into two sequential phases.
The first phase consists of information steps, which provide the necessary context to the model.
The second phase consists of query steps, where the model must answer a question based on the prior information. 
Finally, the evaluation calculates performance metrics by comparing the model's output in the query step with the ground truth.
The prompts of the information step and query step are provided in \textbf{Appendix~\ref{appendix:prompt_baseline}}, and the prompts of LLM-as-a-Judge are presented in \textbf{Appendix~\ref{appendix:prompt_LLM_as_a_judge}}.

\subsection{Overall Performance}
\label{subsec:OverallPerformance}

\input{tables/overall_memop}

\textbf{Memory Operation Tasks.}
The results of MemOps in the gold setting (\textit{i.e.,} MemOps (Gold)) and the Metis test set are presented in \textbf{Table~\ref{tab:results_memoryops}}.
Due to the page limitation, we put the experiment results and analysis of MemOps in the full setting (\textit{i.e.,} MemOps (Full)) in \textbf{Appendix~\ref{appendix:extensive_results_memops_full}}.
As expected, full-context models achieve the strongest overall performance, while removing the context causes a substantial performance drop for standard backbones.
Partial context preserves some information on the Metis test set but performs poorly on MemOps (Gold), showing that incomplete histories cannot reliably support memory operations.
Temp-LoRA and $\delta$-Mem recover part of the lost performance, but their gains remain limited.
Under the same no-context setting, Metis achieves the best average results on both MemOps (Gold) and the Metis test set.
These results suggest that Metis can preserve information in its native memory state and use it in later steps without replaying the original context.

Metis-27B achieves the best average performance on both benchmarks under the no-context setting.
Compared with Metis-4B and Metis-9B, it shows clear gains in remembering, updating, reflection, and overall performance.
The improvement is especially large for forgetting on the Metis test set.
These results suggest that a sufficiently large backbone can better formulate and utilize the native memory state.
In addition, forgetting is still the most difficult operation on the external MemOps (Gold) benchmark, even for Metis-27B.
This suggests that removing or suppressing information in a shared latent state is more difficult to generalize than storing or updating information.
Overall, Metis shows strong performance on memory operation tasks in short-term scenarios.

\input{tables/overall_memqa}

\textbf{Memory-based QA Tasks.}
The results of the memory-based QA tasks are presented in \textbf{Table~\ref{tab:results_memory_based_QA}}.
Full-context models provide a strong upper bound because they can directly attend to the original evidence.
Their performance drops sharply when only partial context is available.
Without context, the original Qwen3.5 models obtain almost zero scores on LoCoMo (Gold), confirming that the answers cannot be reliably recovered from backbone knowledge alone.
In contrast, Metis achieves the best average performance on both benchmarks under the no-context setting.
Metis-27B obtains the highest score in every task category, outperforming other baselines.
These results show that the native memory state can preserve useful information and support question answering without replaying the original context.

The advantage of Metis is especially clear on tasks with complex or long-range memory requirements.
On NextMem, Metis achieves large gains on HotpotQA, LongMemEval, and LoCoMo subset.
It also substantially improves multi-hop and temporal question answering on LoCoMo (Gold).
It indicates that native memory remains effective for relatively simple factual questions while providing larger gains on more demanding tasks.
The strong improvement on temporal questions also suggests that a larger Metis model can better preserve and use relations across different interaction steps.
However, the gain on open-domain LoCoMo (Gold) questions is relatively limited, which indicates that some task types remain difficult even with increased model capacity.

In addition, we find that the improvement between Metis-4B and Metis-9B is modest, whereas Metis-27B substantially improves the average score.
This pattern suggests that backbone scaling can enhance native memory capability once model capacity is reached, although the gains are not uniform across tasks.
These results indicate that Metis provides strong memory-based QA performance in both the short-term QA tasks and the relatively longer LoCoMo (Gold) setting. 
In \textbf{Appendix~\ref{app:backbone_transferability}}, we further apply Metis to Llama~\cite{grattafiori2024llama} and Gemma~\cite{team2026gemma} models of varying sizes. We use the same mid-training and evaluation paradigm to explore its transferability across different backbone families and scales.

\subsection{Ablation Studies}
\label{subsec:ablation_studies}

We conduct ablation studies on Metis-4B from the perspectives of training data and model structure.
Following the main experimental setup, we evaluate LoCoMo (Gold) and NextMem in memory-based QA tasks. We also use the Metis test set and MemOps (Gold) in memory operation tasks.
All ablation models use Metis-4B and the same training configuration and evaluation pipeline as the main results.

\textbf{Data Ablation.}
We evaluate the contribution of different training data through two variants. In \underline{w/o MS}, we remove the Multi-fact Scenario data. In \underline{w/o MS+MP}, we remove the entire auxiliary dataset, including Multi-fact and Memory Pollution data, to examine its overall contribution to memory learning and generalization.

As shown in \textbf{Table~\ref{tab:ablation_4b_data_structure}}, removing the Multi-fact Scenario data consistently reduces performance across both types of tasks.
This result indicates that multi-fact supervision helps Metis integrate related information and maintain a coherent memory state.
The decline is more evident on MemOps (Gold) and the Metis test set, suggesting that such data is particularly important for learning reliable memory operations.
Removing the entire auxiliary dataset leads to a much larger overall degradation. The drop is especially clear on the Metis test set, while performance on LoCoMo (Gold) and NextMem also decreases consistently.
This shows that auxiliary data improves the robustness and generalization of native memory procedures across different scenarios.

\textbf{Structure Ablation.}
We further ablate the main components of the native memory procedure.
In \underline{w/o GDU}, we replace the GDU with a linear update (LU).
In \underline{w/o SA}, we remove the adaptive aggregation mechanism and directly use the last-token hidden state for memory storage.
In \underline{w/o OQ}, we remove the optimizable memory query projection and reuse the query from the original attention.
In \underline{w/o QKN}, we remove query-key normalization from memory attention.

Among all these evaluated variants, removing adaptive aggregation causes the largest performance drop.
Directly using the last token cannot effectively capture information distributed across the input sequence.
As a result, the model fails to construct an informative memory state.
In addition, removing query-key normalization also causes a substantial degradation, particularly on LoCoMo (Gold) and NextMem.
Without this normalization, irrelevant information may introduce stronger interference.
Furthermore, reusing the original attention query also reduces performance across all benchmarks. The decrease is larger on the memory-based QA tasks, indicating that a separate memory query is important for distinguishing relevant historical information from noise. This observation is consistent with our theoretical analysis, where the additional query projection reshapes cross-step similarities and suppresses interference from irrelevant memory.

We also find that replacing the GDU with a linear update has a small effect on the overall average.
The linear update performs slightly better on MemOps (Gold) and the Metis test set but is clearly weaker on LoCoMo (Gold).
This suggests that a linear update can handle simple and short-term memory operations, while the GDU provides a better balance in long-term scenarios.
This result is consistent with our engineering observation that GDU may produce more stable model behavior, motivating its use in Metis. \textbf{Appendix~\ref{app:lu_gdu_scaling}} further compares LU and GDU across model scales, while \textbf{Appendix~\ref{app:lu_downward_ablation}} repeats the data and structure ablations from the LU baseline.

\input{tables/ablation_4b_data_structure}

\subsection{Out-of-Distribution Memory Tasks}
\label{sec:ood-memory-generalization}

To examine whether the strong performance reported in \textbf{Section~\ref{subsec:OverallPerformance}} generalizes beyond the data-construction distribution, we further evaluate Metis on two out-of-distribution (OOD) benchmarks that were not used to construct the training data. All methods are evaluated under the no-context setting.

We evaluate ATM-Bench~\cite{mei2026atm} on its official standard split. For MemDaily~\cite{zhang2026memsim}, we use the subset of the official pre-generated release in which the annotated retrieval-target messages occur before the query. 
In the Gold setting, the model receives only benchmark-annotated evidence: human-annotated memory items represented as text in SGM for ATM-Bench, and retrieval-target messages for MemDaily. ATM-Bench scores list-recall, number, and open-ended questions using Jaccard similarity, post-processed exact match, and an LLM judge, respectively. MemDaily reports deterministic single-choice accuracy for all six question types.

\input{tables/ood_no_context}

As shown in \textbf{Table~\ref{tab:ood-no-context}}, Metis demonstrates strong OOD transfer on ATM-Bench, consistently outperforming the memory baselines across model scales and most question types. The advantage also holds for the deterministically scored number questions, indicating that the improvement is not merely an artifact of the LLM judge used for open-ended questions.
Since ATM-Bench requires models to retain and integrate heterogeneous evidence extracted from personal archives, these results suggest that the native memory procedure learned by Metis transfers beyond the patterns observed during training. 
The results on MemDaily are more mixed: Metis remains competitive but does not consistently lead the memory baselines. Together, the two benchmarks provide evidence that Metis's native memory capability generalizes to benchmarks not used in constructing its training data.

\subsection{Source-Exclusion Study}
\label{subsec:source-exclusion}

Complementing the OOD evaluation, we study how sensitive Metis is to the composition of the public sources used by the synthesis pipeline.
Specifically, we remove all training instances generated from LoCoMo and LongMemEval, and train Metis-4B, Metis-9B, and Metis-27B on the remaining data.
We keep other training and evaluation configurations unchanged.
The evaluation is memory-only, without replaying the original context.
Under the accounting used for this experiment, the exclusion removes only about 2.61\% of training instances, but these removals are concentrated in the remember, reconstruction, and multi-entity or mixed-operation slices.

\input{tables/source_exclusion}

As shown in \textbf{Table~\ref{tab:source-exclusion}}, the six-benchmark macro-average decreases at all three model sizes, but the reductions remain limited rather than producing a capability cliff.
The LoCoMo overall score and the LoCoMo subset of NextMem both decrease across all sizes, indicating measurable sensitivity to source composition.
However, LongMemEval and the NextMem average do not follow the same pattern, and both of them show improvement for Metis-27B.
A related contrast appears between the Metis test set and MemOps. They move in opposite directions, with the direction of the contrast reversing across model sizes.

Taken together, these results suggest that excluding these sources preserves most of the overall memory capability while redistributing performance across benchmarks, rather than causing uniform degradation or improvement.
In addition, the average across the two OOD benchmarks, ATM and MemDaily, decreases at every model size, although the two benchmarks do not change uniformly.

These changes may be explained from two perspectives.
From the perspective of memory as a prediction problem, narrower entity coverage may make the model less sensitive to information involving unseen or low-frequency entities and to how that information may be used in future interactions.
From the training-data perspective, a small but concentrated exclusion can produce a disproportionate change in task proportions, shifting the relative supervision across memory behaviors and potentially producing different performance trade-offs across benchmarks.

\subsection{Memory Capacity Studies}
\label{subsec:memory_capacity}
In this part, we further explore the long-term memory capability of Metis.
We evaluate this capability by modeling memory capacity, which comprises step-level and trajectory-level capacity.
Specifically, step-level capacity refers to the maximum number of tokens accommodated within a single update.
Similarly, trajectory-level capacity denotes the maximum number of update steps within an interaction trajectory.

We construct a testing dataset, which contains 20 fictional users defined over a shared schema of 40 distinct and atomic persona domains.
First, the model generates 40 fine-grained domains such as demographics, education and relationships.
Then, for each user, the domains are shuffled with a fixed seed and instantiated sequentially.
Previously generated attributes are provided as an immutable context to ensure logical consistency within each persona.
Each attribute is initially expanded into a biography-style sentence conditioned on the complete persona.
After that, one direct question is generated for each domain and reused across all users.
Finally, each biography-style message is rewritten as a concise first-person statement that expresses only the corresponding fact.
During evaluation, the 40 records of each user form an ordered trajectory.
The statements are sequentially stored in memory, and the queries are used to probe the model.
In addition, the user attributes serve as the gold answer for LLM-based judging.
Based on this dataset, we compare the performance of Qwen3.5-4B with full context and Metis-4B under two evaluation settings.

\begin{figure*}[t]
    \centering
\includegraphics[width=0.9\textwidth]{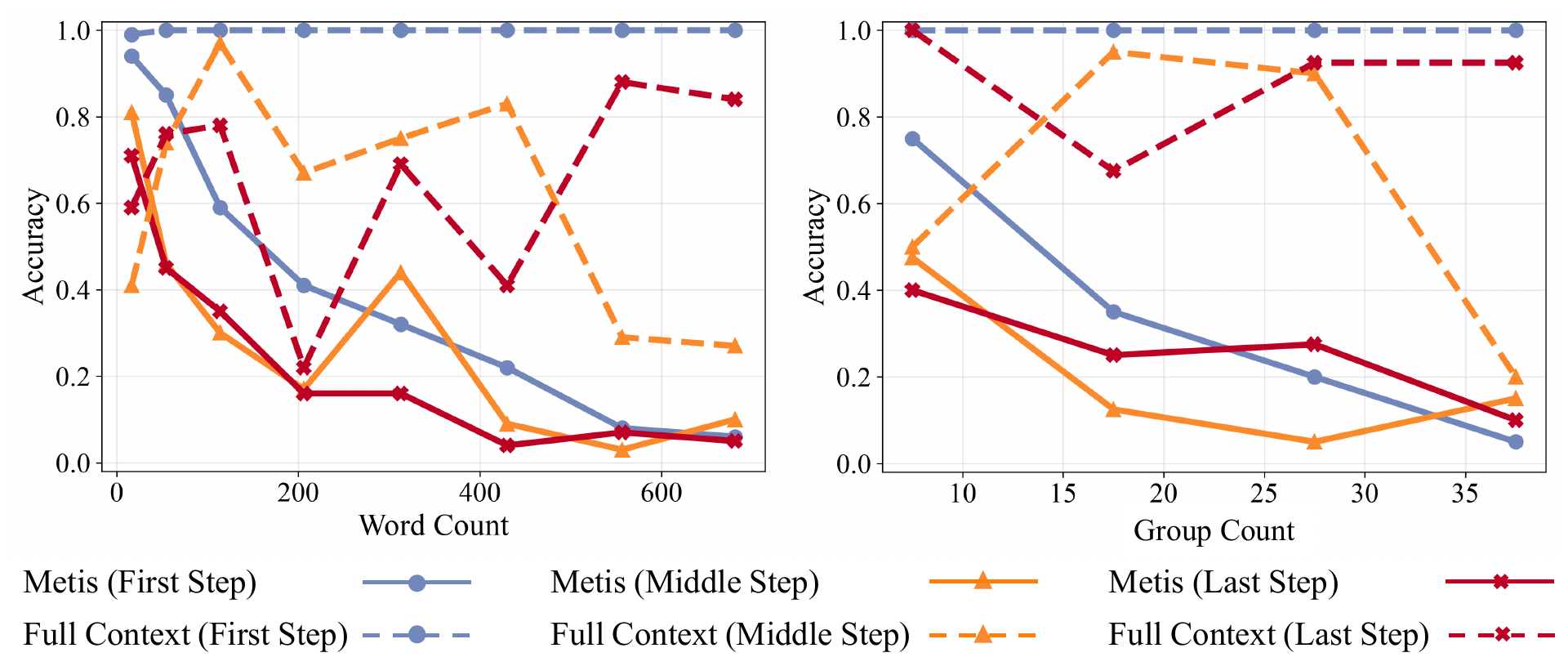}
    \caption{Results of memory capacity at the step-level (left) and trajectory-level (right).}
    \vspace{-0.4cm}
    \label{fig:memory_capacity}
\end{figure*}

\textbf{Step-level Capacity.}
This setting evaluates how much information the model can encode within a single memory update.
For each user of step $t$, the memory state is reset, and the first $t$ statements are concatenated and updated in one operation.
The model is then queried about the first, middle, and last facts in the updated content.
Since the full history is encoded from scratch at each step, this setting isolates the capacity of a single update as the input length increases. 
The results are presented in \textbf{Figure~\ref{fig:memory_capacity} (left)}.

For the step-level setting, Metis performs well when a single update contains only a small amount of information, but its accuracy decreases rapidly as the input becomes longer.
The performance on the first fact shows the clearest downward trend, while the middle and last facts exhibit larger fluctuations.
When the input exceeds several hundred words, performance at all three positions becomes low.
In contrast, the full-context baseline remains much stronger, especially for the first fact.

\textbf{Trajectory-level Capacity.}
This setting evaluates how much information the model can retain over a sequence of memory updates.
For each user, the memory state is reset only once and then accumulates throughout the trajectory.
The 40 statements are divided into consecutive groups of $g$ statements, with $g=5$ by default, and each group is concatenated and updated in Metis sequentially.
After every group-level update, the model is queried about the first, middle, and current updated facts.
This setting measures the capacity of the evolving memory state under repeated updates, with performance reported against the number of updated trajectory steps.
The results are presented in \textbf{Figure~\ref{fig:memory_capacity} (right)}.

For the trajectory-level setting, performance also declines as the number of updated steps increases.
The accuracy of the first fact decreases almost continuously, showing that early information is gradually weakened by later updates.
The middle and most recent facts also remain unstable, which suggests that new updates introduce interference throughout the whole memory state rather than only overwriting the oldest information.
Although the amount of information in each update is fixed, performance drops clearly as the trajectory becomes longer.
This confirms that repeated state transitions and accumulated compression errors form another major limitation of native memory.

\subsection{General Capability Studies}
\label{subsec:general_capability}
Previous experiments have demonstrated the effectiveness of Metis on memory-related tasks.
However, integrating native memory into the forward computation potentially influences the backbone's original behavior, which possibly decreases its general capabilities.
In this part, we further explore how Metis performs on the general tasks compared with its original backbone.
We compare Metis-4B with Qwen3.5-4B and report the performance difference between them.
Specifically, we design two settings to evaluate the general capabilities of Metis in different stages.
The first is the \underline{Initial Stage}, which measures performance on general tasks before any information is stored in memory, corresponding to step $t=1$.
In this setting, Metis is reset to an empty memory state before receiving the original prompt in benchmarks, and the backbone receives the same prompt. 
The second is the \underline{Active Stage}, which evaluates Metis after it has accumulated irrelevant information over previous interaction steps, where we have the step $t>1$.
For Metis, we reset the memory state and store task-irrelevant messages before providing the benchmark prompt.
For the backbone, we prepend the same messages to its prompt.
Our experiments are conducted under MMLU-Pro~\cite{wang2024mmlu}, IFEval~\cite{zhou2023instruction}, GSM8K~\cite{cobbe2021training}, and MMMLU~\cite{hendrycks2020measuring}.
In IFEval, we adopt the strict evaluation setting, which verifies instruction compliance directly on the original model response without applying the response transformations used by the loose criterion.
The detailed prompts of irrelevant messages are provided in \textbf{Appendix~\ref{appendix:prompt_general_capability}}.
We present the results in \textbf{Table~\ref{tab:general_capability_history}}.

\input{tables/general_capability_history}

The results show that Metis largely preserves the general capabilities of its original backbone at the initial stage. It shows only minor decreases on the other tasks. This indicates that the added memory architecture and memory-specific training do not substantially change the model's behavior when the memory state is empty.
A different trend appears at the active stage.
After irrelevant information is stored, Metis shows consistent performance drops across all benchmarks.
The degradation is moderate on MMLU-Pro, GSM8K, and MMMLU, but is much larger on IFEval.
This suggests that irrelevant native memory may introduce noise into the forward computation and interfere with the processing of the current general task, especially on strict instruction following.
Overall, Metis retains most of its original general capability before memory is activated, but drops as more information is stored in the memory states.

\subsection{Low-rank Decomposition}

\begin{figure*}[t]
    \centering
    \includegraphics[width=0.5\textwidth]{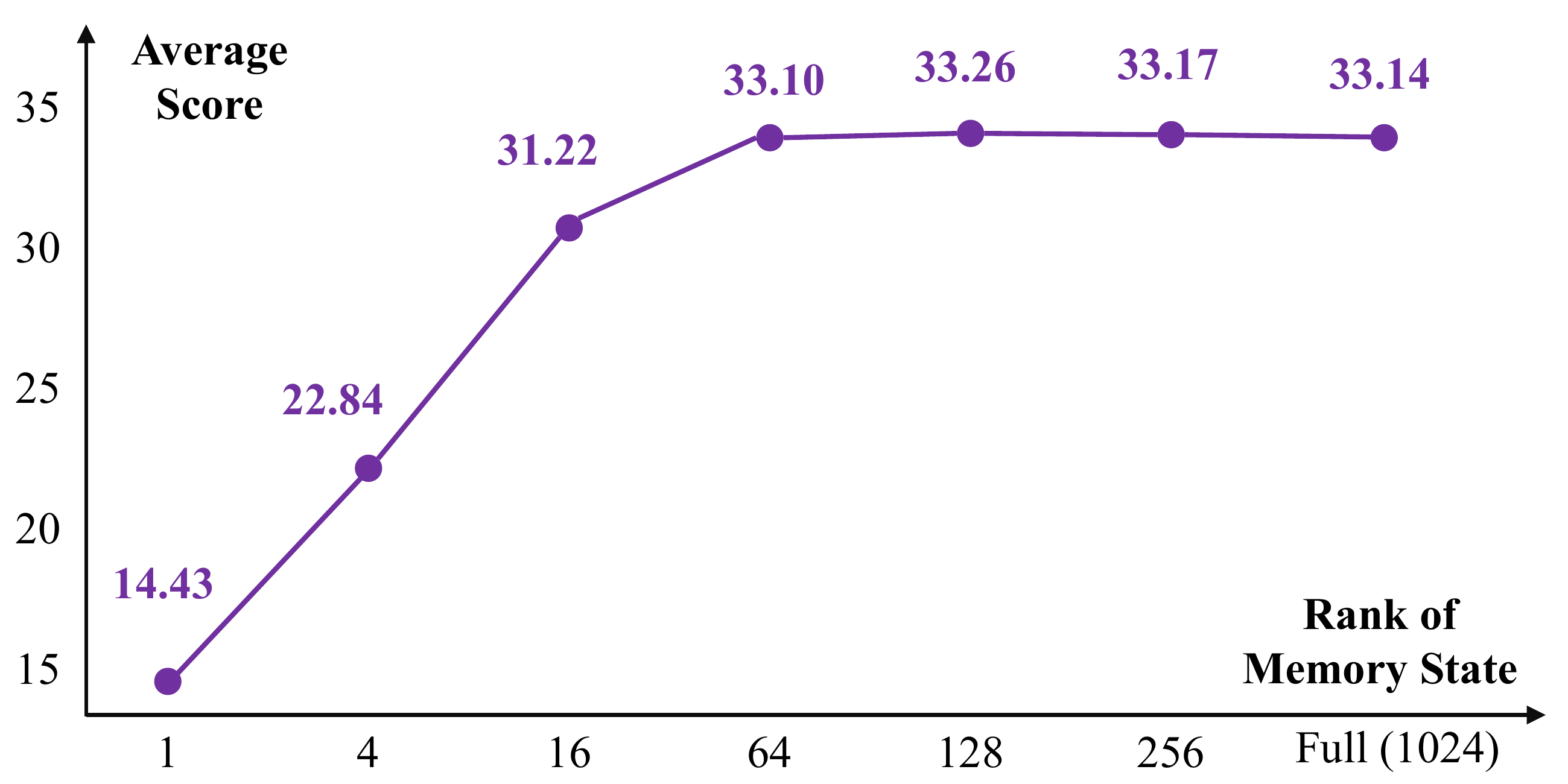}
    \caption{Results of low-rank decomposition under different ranks of memory states.}
    \label{fig:lowrank_4b_overall}
    \vspace{-0.2cm}
\end{figure*}

Storage overhead is a key efficiency metric for memory. In short-term tasks with limited information, parametric memory representations can be further compressed. Therefore, to explore the storage optimization potential of Metis, we apply low-rank decomposition to the memory states for efficient storage and reconstruct them before memory utilization.
Specifically, we cast the memory state $\mathbf{M}^{(l)}_t$ to FP32 and compute an SVD along its last two dimensions.
Therefore, for the retained rank $k$, the decomposition process can be represented as
\begin{equation*}
    \mathbf{M}^{(l)}_t = \mathbf{U}_t^{(l)}\mathbf{\Sigma}_t^{(l)} \mathbf{V}_t^{(l)\top}.
\end{equation*}
We maintain the low-rank approximation $\hat{\mathbf{U}}_t^{(l)}=\mathbf{U}_{t,:,1:k}^{(l)}$, $\hat{\mathbf{\Sigma}}_t^{(l)}=\mathbf{\Sigma}_{t,1:k,1:k}^{(l)}$, and $\hat{\mathbf{V}}_t^{(l)}=\mathbf{V}_{t,:,1:k}^{(l)}$ instead of the original full-rank factors.
Here, $\mathbf{U}_{t,:,1:k}^{(l)}$ and $\mathbf{V}_{t,:,1:k}^{(l)}$ denote the first $k$ columns of $\mathbf{U}_t^{(l)}$ and $\mathbf{V}_t^{(l)}$, respectively, while $\mathbf{\Sigma}_{t,1:k,1:k}^{(l)}$ denotes the leading $k \times k$ diagonal submatrix of $\mathbf{\Sigma}_t^{(l)}$.
Then, the reconstruction can be expressed as
\begin{equation*}
\hat{\mathbf{M}}^{(l)}_t
            = \hat{\mathbf{U}}_t^{(l)}\hat{\mathbf{\Sigma}}_t^{(l)} \hat{\mathbf{V}}_t^{(l)\top}.
\end{equation*}
We evaluate $k\in\{1,4,16,64,128,256\}$ on the four benchmarks with Metis-4B aligned with \textbf{Section~\ref{subsec:OverallPerformance}}.
We compare these models with low-rank memory states to the original Metis models to characterize the trends of their performance degradation, where the full memory-state dimension is 1024.
The results are presented in \textbf{Figure~\ref{fig:lowrank_4b_overall}}, which shows a clear trend as the retained rank increases.
Extremely low ranks, such as 1 and 4, cause substantial performance degradation. This indicates that a few singular directions are insufficient to preserve the semantic information stored in the memory states. Performance improves rapidly when the rank increases to 16 and becomes close to the full-rank model at rank 64.
Further increasing the rank from 64 to 256 brings almost no additional improvement. These results suggest that useful information in the memory states is mainly concentrated in a relatively low-dimensional subspace.

\input{tables/lowrank_4b_per_benchmark}

\textbf{Table~\ref{tab:lowrank_4b_per_benchmark}} further shows the sensitivity to low-rank decomposition in different datasets.
The results show that different benchmarks have different levels of sensitivity to aggressive low-rank compression.
At very small ranks, performance drops clearly on all datasets. The decrease is particularly large on the Metis test set and MemOps (Gold), suggesting that memory operation tasks require sufficient representational capacity to preserve operation-related information.
LoCoMo (Gold) is relatively less sensitive at $k=1$, although its performance fluctuates at very small ranks. This may indicate that some useful conversational information is concentrated in a small number of dominant directions. However, these directions alone are not sufficient to support stable memory utilization.
MemOps (Gold) recovers its full-rank performance at $k=16$ and remains stable at larger ranks. In contrast, the Metis test set continues to improve from $k=16$ to $k=64$. This difference suggests that the two memory operation benchmarks require different levels of memory capacity.
Across all datasets, performance becomes close to the full model at $k=64$.
The overall recovery reaches 99.9\%, while further increasing the rank brings little additional benefit.
These results confirm that the memory states still contain substantial redundancy and that most useful information lies in a relatively low-dimensional subspace.
They also show that the appropriate compression level depends on the task, since different benchmarks require different amounts and types of memory information.

\subsection{Case Studies}
\begin{figure*}[t]
    \centering
\includegraphics[width=\textwidth]{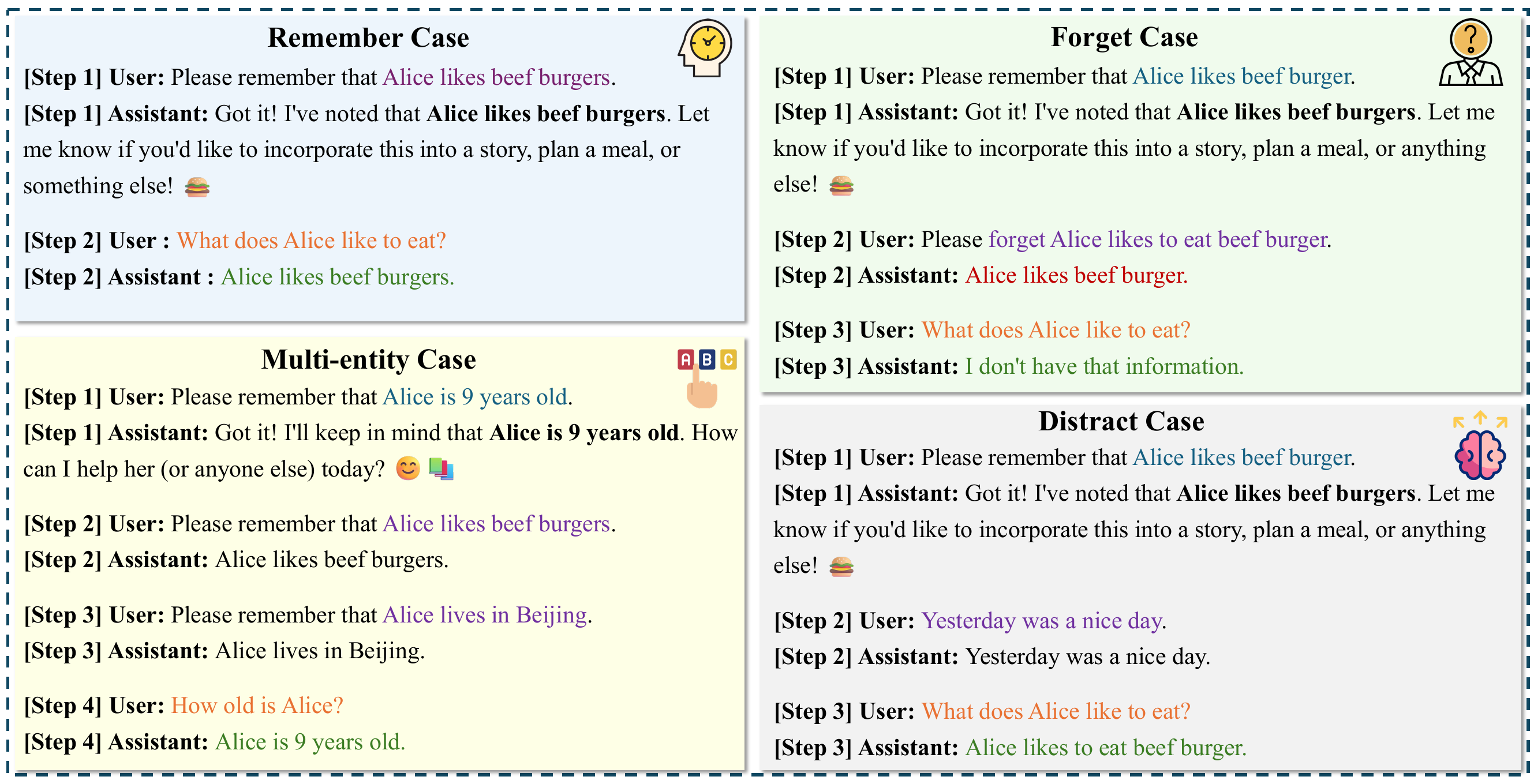}
    \caption{Case studies of Metis on different conversational scenarios.}
    \label{fig:case_studies}
    \vspace{-0.4cm}
\end{figure*}

We conduct qualitative case studies to show the behavior of Metis-4B under different scenarios.
At each turn, the model first generates a response based on the current input and existing memory, and then updates the user input in the memory state.
We present several representative cases in \textbf{Figure~\ref{fig:case_studies}} to qualitatively examine the native memory behaviors of Metis-4B.
Each case starts from an empty native memory state, and the model must use information stored in previous interaction steps.
In the remembering case, Metis correctly stores Alice's food preference and retrieves it in a later query.
In the multi-fact case, the model retains several attributes about Alice and correctly selects her age after other attributes are introduced.
This result suggests that Metis can bind different values to their corresponding attributes and reduce interference among related facts.
The distractor case further shows that an unrelated dialogue turn does not overwrite the stored preference.
Metis can therefore distinguish useful memory from ordinary conversational content.

The forgetting case demonstrates that the native memory state is not append-only.
After receiving a forgetting instruction, Metis no longer provides the removed preference in the subsequent query.
This indicates that the model can modify its latent memory state according to the semantic intent of an instruction.
However, the immediate response to the forgetting instruction still repeats the old fact instead of explicitly confirming its removal.
The final memory state is correct, but the response at the operation step is not fully aligned with the intended memory operation.
This behavior may result from the current step over-emphasizing previous memory states.
Overall, these cases show the effectiveness of Metis, while also revealing room for improvement in its consistency.

\section{Related Work}
\label{sec:related_work}
\subsection{Memory of LLMs and Agents}
In recent years, large foundation models and agents have been widely applied to fields such as personal assistants~\cite{lewis2020retrieval,zhang2026memsim}, deep research~\cite{huang2025deep,zheng2025deepresearcher,du2025deepresearch}, and coding agents~\cite{jiang2026survey,chen2021evaluating,roziere2023code}.
A critical capability of these systems is memory, which stores past information to support future inference~\cite{zhang2025survey}.
Based on their representation forms, memory mechanisms of large foundation models and agents are generally categorized into three types, including textual memory, latent memory, and parametric memory~\cite{zhang2025survey, hu2025memory}.
Textual memory typically represents information as text, relying on RAG for storage and retrieval.
These methods provide information for backbones to support inference by In-Context Learning (ICL)~\cite{dong2024survey}.
For example, MemoryBank~\cite{zhong2024memorybank} proposes a hierarchical storage approach with dual-tower dense retrieval to maintain historical conversations with users.
MemTree~\cite{rezazadeh2025isolated} designs a tree-structured memory mechanism to model the abstraction levels of information, which dynamically updates based on semantic embeddings.
In contrast, latent memory captures memory through intermediate activations of models.
For example, NextMem~\cite{zhang2026nextmem} compresses factual memory into latent representations through an autoregressive autoencoder, while MemGen~\cite{zhang2025memgen} generates latent memory tokens that are interwoven into the reasoning process.
Additionally, parametric memory injects knowledge into internal model parameters.
For example, Locas~\cite{lu2026locas} views the FFN as a soft look-up table. By adding a bypass FFN, it stores test-time information from a key-value perspective. Furthermore, knowledge editing can also be considered a parametric memory method~\cite{zhang2025survey}.
ROME~\cite{meng2022locating} treats the projection matrix as an associative memory and inserts a new factual association through a rank-one update.
Although textual memory remains the most effective approach in industry, latent memory and parametric memory are emerging as promising research directions.

\subsection{Fast Weight Programming}

Recently, FWP has attracted widespread attention.
This paradigm not only uses parameters learned during training (\textit{i.e.,} slow weights), but also maintains dynamic parameters (\textit{i.e.,} fast weights) during inference to capture sequence-dependent information~\cite{ba2016using}.
Existing methods in this line of work generally follow several main directions.
Linear attention replaces the softmax kernel with feature maps to achieve linear complexity and a recurrent state~\cite{katharopoulos2020transformers}.
In addition, it has been shown that linear transformers are secretly fast weight programmers~\cite{schlag2021linear}.
Subsequent works enrich the update rule, such as RetNet~\cite{sun2023retentive} and RWKV~\cite{peng2023rwkv}.
Furthermore, state space models compress a sequence into a fixed-size recurrent state with linear-time computation, such as S4~\cite{gu2021efficiently} and Mamba~\cite{gu2023mamba}, while Mamba-2~\cite{dao2024transformers} further reveals a duality between state space models and attention.
TTT also treats the recurrent state as fast weights that are optimized by self-supervised gradient descent during inference, such as the TTT layer~\cite{sun2024learning} and Titans~\cite{behrouz2026titans}.

\subsection{Memory-Augmented Neural Networks}
MANNs introduce explicit memory modules to improve a model's ability to store and retrieve task-specific information during inference.
Early work, such as Memory Networks~\cite{weston2014memory} and Neural Turing Machines~\cite{graves2014neural}, augments neural controllers with external memory and learns differentiable read and write operations over memory slots. 
These methods show that neural models can use non-parametric memory to support associative recall, algorithmic reasoning, and few-shot adaptation.
However, their memory is usually maintained as a separate storage module, and the memory procedures are often designed independently from the backbone computation.
Recent models also maintain dynamic states during inference, such as recurrent memory~\cite{bulatov2022recurrent}.
Unlike static model parameters learned during training, these dynamic states are updated according to the current input sequence and capture information that changes over time.
Our work follows this general direction, but focuses on integrating memory storage and utilization directly into the model computation, so that the model can maintain sequence-dependent information more natively.

\section{Conclusion}
\label{sec:conclusion}
In this paper, we introduce memory foundation models and provide formal definitions of native memory based on the memory state and memory procedures.
Based on this formulation, we propose Metis, the first prototype of memory foundation models.
We introduce Metis blocks composed of local memory blocks and hyper memory blocks, enabling the model to maintain compact dense memory states across interaction steps and to update them according to the current input and generated response.
We further construct a memory-specific dataset from public benchmarks and design a mid-training framework with memory reconstruction, memory operation, and regularization objectives.
Our experiments verify the effectiveness of Metis and analyze its behavior from multiple perspectives.

Despite these promising results, Metis is still an early step toward memory foundation models.
Since the current native memory state compresses information into fixed-size latent parameters, performance may degrade in extremely long-term scenarios, and semantically similar facts may sometimes be confused in the latent space.
Therefore, native memory still cannot be viewed as a complete replacement for external memory.
Instead, we believe it opens a complementary direction for building future foundation models with more efficient, optimizable, and deeply integrated memory capabilities.
Future work may further improve memory capacity, controllability, and interpretability, explore hybrid systems that combine native and external memory, and scale native memory training to broader domains and longer interactions.

\bibliographystyle{unsrt}
\bibliography{reference}

\newpage

\appendix

\section{Roadmap for Memory Foundation Models}
\label{sec:future_memory_foundation_models}

\begin{figure}[htp]
    \centering
    \includegraphics[width=\textwidth]
    {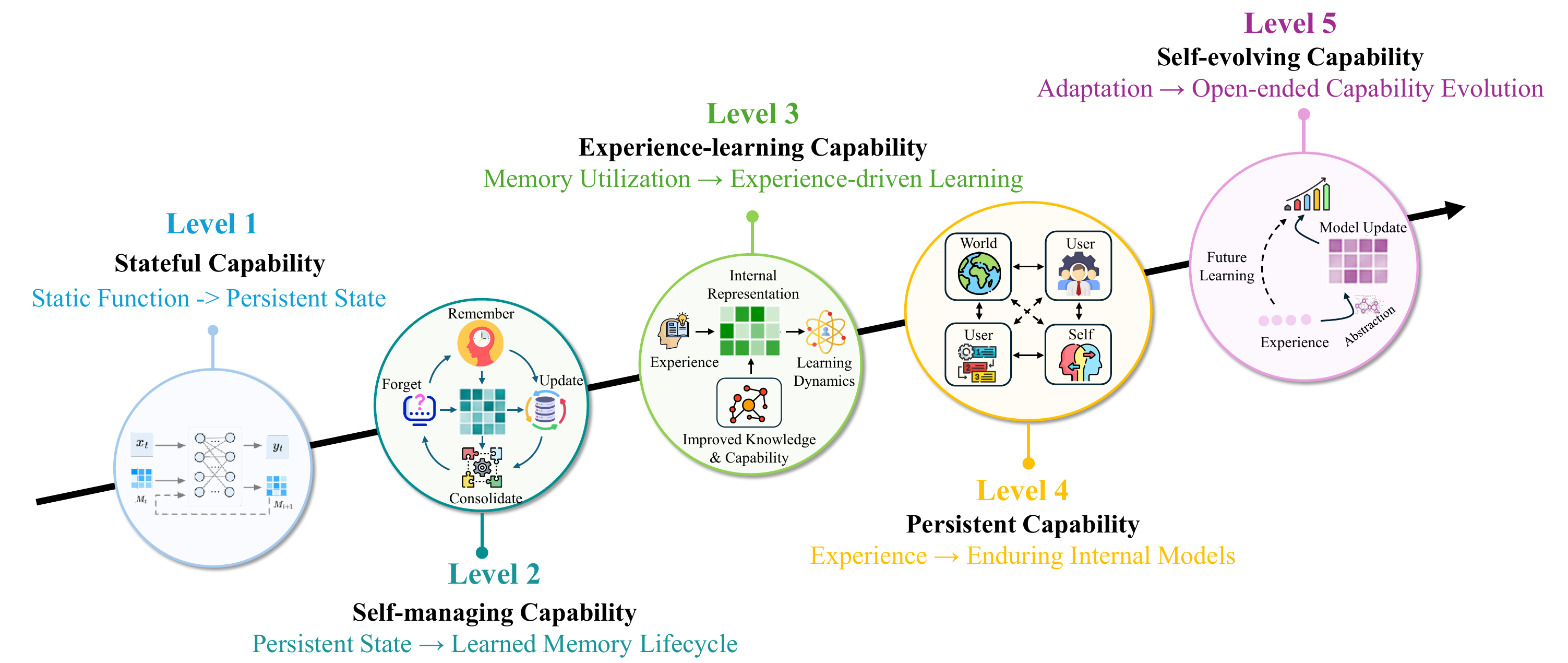}
    \caption{
    \textbf{Roadmap for memory foundation models.}
    Native memory aims to progressively transform foundation models from stateless predictors into stateful learners and, ultimately, into models capable of continual self-evolution.
    The five capabilities represent increasingly deep changes to the model's computation, memory organization, learning process, cognitive representations, and capability formation.
    Their development is supported by advances in memory architecture, learning objectives, scalable training, interpretability and control, and long-horizon evaluation.
    }
    \label{fig:memory_native_stages}
\end{figure}

Metis represents an initial exploration of memory foundation models.
It transforms memory from an external information-management module into a persistent internal state, integrating memory storage and utilization directly into the forward computation of the model.
The significance of native memory, however, extends beyond improving information retention capability.
In the longer term, it may reshape the computational paradigm, learning process, cognitive structure, and capability development of foundation models.
As illustrated in \textbf{Figure~\ref{fig:memory_native_stages}}, we envision five progressive levels of capabilities in the development of memory foundation models: \textbf{stateful capability}, \textbf{self-managing capability}, \textbf{experience-learning capability}, \textbf{persistent cognitive capability}, and \textbf{self-evolving capability}.
These capabilities characterize how memory may become progressively integrated into the foundation model itself, progressing from persistent state to autonomous memory organization, experience-driven learning, persistent cognition, and continual capability evolution.

\textbf{Level I: Stateful Capability.} Most existing foundation models remain fundamentally stateless conditional predictors.
At each inference step, their outputs are determined by fixed model parameters and the current context, while continuity across interactions is primarily maintained by repeatedly supplying historical information as external context.
Therefore, the first level is a transition from a \textbf{static function to a persistent state}.
A memory foundation model maintains a dynamic internal state across inference steps.
Its output is jointly determined by the current input and its previous state, while each interaction updates the state used in subsequent computation:
\begin{equation*}
    (Y_t, \mathbf{M}_{t+1}) = f_{\Phi}(X_t, \mathbf{M}_t),
    \label{eq:stateful_foundation_model}
\end{equation*}
where $\Phi$ denotes the static model parameters and $\mathbf{M}_t$ denotes the native memory state at step $t$.
Unlike textual context supplied from outside the model, $\mathbf{M}_t$ is directly coupled with the model's internal computation and evolves continuously during interaction.
This transition changes the basic computational unit of a foundation model. The model is no longer merely a static mapping from inputs to outputs, but rather a stateful computational system that evolves over time.

\textbf{Level II: Self-Managing Capability.}
Possessing a persistent state does not by itself constitute a complete memory capability.
Information observed in real environments differs in importance, validity, abstraction level, time scale, and security requirements.
A model must therefore learn how to organize and maintain its internal state autonomously.
The second level moves from \textbf{persistent state to a learned memory lifecycle}.
The model should determine what to remember, update, consolidate, and forget according to the semantics of incoming information, its expected future utility, and applicable privacy and safety constraints.
At this level, memory operations are no longer implemented primarily through external rules or discrete workflows.
Instead, both the memory state and the procedures that transform it become native, trainable components of the foundation model.
This enables memory to be selected, organized, and evolved within the model's continuous computational space.

\textbf{Level III: Experience-Learning Capability.}
Once a model can autonomously maintain memory, the role of memory can expand from information support to capability adaptation.
The model should not only remember what happened, but also change as a consequence of what it has experienced.
The third level marks a transition from \textbf{memory utilization to experience-driven learning}.
Interaction histories become reward signals that the model can use to refine representations, knowledge, and behavioral regularities.
Experiences involving success, failure, feedback, or environmental change can be transformed into reusable internal capabilities rather than remaining isolated records.
This direction may gradually connect pre-training, in-context learning, test-time adaptation, and continual learning within a unified framework.
Instead of remaining completely fixed after training, a foundation model could continuously adapt to new users, tasks, and environments while preserving previously acquired capabilities.

\textbf{Level IV: Persistent Cognitive Capability.}
Experience learning explains how a model may adapt through interaction, but more advanced intelligence requires the formation of structured, persistent, and continually updated internal cognition.
The next level moves from \textbf{accumulated experience to enduring internal models}.
A foundation model should maintain evolving representations of the world, users, tasks, time, and itself. These representations should capture temporal changes, causal dependencies, uncertainty, and conflicts between new evidence and existing beliefs. When the environment changes, the model should be able to revise its internal representations while preserving global consistency.
At this level, memory no longer consists of disconnected pieces of historical information.
It becomes the substrate through which the model maintains cognitive continuity over extended periods.
Planning, personalization, and complex decision-making may emerge as downstream expressions of this capability, but the defining transformation occurs within the foundation model's internal cognitive representations rather than in an external agent workflow.

\textbf{Level V: Self-Evolving Capability.}
The long-term objective of memory foundation models is to convert accumulated experience into the continual development of the model's own capabilities.
A model should not only adapt its current state, but also reflect on, abstract, and reorganize past experience to discover new knowledge structures and learning strategies.
The final level represents a transition from \textbf{local adaptation to open-ended capability evolution}.
The model identifies experiences with long-term value, abstracts transferable knowledge from specific interactions, and incorporates the resulting insights into future learning.
Then, the exploration and exploitation will constitute a continual feedback loop between memory foundation models and the environment.

\textbf{Outlook.}
Memory foundation models should not be viewed as conventional foundation models equipped with a stronger storage module. 
They point toward a more fundamental change in the paradigm of the foundation model.
Under this view, memory may become a foundational mechanism connecting computation, learning, cognition, and continual self-evolution.

\begin{keyinsight}[A possible paradigm shift]
Foundation models may evolve from \textbf{stateless predictors} into \textbf{stateful learners}, and ultimately into learning systems that autonomously organize memory, maintain persistent cognition, and develop new capabilities through accumulated experience.
\end{keyinsight}

\section{Extensive Experiment Results}

\input{appendix/appendix_memory_operation_results}

\section{Further Analysis of Update Designs}

\input{appendix/lu_gdu_scaling}

\input{appendix/lu_downward_ablation}

\input{appendix/backbone_transferability}

\input{appendix/appendix_exp_impl_details}

\section{Prompts}
\input{appendix/experiment_prompts}

\section{Efficiency}
\input{appendix/Efficiency}

\end{document}

%% file: tables/overall_memop.tex
\begingroup
\newcommand{\refgray}[1]{\textcolor[gray]{0.45}{#1}}

\begin{table}[t]
  \centering
  \renewcommand{\arraystretch}{1.1}
  \caption{The overall performance on memory operation tasks. Full-context and partial-context results are shown in gray to visually distinguish context-access settings from the no-context comparison. Within the No Context setting, the best and second-best scores are \textbf{bolded} and \underline{underlined}, respectively. Avg. represents the micro-average performance.}
  \vspace{-0.1cm}
  \resizebox{1.0\textwidth}{!}
  {
    \begin{tabular}{cccccccccccc}
      \hline
      \hline
      \multirow{2}[4]{*}{\textbf{Type}}
      & \multirow{2}[4]{*}{\textbf{Method}}
      & \multicolumn{5}{c}{\textbf{MemOps (Gold)}}
      & \multicolumn{5}{c}{\textbf{Metis Test Set}}
      \bigstrut\\

      \cline{3-12}
      &
      & \textbf{Remember}
      & \textbf{Update}
      & \textbf{Forget}
      & \textbf{Reflect}
      & \textbf{Avg.}
      & \textbf{Remember}
      & \textbf{Update}
      & \textbf{Forget}
      & \textbf{Reflect}
      & \textbf{Avg.}
      \bigstrut\\
      \hline

      \multirow{3}[2]{*}{\refgray{Full Context}}
      & \refgray{Qwen3.5-4B}
      & \refgray{84.97}
      & \refgray{86.34}
      & \refgray{81.36}
      & \refgray{85.17}
      & \refgray{84.56}
      & \refgray{80.07}
      & \refgray{70.31}
      & \refgray{70.42}
      & \refgray{83.13}
      & \refgray{75.18}
      \bigstrut[t]\\

      & \refgray{Qwen3.5-9B}
      & \refgray{88.54}
      & \refgray{88.43}
      & \refgray{82.73}
      & \refgray{86.90}
      & \refgray{86.86}
      & \refgray{78.18}
      & \refgray{69.48}
      & \refgray{67.50}
      & \refgray{84.38}
      & \refgray{73.89}\\

      & \refgray{Qwen3.5-27B}
      & \refgray{91.37}
      & \refgray{90.74}
      & \refgray{84.32}
      & \refgray{84.48}
      & \refgray{87.90}
      & \refgray{81.01}
      & \refgray{73.44}
      & \refgray{75.83}
      & \refgray{88.75}
      & \refgray{78.87}
      \bigstrut[b]\\
      \hline

      \multirow{3}[2]{*}{\refgray{Partial Context}}
      & \refgray{Qwen3.5-4B}
      & \refgray{38.84}
      & \refgray{33.56}
      & \refgray{24.55}
      & \refgray{21.90}
      & \refgray{30.18}
      & \refgray{70.05}
      & \refgray{63.12}
      & \refgray{59.90}
      & \refgray{67.81}
      & \refgray{64.82}
      \bigstrut[t]\\

      & \refgray{Qwen3.5-9B}
      & \refgray{30.51}
      & \refgray{26.62}
      & \refgray{20.23}
      & \refgray{11.55}
      & \refgray{22.41}
      & \refgray{70.40}
      & \refgray{55.10}
      & \refgray{55.00}
      & \refgray{64.69}
      & \refgray{60.68}\\

      & \refgray{Qwen3.5-27B}
      & \refgray{37.05}
      & \refgray{35.88}
      & \refgray{22.05}
      & \refgray{16.21}
      & \refgray{28.01}
      & \refgray{70.28}
      & \refgray{63.33}
      & \refgray{66.46}
      & \refgray{60.47}
      & \refgray{65.40}
      \bigstrut[b]\\

      \hline

      \multirow{10}[4]{*}{No Context}
      & Qwen3.5-4B
      & 4.17
      & 0.00
      & 1.59
      & 0.00
      & 1.65
      & 12.03
      & 0.00
      & 49.58
      & 0.00
      & 16.96
      \bigstrut[t]\\

      & Qwen3.5-9B
      & 4.17
      & 1.85
      & 0.91
      & 0.00
      & 1.88
      & 11.79
      & 5.42
      & \underline{49.90}
      & 0.63
      & 18.64\\

      & Qwen3.5-27B
      & 3.57
      & 0.93
      & 0.91
      & 0.69
      & 1.69
      & 10.73
      & 2.08
      & 47.50
      & 1.25
      & 16.87\\

      & Temp-LoRA-4B
      & 15.33
      & 10.19
      & 2.95
      & 4.83
      & 8.85
      & 15.80
      & 27.71
      & 15.21
      & 17.66
      & 19.34\\

      & Temp-LoRA-9B
      & 23.81
      & 13.43
      & 5.00
      & 8.10
      & 13.51
      & 20.05
      & 17.71
      & 20.21
      & 20.47
      & 19.51\\

      & Temp-LoRA-27B
      & 20.68
      & 6.48
      & 2.50
      & 4.83
      & 9.70
      & 25.94
      & 18.65
      & 25.21
      & 26.87
      & 23.86\\

      & $\delta$-Mem
      & 7.44
      & 6.02
      & 1.82
      & 1.55
      & 4.38
      & 13.92
      & 21.77
      & 12.40
      & 10.31
      & 15.03
      \bigstrut[b]\\

      \cline{2-12}

      & Metis-4B
      & 19.35
      & \underline{27.55}
      & 7.27
      & \underline{16.90}
      & 17.84
      & 52.24
      & \underline{63.85}
      & 31.25
      & 90.16
      & 56.72
      \bigstrut[t]\\

      & Metis-9B
      & \underline{25.89}
      & 23.61
      & \textbf{11.59}
      & 15.52
      & \underline{19.63}
      & \underline{58.14}
      & 63.33
      & 30.42
      & \underline{90.78}
      & \underline{57.92}\\

      & Metis-27B
      & \textbf{28.27}
      & \textbf{31.02}
      & \underline{10.91}
      & \textbf{26.55}
      & \textbf{24.76}
      & \textbf{61.08}
      & \textbf{68.13}
      & \textbf{77.50}
      & \textbf{93.44}
      & \textbf{73.77}
      \bigstrut[b]\\

      \hline
      \hline
    \end{tabular}
  }
  \label{tab:results_memoryops}
  \vspace{-0.4cm}
\end{table}

\endgroup

%% file: tables/overall_memqa.tex
\begingroup
\newcommand{\refgray}[1]{\textcolor[gray]{0.45}{#1}}

\begin{table}[t]
  \centering
  \renewcommand{\arraystretch}{1.1}
  \caption{The overall performance on memory-based QA tasks. Full-context and partial-context results are shown in gray to visually distinguish context-access settings from the no-context comparison. Within the No Context setting, the best and second-best scores are \textbf{bolded} and \underline{underlined}, respectively.
  Single and Multi indicate Single-hop Retrieval and Multi-hop Retrieval settings in LoCoMo (Gold), respectively.
  Temporal and Open refer to the temporal reasoning setting and open domain knowledge settings in LoCoMo (Gold), respectively.
  A dash indicates that the corresponding result is not applicable. Avg. represents the micro-average performance.}
  \vspace{-0.2cm}
  \resizebox{1.0\textwidth}{!}
  {
    \begin{tabular}{cccccccccccc}
      \hline
      \hline

      \multirow{2}[4]{*}{\textbf{Type}}
      & \multirow{2}[4]{*}{\textbf{Method}}
      & \multicolumn{5}{c}{\textbf{LoCoMo (Gold)}}
      & \multicolumn{5}{c}{\textbf{NextMem}}
      \bigstrut\\

      \cline{3-12}
      &
      & \textbf{Single}
      & \textbf{Multi}
      & \textbf{Temporal}
      & \textbf{Open}
      & \textbf{Avg.}
      & \textbf{SQuAD}
      & \textbf{HotpotQA}
      & \textbf{LongMemEval}
      & \textbf{LoCoMo}
      & \textbf{Avg.}
      \bigstrut\\

      \hline

      \multirow{3}[2]{*}{\refgray{Full Context}}
      & \refgray{Qwen3.5-4B}
      & \refgray{85.12}
      & \refgray{65.92}
      & \refgray{15.78}
      & \refgray{23.88}
      & \refgray{63.52}
      & \refgray{91.00}
      & \refgray{88.28}
      & \refgray{45.21}
      & \refgray{61.00}
      & \refgray{77.15}
      \bigstrut[t]\\

      & \refgray{Qwen3.5-9B}
      & \refgray{84.43}
      & \refgray{64.93}
      & \refgray{16.64}
      & \refgray{21.35}
      & \refgray{63.00}
      & \refgray{91.00}
      & \refgray{87.61}
      & \refgray{45.79}
      & \refgray{60.65}
      & \refgray{77.05}\\

      & \refgray{Qwen3.5-27B}
      & \refgray{85.83}
      & \refgray{69.96}
      & \refgray{15.55}
      & \refgray{31.18}
      & \refgray{65.03}
      & \refgray{91.80}
      & \refgray{89.18}
      & \refgray{48.43}
      & \refgray{64.47}
      & \refgray{78.80}
      \bigstrut[b]\\

      \hline

      \multirow{3}[2]{*}{\refgray{Partial Context}}
      & \refgray{Qwen3.5-4B}
      & \refgray{36.28}
      & \refgray{10.79}
      & \refgray{6.17}
      & \refgray{5.62}
      & \refgray{23.54}
      & \refgray{-}
      & \refgray{-}
      & \refgray{-}
      & \refgray{-}
      & \refgray{-}
      \bigstrut[t]\\

      & \refgray{Qwen3.5-9B}
      & \refgray{34.05}
      & \refgray{8.27}
      & \refgray{7.50}
      & \refgray{2.81}
      & \refgray{21.97}
      & \refgray{-}
      & \refgray{-}
      & \refgray{-}
      & \refgray{-}
      & \refgray{-}\\

      & \refgray{Qwen3.5-27B}
      & \refgray{35.63}
      & \refgray{10.07}
      & \refgray{7.73}
      & \refgray{1.97}
      & \refgray{23.17}
      & \refgray{-}
      & \refgray{-}
      & \refgray{-}
      & \refgray{-}
      & \refgray{-}
      \bigstrut[b]\\

      \hline

      \multirow{10}[4]{*}{No Context}
      & Qwen3.5-4B
      & 0.00
      & 0.36
      & 0.00
      & 1.97
      & 0.18
      & 11.24
      & 25.22
      & 2.86
      & 0.48
      & 11.86
      \bigstrut[t]\\

      & Qwen3.5-9B
      & 0.00
      & 0.36
      & 0.00
      & 0.00
      & 0.07
      & 14.92
      & 34.98
      & 2.86
      & 0.48
      & 15.93\\

      & Qwen3.5-27B
      & 0.00
      & 0.36
      & 0.00
      & 0.00
      & 0.07
      & 16.73
      & 38.90
      & 3.14
      & 0.48
      & 17.75\\

      & Temp-LoRA-4B
      & 10.92
      & 11.24
      & 1.80
      & 26.69
      & 9.99
      & 26.19
      & 38.62
      & 9.71
      & 11.12
      & 24.20\\

      & Temp-LoRA-9B
      & 13.33
      & 13.31
      & 2.42
      & 25.00
      & 11.72
      & 29.52
      & 45.07
      & 12.93
      & 12.80
      & 28.12\\

      & Temp-LoRA-27B
      & 4.29
      & 5.49
      & 1.25
      & 10.67
      & 4.24
      & \underline{37.68}
      & 51.46
      & 6.57
      & 5.86
      & 30.97\\

      & $\delta$-Mem
      & 12.86
      & 10.16
      & 3.28
      & 20.22
      & 10.79
      & 20.74
      & 33.02
      & 9.79
      & 10.29
      & 20.42
      \bigstrut[b]\\

      \cline{2-12}

      & Metis-4B
      & \underline{18.90}
      & 15.29
      & \underline{7.03}
      & \underline{28.37}
      & 16.31
      & 29.62
      & 58.13
      & \underline{39.36}
      & 50.48
      & 41.69
      \bigstrut[t]\\

      & Metis-9B
      & 18.87
      & \underline{18.53}
      & \underline{7.03}
      & 27.25
      & \underline{16.81}
      & 33.06
      & \underline{63.45}
      & 33.36
      & \underline{51.56}
      & \underline{43.39}\\

      & Metis-27B
      & \textbf{31.01}
      & \textbf{27.97}
      & \textbf{13.83}
      & \textbf{28.93}
      & \textbf{26.74}
      & \textbf{43.42}
      & \textbf{66.54}
      & \textbf{39.71}
      & \textbf{60.41}
      & \textbf{50.82}
      \bigstrut[b]\\

      \hline
      \hline
    \end{tabular}
  }
  \label{tab:results_memory_based_QA}
  \vspace{-0.2cm}
\end{table}

\endgroup

%% file: tables/ablation_4b_data_structure.tex
\begin{table}[t]
  \centering
  \renewcommand{\arraystretch}{1.1}
  \caption{Results of the ablation studies on Metis-4B. We use Avg. to report the macro-average performance in different groups, and utilize $\Delta$Avg. to represent the relative performance gap compared with the full model.}
  \vspace{-0.2cm}
  \resizebox{1.0\textwidth}{!}
		{    \begin{tabular}{cccccccccccc}
    \hline
    \hline
    \multirow{2}[4]{*}{\textbf{Type}} & \multirow{2}[4]{*}{\textbf{Model}} & \multicolumn{4}{c}{\textbf{Memory Operation Task}} & \multicolumn{4}{c}{\textbf{Memory-based QA Task}} & \multicolumn{2}{c}{\textbf{Overall}} \bigstrut\\
\cline{3-12}          &       & \textbf{MemOps (Gold)} & \textbf{Metis Test Set} & \textbf{Avg.} & \boldmath{}\textbf{$\Delta$Avg.}\unboldmath{} & \textbf{LoCoMo (Gold)} & \textbf{NextMem} & \textbf{Avg.} & \boldmath{}\textbf{$\Delta$Avg.}\unboldmath{} & \textbf{Avg.} & \boldmath{}\textbf{$\Delta$Avg.}\unboldmath{} \bigstrut\\
    \hline
    Full Model & Metis & 17.84  & 56.72  & 37.28  & -     & 16.31  & 41.69  & 29.00  & -     & 33.14  & - \bigstrut\\
    \hline
    \multirow{2}[2]{*}{\shortstack{Data\\Ablation}} & w/o MS & 14.64  & 51.53  & 33.08  & -11.26\%  & 14.78  & 37.79  & 26.29  & -9.36\%  & 29.68  & -10.43\%  \bigstrut[t]\\
          & w/o MS+MP
& 14.45 & 42.17 & 28.31 & -24.07\%
& 14.18 & 36.17 & 25.17 & -13.20\%
& 26.74 & -19.31\% \bigstrut[b]\\
    \hline
    \multirow{4}[2]{*}{\shortstack{Structure\\Ablation}} & w/o GDU & 18.50  & 58.54  & 38.52  & 3.32\%  & 11.97  & 42.78  & 27.37  & -5.60\%  & 32.95  & -0.58\%  \bigstrut[t]\\
          & w/o SA & 3.67  & 19.72  & 11.70  & -68.63\%  & 9.84  & 18.49  & 14.16  & -51.16\%  & 12.93  & -60.98\%  \\
          & w/o OQ & 13.89  & 53.93  & 33.91  & -9.04\%  & 11.46  & 37.07  & 24.26  & -16.33\%  & 29.09  & -12.23\%  \\
          & w/o QKN & 9.32  & 48.74  & 29.03  & -22.13\%  & 10.00  & 26.80  & 18.40  & -36.55\%  & 23.72  & -28.44\%  \bigstrut[b]\\
    \hline
    \hline
    \end{tabular}}
  \label{tab:ablation_4b_data_structure}%
\end{table}%

%% file: tables/ood_no_context.tex
\begin{table}[t]
    \centering
    \small
    \vspace{-0.2cm}
    \renewcommand{\arraystretch}{1.18}
    \caption{Results on OOD memory benchmarks. The best and unique second-best scores are \textbf{bolded} and \underline{underlined}, respectively. The average score is calculated according to the official category counts.}
    \vspace{-0.2cm}
    \label{tab:ood-no-context}
    \resizebox{\textwidth}{!}{%
    \begin{tabular}{lcccc@{\hspace{1.2em}}ccccccc}        \hline
        \hline
        \multirow{2}{*}{\textbf{Method}}
        & \multicolumn{4}{c}{\textbf{ATM-Bench (Gold)}}
        & \multicolumn{7}{c}{\textbf{MemDaily (Gold)}} \\
        \cmidrule(lr){2-5}\cmidrule(lr){6-12}
        & \textbf{List} & \textbf{Number} & \textbf{Open} & \textbf{Avg.}
        & \textbf{Aggreg.} & \textbf{Comp.} & \textbf{Cond.} & \textbf{Noisy} & \textbf{Post-proc.} & \textbf{Simple} & \textbf{Avg.} \\
        \hline
        $\delta$-Mem
        & 0.00 & 1.94 & 3.11 & 2.27
        & 29.44 & 21.14 & 44.40 & 38.40 & 59.00 & 45.58 & 39.84 \\
        Temp-LoRA-4B
        & 0.00 & 0.00 & 5.06 & 2.57
        & 30.74 & 30.49 & 50.80 & 43.60 & 60.60 & 52.01 & 44.92 \\
        Temp-LoRA-9B
        & 0.00 & 0.00 & 5.64 & 2.86
        & \underline{45.24} & 31.91 & \underline{58.40} & 46.40 & 66.80 & 59.04 & 51.42 \\
        Temp-LoRA-27B
        & 0.00 & 0.00 & 5.06 & 2.57
        & \textbf{61.90} & 40.24 & \textbf{61.00} & \underline{50.60} & \underline{74.20} & \textbf{68.67} & \textbf{59.45} \\
        \cline{1-12}
        Metis-4B
        & \textbf{1.08} & 14.17 & 9.92 & 10.22
        & 45.02 & \underline{54.67} & 51.00 & 44.60 & 64.80 & 54.62 & 52.54 \\
        Metis-9B
        & 0.00 & \underline{24.72} & \textbf{15.18} & \underline{16.49}
        & 30.30 & 34.35 & 53.00 & 43.60 & 66.80 & 54.22 & 47.29 \\
        Metis-27B
        & 0.00 & \textbf{31.39} & \underline{14.59} & \textbf{18.56}
        & 40.69 & \textbf{66.06} & 56.60 & \textbf{52.80} & \textbf{75.40} & \underline{61.45} & \underline{59.04} \\
        \hline
        \hline
    \end{tabular}%
    }
    \vspace{-0.4cm}
\end{table}

%% file: tables/source_exclusion.tex
\begingroup
\newcommand{\sourceexclneg}[1]{\textcolor[gray]{0.45}{#1}}
\newcommand{\sourceexclpos}[1]{\raisebox{0.02ex}{\scalebox{0.87}{\textbf{#1}}}}

\begin{table*}[t]
  \centering
  \small
  \setlength{\tabcolsep}{3.0pt}
  \renewcommand{\arraystretch}{1.1}
  \caption{Results of the source-exclusion study. Parentheses give percentage-point differences from the Metis results (bold: positive; gray: negative). 6-Bench Avg is the unweighted mean of LoCoMo, NextMem Avg, Metis Test, MemOps, ATM, and MemDaily.}
  \label{tab:source-exclusion}
  \vspace{-0.2cm}
  \resizebox{1.0\textwidth}{!}
  {
    \begin{tabular}{lccccccccc}
      \hline
      \hline
      \multirow{2}[4]{*}{\textbf{Model}}
      & \multirow{2}[4]{*}{\textbf{LoCoMo}}
      & \multicolumn{3}{c}{\textbf{NextMem}}
      & \multirow{2}[4]{*}{\textbf{Metis Test}}
      & \multirow{2}[4]{*}{\textbf{MemOps}}
      & \multirow{2}[4]{*}{\textbf{ATM}}
      & \multirow{2}[4]{*}{\textbf{MemDaily}}
      & \multirow{2}[4]{*}{\textbf{6-Bench Avg}}
      \bigstrut\\
      \cline{3-5}
      &
      & \textbf{LongMemEval}
      & \textbf{LoCoMo}
      & \textbf{Avg.}
      &
      &
      &
      &
      \bigstrut\\
      \hline
      Metis-4B
      & 13.18 \sourceexclneg{(-3.13)}
      & 29.71 \sourceexclneg{(-9.65)}
      & 42.46 \sourceexclneg{(-8.02)}
      & 37.30 \sourceexclneg{(-4.39)}
      & 63.26 \sourceexclpos{(+6.54)}
      & 16.24 \sourceexclneg{(-1.60)}
      & 12.54 \sourceexclpos{(+2.32)}
      & 42.85 \sourceexclneg{(-9.69)}
      & 30.90 \sourceexclneg{(-1.66)}
      \bigstrut[t]\\
      Metis-9B
      & 15.72 \sourceexclneg{(-1.09)}
      & 32.00 \sourceexclneg{(-1.36)}
      & 43.06 \sourceexclneg{(-8.50)}
      & 41.54 \sourceexclneg{(-1.85)}
      & 66.93 \sourceexclpos{(+9.01)}
      & 19.30 \sourceexclneg{(-0.33)}
      & 11.65 \sourceexclneg{(-4.84)}
      & 41.84 \sourceexclneg{(-5.45)}
      & 32.83 \sourceexclneg{(-0.76)}
      \\
      Metis-27B
      & 21.15 \sourceexclneg{(-5.59)}
      & 44.21 \sourceexclpos{(+4.50)}
      & 52.03 \sourceexclneg{(-8.38)}
      & 53.31 \sourceexclpos{(+2.49)}
      & 67.66 \sourceexclneg{(-6.11)}
      & 30.89 \sourceexclpos{(+6.13)}
      & 12.44 \sourceexclneg{(-6.12)}
      & 52.10 \sourceexclneg{(-6.94)}
      & 39.59 \sourceexclneg{(-2.69)}
      \bigstrut[b]\\
      \hline
      \hline
    \end{tabular}
  }
\end{table*}

\endgroup

%% file: tables/general_capability_history.tex
\begin{table*}[t]
  \centering
  \small
  \renewcommand{\arraystretch}{1.0}
  \caption{Results of general capability tasks. The gap is calculated as the performance of Metis-4B minus that of Qwen3.5-4B.
  }
  \vspace{-0.2cm}
  \resizebox{0.7\textwidth}{!}
  {\begin{tabular}{ccccccc}
    \hline
    \hline
    \multirow{2}[4]{*}{\textbf{Benchmark}} & \multicolumn{3}{c}{\textbf{Initial Stage}} & \multicolumn{3}{c}{\textbf{Active Stage}} \bigstrut\\
    \cline{2-7}
     & \textbf{Qwen3.5-4B} & \textbf{Metis-4B} & \textbf{Gap} & \textbf{Qwen3.5-4B} & \textbf{Metis-4B} & \textbf{Gap} \bigstrut\\
    \hline
    MMLU-Pro & 46.00 & 45.20 & $-0.80$ & 46.00 & 40.90 & $-5.10$ \bigstrut\\
    IFEval & 79.30 & 79.85 & $+0.55$ & 76.71 & 54.53 & $-22.18$ \bigstrut\\
    GSM8K & 83.09 & 82.03 & $-1.06$ & 84.53 & 78.92 & $-5.61$ \bigstrut\\
    MMMLU & 61.30 & 60.50 & $-0.80$ & 59.90 & 56.60 & $-3.30$ \bigstrut\\
    \hline
    \hline
  \end{tabular}}
  \label{tab:general_capability_history}
  \vspace{-0.4cm}
\end{table*}

%% file: tables/lowrank_4b_per_benchmark.tex
\begin{table*}[t]
\centering
\renewcommand{\arraystretch}{1.15}
\providecommand{\lrcell}[2]{#1\,{\color{black!60}(#2\%)}}
\caption{Results of low-rank decomposition across different datasets.
Values in parentheses show recovery relative to Full, and the values above 100\% are treated as judge variation rather than improvements.}
\vspace{-0.2cm}
\label{tab:lowrank_4b_per_benchmark}
\resizebox{\textwidth}{!}{%
\begin{tabular}{@{}c*{7}{c}@{}}
\hline
\hline
\textbf{Dataset} & \textbf{$k=1$} & \textbf{$k=4$} & \textbf{$k=16$} & \textbf{$k=64$} & \textbf{$k=128$} & \textbf{$k=256$} & Full \\
\hline
LoCoMo (Gold) & \lrcell{11.31}{69.4} & \lrcell{10.81}{66.3} & \lrcell{14.21}{87.1} & \lrcell{16.00}{98.1} & \lrcell{16.36}{100.3} & \lrcell{16.14}{99.0} & \lrcell{16.31}{100.0} \\
NextMem & \lrcell{22.80}{54.7} & \lrcell{27.83}{66.7} & \lrcell{38.69}{92.8} & \lrcell{41.43}{99.4} & \lrcell{41.78}{100.2} & \lrcell{41.71}{100.0} & \lrcell{41.69}{100.0} \\
Metis Test & \lrcell{17.99}{31.7} & \lrcell{41.26}{72.7} & \lrcell{53.64}{94.6} & \lrcell{56.63}{99.8} & \lrcell{56.75}{100.1} & \lrcell{56.04}{98.8} & \lrcell{56.72}{100.0} \\
MemOps (Gold) & \lrcell{5.60}{31.4} & \lrcell{11.49}{64.4} & \lrcell{18.36}{102.9} & \lrcell{18.36}{102.9} & \lrcell{18.17}{101.8} & \lrcell{18.79}{105.3} & \lrcell{17.84}{100.0} \\
\hline
\textbf{Overall} & \lrcell{14.43}{43.5} & \lrcell{22.84}{68.9} & \lrcell{31.22}{94.2} & \lrcell{33.10}{99.9} & \lrcell{33.26}{100.4} & \lrcell{33.17}{100.1} & \lrcell{33.14}{100.0} \\
\hline
\hline
\end{tabular}%
}
\vspace{-0.4cm}
\end{table*}

%% file: appendix/appendix_memory_operation_results.tex
\label{appendix:extensive_results_memops_full}
The results of memory operation tasks on MemOps (Full) are presented in \textbf{Table~\ref{tab:results_memoryops_full}}.
According to the results, access to complete textual evidence remains the strongest setting.
Full-context models achieve consistently high performance across all four operations, with moderate overall gains from increasing the backbone size.
In contrast, performance drops sharply when only partial context is available.
Standard Qwen models obtain near-zero scores without context, confirming that these operations cannot be performed reliably using backbone knowledge alone.
The partial-context results also show that increasing model size cannot compensate for missing historical evidence.

Among no-context methods, Metis-27B achieves the best overall performance.
It obtains the highest scores on updating and reflection, while Metis-9B performs best on forgetting.
In particular, the substantial improvement on reflection suggests that native memory can support higher-level reasoning over stored information rather than only preserving individual facts.
Temp-LoRA remains strongest on remembering, indicating that temporary parameter adaptation is effective for direct information retention.
However, Metis provides a more balanced advantage across memory operations and outperforms Temp-LoRA in the overall average.

Scaling Metis from 4B to 9B produces only moderate improvements, whereas Metis-27B substantially increases the average score.
This demonstrates that a larger backbone can strengthen native memory operations when sufficient model capacity is available.
However, the gains are not uniform across operations.
For example, Metis-27B improves updating and reflection but performs worse than Metis-9B on forgetting.
This suggests that different memory operations may require different mechanisms and may not benefit equally from backbone scaling.
Despite these improvements, a large gap from the full-context setting remains, showing that accurately storing, modifying, and reasoning over complex histories is still challenging.

\begingroup
\newcommand{\refgray}[1]{\textcolor[gray]{0.45}{#1}}

\begin{table}[t]
  \centering
  \renewcommand{\arraystretch}{1.1}
  \caption{The performance on memory operation tasks under MemOps (Full). Full-context and partial-context results are shown in gray to visually distinguish context-access settings from the no-context comparison. Within the No Context setting, the best and second-best scores are \textbf{bolded} and \underline{underlined}, respectively. Average represents the micro-average performance.}
  \resizebox{0.8\textwidth}{!}
  {
    \begin{tabular}{ccccccc}
      \hline
      \hline

      \textbf{Type}
      & \textbf{Method}
      & \textbf{Remember}
      & \textbf{Update}
      & \textbf{Forget}
      & \textbf{Reflect}
      & \textbf{Average}
      \bigstrut\\

      \hline

      \multirow{3}[2]{*}{\refgray{Full Context}}
      & \refgray{Qwen3.5-4B}
      & \refgray{87.05}
      & \refgray{85.42}
      & \refgray{81.36}
      & \refgray{79.66}
      & \refgray{83.52}
      \bigstrut[t]\\

      & \refgray{Qwen3.5-9B}
      & \refgray{89.58}
      & \refgray{89.58}
      & \refgray{77.95}
      & \refgray{84.14}
      & \refgray{85.69}\\

      & \refgray{Qwen3.5-27B}
      & \refgray{91.52}
      & \refgray{92.13}
      & \refgray{81.82}
      & \refgray{82.59}
      & \refgray{87.19}
      \bigstrut[b]\\

      \hline

      \multirow{3}[2]{*}{\refgray{Partial Context}}
      & \refgray{Qwen3.5-4B}
      & \refgray{31.55}
      & \refgray{23.15}
      & \refgray{20.23}
      & \refgray{14.48}
      & \refgray{22.83}
      \bigstrut[t]\\

      & \refgray{Qwen3.5-9B}
      & \refgray{25.30}
      & \refgray{12.96}
      & \refgray{15.23}
      & \refgray{7.24}
      & \refgray{15.77}\\

      & \refgray{Qwen3.5-27B}
      & \refgray{28.42}
      & \refgray{15.28}
      & \refgray{17.50}
      & \refgray{10.17}
      & \refgray{18.50}
      \bigstrut[b]\\

      \hline

      \multirow{10}[4]{*}{No Context}
      & Qwen3.5-4B
      & 3.57
      & 0.00
      & 1.82
      & 0.00
      & 1.51
      \bigstrut[t]\\

      & Qwen3.5-9B
      & 3.57
      & 1.85
      & 0.68
      & 0.00
      & 1.65\\

      & Qwen3.5-27B
      & 4.17
      & 0.93
      & 0.68
      & 0.69
      & 1.84\\

      & Temp-LoRA-4B
      & 14.88
      & 14.35
      & 2.95
      & 6.38
      & 9.98\\

      & Temp-LoRA-9B
      & \underline{19.79}
      & 14.58
      & 4.09
      & 7.76
      & \underline{12.19}\\

      & Temp-LoRA-27B
      & \textbf{20.68}
      & 10.88
      & 2.50
      & 5.69
      & 10.83\\

      & $\delta$-Mem
      & 4.46
      & 5.09
      & 1.36
      & 1.21
      & 3.06
      \bigstrut[b]\\

      \cline{2-7}

      & Metis-4B
      & 10.42
      & 13.66
      & \underline{4.77}
      & \underline{9.66}
      & 9.70
      \bigstrut[t]\\

      & Metis-9B
      & 13.84
      & \underline{14.81}
      & \textbf{7.27}
      & \underline{9.66}
      & 11.53\\

      & Metis-27B
      & 17.26
      & \textbf{17.13}
      & 3.86
      & \textbf{20.52}
      & \textbf{15.35}
      \bigstrut[b]\\

      \hline
      \hline
    \end{tabular}
  }
  \label{tab:results_memoryops_full}
  \vspace{-0.3cm}
\end{table}

\endgroup

%% file: appendix/lu_gdu_scaling.tex
\subsection{LU and GDU across Model Scales}
\label{app:lu_gdu_scaling}

As discussed in \textbf{Section~\ref{subsec:ablation_studies}}, using a linear update (LU) to replace the GDN-based update (GDU) yields competitive performance at the 4B scale.
In this part, we further explore their performance at the 9B and 27B scales. 
During the training phase, the Metis-9B LU did not show sharp fluctuations on the validation curve, so it was trained to 14k steps without early stopping.
The reported Metis-9B GDU result is based on the 8k checkpoint.

Across all three scales, LU scores higher on the Metis test set but lower on LoCoMo (Gold).
This recurring split points to a task-dependent trade-off between direct memory operations and long conversational memory.
The aggregate comparison is driven by the Metis test set at 9B and LoCoMo (Gold) at 27B, while the two updates remain nearly tied at 4B.
We also find that LU has a sharp drop in LoCoMo (Gold), which may indicate that LU is more vulnerable over long conversational trajectories.
In addition, LU consistently performs better on the Metis test set, possibly because its simpler update rule is easier to fit to the memory operations emphasized during training.

\begin{table*}[t]
  \centering
  \small
  \renewcommand{\arraystretch}{1.08}
  \caption{Performance comparison of Metis with LU and GDU across different model scales. Overall is the equal-weight macro-average score of the four benchmarks.}
  \vspace{-0.2cm}
  \resizebox{0.85\textwidth}{!}
  {\begin{tabular}{ccccccc}
    \hline
    \hline
    \textbf{Scale} & \textbf{Update} & \textbf{LoCoMo (Gold)} & \textbf{NextMem} & \textbf{Metis Test Set} & \textbf{MemOps (Gold)} & \textbf{Overall} \\
    \hline
    \multirow{2}[2]{*}{4B}  & GDU & 16.31 & 41.69 & 56.72 & 17.84 & 33.14 \bigstrut[t]\\
                              & LU  & 11.97 & 42.78 & 58.54 & 18.50 & 32.95 \bigstrut[b]\\
    \hline
    \multirow{2}[2]{*}{9B}  & GDU & 16.81 & 43.39 & 57.92 & 19.63 & 34.44 \bigstrut[t]\\
                              & LU  & 15.37 & 40.16 & 68.66 & 18.17 & 35.59 \bigstrut[b]\\
    \hline
    \multirow{2}[2]{*}{27B} & GDU & 26.74 & 50.82 & 73.77 & 24.76 & 44.02 \bigstrut[t]\\
                              & LU  & 14.16 & 52.09 & 75.32 & 24.44 & 41.50 \bigstrut[b]\\
    \hline
    \hline
  \end{tabular}}
  \label{tab:lu_gdu_scaling}%
\end{table*}

%% file: appendix/lu_downward_ablation.tex
\subsection{Ablations from the LU Baseline}
\label{app:lu_downward_ablation}

The major ablation study in \textbf{Section~\ref{tab:ablation_4b_data_structure}} takes the GDU-based Metis-4B as its reference.
In this section, we replace the GDU with LU and repeat the same data and structure ablations, testing whether their effects depend on the update rule.
All variants are trained for 14{,}000 steps.
We take the LU result reported as \underline{w/o GDU} in \textbf{Table~\ref{tab:ablation_4b_data_structure}} as the reference, and follow the main table in reporting relative changes in Overall.

\begin{table*}[t]
  \centering
  \small
  \renewcommand{\arraystretch}{1.08}
  \caption{Data and structure ablations from the LU baseline on Metis-4B. All scores are percentages. Overall is the equal-weight macro-average of the four benchmarks, and $\Delta$Avg. is the relative performance gap from the reported LU reference.}
  \vspace{-0.2cm}
  \resizebox{1.0\textwidth}{!}
  {\begin{tabular}{cccccccc}
    \hline
    \hline
    \textbf{Type} & \textbf{Model} & \textbf{LoCoMo (Gold)} & \textbf{NextMem} & \textbf{Metis Test Set} & \textbf{MemOps (Gold)} & \textbf{Overall} & \boldmath{}\textbf{$\Delta$Avg.}\unboldmath{} \\
    \hline
    Reference & LU full & 11.97 & 42.78 & 58.54 & 18.50 & 32.95 & -- \\
    \hline
    \multirow{2}[2]{*}{\shortstack{Data\\Ablation}} & LU w/o MS & 16.98 & 40.37 & 62.56 & 14.78 & 33.67 & +2.20\% \bigstrut[t]\\
      & LU w/o MS+MP & 15.44 & 37.70 & 44.84 & 17.42 & 28.85 & -12.44\% \bigstrut[b]\\
    \hline
    \multirow{3}[2]{*}{\shortstack{Structure\\Ablation}} & LU w/o SA & 12.25 & 19.61 & 26.85 & 6.12 & 16.21 & -50.81\% \bigstrut[t]\\
      & LU w/o OQ & 10.56 & 32.50 & 58.25 & 14.03 & 28.83 & -12.49\% \\
      & LU w/o QKN & 12.69 & 36.89 & 50.35 & 12.15 & 28.02 & -14.96\% \bigstrut[b]\\
    \hline
    \hline
  \end{tabular}}
  \label{tab:lu_downward_ablation}%
\end{table*}

\textbf{Table~\ref{tab:lu_downward_ablation}} shows that, except for removing MS alone, the LU ablations produce the expected declines.
Removing SA causes by far the largest degradation, identifying adaptive aggregation as the most consequential structural component under LU as well.
Removing OQ or QKN also hurts Overall, matching the direction of the GDU-based study and showing that their contributions are not specific to GDU.
Removing MS alone slightly improves Overall, in contrast to its decrease in the main GDU-based ablation.
This sign reversal suggests that the effect of MS depends on the update rule.
MS trains the model to maintain and revise multiple entities within a shared memory state.
One possible explanation is that GDU's delta-rule update performs a key-conditioned correction that approximately overwrites the target entity's existing value association.
This mechanism may make multiple entities easier to manage, whereas LU globally decays the memory state and adds new key-value content, potentially increasing cross-entity interference.
Finally, removing both MS and MP produces a clear overall degradation, showing that the auxiliary data remains important as a whole.

%% file: appendix/backbone_transferability.tex
\section{Transfer Across Backbone Families and Scales}
\label{app:backbone_transferability}

Metis can be applied to various compatible causal decoder-only Transformer backbones by integrating its native-memory components into their Transformer layers.
Since the main experiments use Qwen3.5 backbones, we test this architectural flexibility by applying Metis to Llama3.1-8B~\cite{grattafiori2024llama}, Gemma4-12B~\cite{team2026gemma}, Gemma4-31B~\cite{team2026gemma}, and Llama3.1-70B~\cite{grattafiori2024llama}.

\textbf{Setup.}
We replace the Qwen3.5 backbone with each transferred backbone and otherwise follow the same training and evaluation setup as the main experiments, reporting the checkpoint at step 14,000. For each backbone, we compare three conditions. No Context provides no interaction history, Metis uses only its internal memory state without replaying the original context, and Full Context directly provides the complete interaction history.
We also retain the Qwen3.5-based Metis-4B and Metis-9B results from the main evaluation as references.

\begin{table*}[t]
  \centering
  \small
  \renewcommand{\arraystretch}{1.08}
  \caption{Backbone-transfer evaluation. Overall is the equal-weight macro-average across the four benchmarks.}
  \vspace{-0.2cm}
  \resizebox{0.95\textwidth}{!}
  {\begin{tabular}{lccccc}
    \hline
    \hline
    \textbf{Method} & \textbf{LoCoMo (Gold)} & \textbf{NextMem} & \textbf{Metis Test Set} & \textbf{MemOps (Gold)} & \textbf{Overall} \\
    \hline
    Llama3.1-8B (No Context)   & 0.25  & 19.27 & 19.98 & 3.25  & 10.69 \bigstrut[t]\\
    Llama3.1-8B (Full Context) & 62.80 & 75.75 & 73.42 & 70.48 & 70.61 \\
    Gemma4-12B (No Context)      & 0.07  & 15.69 & 16.40 & 1.27  & 8.36 \\
    Gemma4-12B (Full Context)    & 66.39 & 81.05 & 78.05 & 83.71 & 77.30 \\
    Gemma4-31B (No Context)      & 0.00  & 10.55 & 10.39 & 0.66  & 5.40 \\
    Gemma4-31B (Full Context)    & 70.38 & 80.79 & 77.96 & 83.57 & 78.18 \\
    Llama3.1-70B (No Context)      & 0.13  & 26.44 & 21.24 & 3.01  & 12.71 \\
    Llama3.1-70B (Full Context)    & 64.93 & 77.22 & 74.56 & 83.52 & 75.06 \bigstrut[b]\\
    \hline
    \textbf{Metis (Llama3.1-8B)} & 20.56 & 39.98 & 72.33 & 11.25 & 36.03 \bigstrut[t]\\
    \textbf{Metis (Gemma4-12B)} & 21.68 & 49.00 & 58.77 & 21.28 & 37.68 \\
    \textbf{Metis (Gemma4-31B)} & 17.80 & 44.98 & 53.37 & 17.61 & 33.44 \\
    \textbf{Metis (Llama3.1-70B)} & 22.17 & 52.36 & 70.95 & 11.68 & 39.29 \bigstrut[b]\\
    \hline
    Metis-4B (Qwen3.5-4B) & 16.31 & 41.69 & 56.72 & 17.84 & 33.14 \bigstrut[t]\\
    Metis-9B (Qwen3.5-9B) & 16.81 & 43.39 & 57.92 & 19.63 & 34.44 \bigstrut[b]\\
    \hline
    \hline
  \end{tabular}}
  \label{tab:backbone_transferability}%
\end{table*}

\textbf{Results.}
As shown in \textbf{Table~\ref{tab:backbone_transferability}}, all four transferred Metis variants exhibit the same broad pattern. On each benchmark, memory-only performance is higher than the corresponding no-context control but lower than the full-context control.
Relative to the Qwen-based Metis-4B and Metis-9B references, some transferred variants improve the overall performance, but these gains do not hold across every task.

These results show that Metis's memory-specific mid-training transfers to the tested compatible Gemma and Llama backbones and model scales.
However, we find that the performance varies across model sizes and benchmarks without a monotonic scaling trend.
One possible reason is that the shared training recipe may not be equally well matched to every backbone and scale.
Thus, the observed transferability does not imply backbone-independent behavior, universal superiority, or complete preservation of full-context capability.

%% file: appendix/appendix_exp_impl_details.tex
\section{Evaluation Implementation Details}
\label{app:evaluation_details}

\textbf{DenseRAG.}
We construct the retrieval corpus from only the context visible to the current test instance. The context is divided at sentence boundaries, and sentences longer than 256 embedding-model tokens are further split into contiguous chunks.  Retrieval is performed by cosine similarity, and the top-$5$ chunks are provided to the corresponding
Qwen3.5-4B, 9B, or 27B generator. Gold answers, evidence identifiers, and future turns are never included in the retrieval corpus.

\textbf{Temp-LoRA.}
Following the official repository's raw-text adaptation design, we implement an official-like memory-task adaptation of Temp-LoRA for Qwen3.5-4B, 9B, and 27B.
For each test instance, the LoRA and optimizer are reset, and each memory step is capped at 4,096 tokens (enough for reported benchmarks) and split into 1,024-token chunks. We perform two updates per chunk in BF16 with batch size 1 and AdamW (learning rate $5\times10^{-5}$, zero weight decay, and no scheduler).
The LoRA uses rank 64, scaling factor 64, and dropout 0.05, and is applied to
the attention and feed-forward projections. Its parameters and optimizer state
persist within an instance and are discarded before the next instance.

\textbf{$\delta$-Mem.}
We use the officially released \texttt{delta-mem\_qwen3\_4b-instruct}
adapter with the Qwen3-4B-Instruct-2507 backbone and the official
$\delta$-Mem runtime. Each raw memory step is ingested as a separate user
message. Before issuing the question-only query, we retain only the online
$\delta$-Mem state and clear the chat history, processed input IDs, and KV
cache. We reset $\delta$-Mem before each instance.

\textbf{Metis.}
We reset Metis to an empty \texttt{LocalMemory} state before each instance.
During the query phase, Metis receives only the question prompt.

%% file: appendix/experiment_prompts.tex
\label{app:experiment_prompts}

\newsavebox{\metispromptbox}
\newenvironment{metisprompt}
  {%
    \par\smallskip\noindent
    \setlength{\fboxsep}{5pt}%
    \begin{lrbox}{\metispromptbox}%
    \begin{minipage}{\dimexpr\linewidth-2\fboxsep-2\fboxrule\relax}%
    \small\ttfamily\raggedright
    \setlength{\parindent}{0pt}%
    \setlength{\parskip}{0pt}%
  }
  {%
    \end{minipage}%
    \end{lrbox}%
    \fcolorbox{black!35}{black!3}{\usebox{\metispromptbox}}%
    \par\smallskip
  }
\newcommand{\metisprompttitle}[1]{\par\medskip\noindent\textbf{#1}\par\nopagebreak}
\newcommand{\metispromptblank}{\par\vspace{\baselineskip}}
\newcommand{\metispromptindent}{\hspace*{2\fontdimen2\font}}

\subsection{Prompt Notation and Coverage}

Double braces, such as \texttt{\{\{question\}\}}, denote values inserted at runtime.
Only prompt text is shown verbatim.
Per-instance questions, contexts, retrieved chunks, and answers are omitted.
Automatic wrapping inside the prompt boxes is typographical only.
Ablation and LowRank runs add no natural-language prompt.

\subsection{Information and Query Prompts of Baselines}
\label{appendix:prompt_baseline}

\subsubsection{Qwen3.5 Backbone}

\metisprompttitle{Query of Memory-based QA Tasks (No-context).}

\begin{metisprompt}
Question: \{\{question\}\}\par
Answer with a short phrase. If the answer is not known from the given information, say "No information available".
\end{metisprompt}

\metisprompttitle{Query of Memory Operation Tasks (No-context).}

\begin{metisprompt}
Question: \{\{question\}\}\par
Answer the question using the memory context. Be concise, but include all necessary details. If the answer is not known from the given information, say "No information available".
\end{metisprompt}

\metisprompttitle{Query of Memory-based QA Tasks (Partial-context).}
Optional metadata fields are emitted only when available.

\begin{metisprompt}
Retrieved context:\par
[chunk 1 | date=\{\{optional\_date\_time\}\} | session=\{\{optional\_session\_id\}\} | speaker=\{\{optional\_speaker\}\}]\par
\{\{retrieved\_chunk\_text\}\}\metispromptblank
Question: \{\{question\}\}\par
Answer with a short phrase using only the retrieved context. If the answer is not known from the retrieved context, say "No information available".\par
Short answer:
\end{metisprompt}

\metisprompttitle{Query of Memory Operation Tasks (Partial-context).}

\begin{metisprompt}
Retrieved context:\par
[chunk 1]\par
\{\{retrieved\_chunk\_text\}\}\metispromptblank
Question: \{\{question\}\}\par
Answer with a short phrase using only the retrieved context. If the answer is not known from the retrieved context, say "No information available".\par
Short answer:
\end{metisprompt}

\metisprompttitle{Query of LoCoMo (Gold) in Memory-based QA Tasks (Full-context).}

\begin{metisprompt}
Evidence-session context:\par
SESSION: \{\{session\_id\}\}\par
DATE: \{\{date\_time\}\}\par
\{\{dialogue\_turn\_id\}\} \{\{speaker\}\} said: "\{\{dialogue\_text\}\}" Shared image caption: \{\{optional\_image\_caption\}\}\metispromptblank
Question: \{\{question\}\}\par
Answer with a short phrase. If the answer is not known from the given information, say "No information available".
\end{metisprompt}

\metisprompttitle{Query of NextMem in Memory-based QA Tasks (Full-context).}

\begin{metisprompt}
Reference context:\par
\{\{reference\_context\}\}\metispromptblank
Question: \{\{question\}\}\par
Answer with a short phrase. If the answer is not known from the given information, say "No information available".
\end{metisprompt}

\metisprompttitle{Query of Memory Operation Tasks (Full-context).}
The context block repeats for each normalized context item.

\begin{metisprompt}
Memory context:\metispromptblank
\{\{context\_id\}\}:\par
\{\{memory\_context\}\}\metispromptblank
Question: \{\{question\}\}\par
Answer the question using the memory context. Be concise, but include all necessary details. If the answer is not known from the given information, say "No information available".
\end{metisprompt}

\subsubsection{Metis, Temp-LoRA, and $\delta$-Mem}

During the information stage, we retain each method's method-specific write interface while keeping the ordered memory-step contents fixed. 
Metis prepends a fixed, answer-independent commit instruction to each memory step before its memory-commit operation. Temp-LoRA updates its temporary LoRA directly on the memory context. $\delta$-Mem passes each raw memory step as a user
message through the officially released runtime's native chat-message ingestion
path to update its online state. At query time, all three methods use the same prompt.

\metisprompttitle{Share Query of Metis, Temp-LoRA, and $\delta$-Mem.}

\begin{metisprompt}
Answer from the learned memory state produced during the information phase.\par
Give the shortest factual answer you can. Do not explain.\par
Question: \{\{question\}\}\par
Short answer:
\end{metisprompt}

\metisprompttitle{Information-stage Prompt of Metis in Memory-based QA Tasks. }

\begin{metisprompt}
Conversation memory segment.\par
Commit the following dated dialogue segment to memory for later question answering.\metispromptblank
\{\{memory\_step.content\}\}
\end{metisprompt}

\metisprompttitle{Information-stage Prompt of Metis in Memory Operation Tasks. }

\begin{metisprompt}
Conversation memory segment.\par
Commit the following dialogue segment to memory for later question answering.\metispromptblank
\{\{memory\_step.content\}\}
\end{metisprompt}

\metisprompttitle{Information-stage Payload of Temp-LoRA and $\delta$-Mem.}
These methods pass \texttt{memory\_steps[*].content} directly during the information step.

\begin{metisprompt}
\{\{memory\_step.content\}\}
\end{metisprompt}

\subsection{Prompts for LLM-as-a-Judge }
\label{appendix:prompt_LLM_as_a_judge}

Formal scoring uses \texttt{gpt-4.1-mini} with temperature 0, three repeats. For each example, the score is the median of the three LLM-as-a-judge scores.

\metisprompttitle{System Message.}

\begin{metisprompt}
You are a conservative but fair evaluator for a memory question-answering benchmark. Your job is to avoid overly generous partial credit while still accepting truly equivalent answers, aliases, abbreviations, and harmless formatting differences. Return JSON only.
\end{metisprompt}

\metisprompttitle{User Instruction.}

\begin{metisprompt}
Grade model\_output against gold\_answer for the question. Return JSON with keys: score (0 to 1), pass (boolean), matched\_points (array of strings), missed\_points (array of strings), and rationale (short string). Use this strict rubric: give 1.0 only when the answer contains the correct core entity/value/date/relationship asked for, allowing aliases and semantically equivalent wording. Give 0.5 to 0.75 only when the output includes the correct core answer but has minor extra wording, minor imprecision, or one secondary omission. Give 0 for a different person, organization, place, number, date, title, relation, or answer choice; for a broad category when the gold answer is a specific entity; for answers that merely share common words with gold; for plausible guesses unsupported by the exact answer; or when the model says the answer is unknown/unavailable while gold is answerable. Do not reward explanation quality if the final answer is wrong. If the question asks for a country/state/type and the model gives exactly that correct country/state/type, it is correct even if it could be guessed from world knowledge.
\end{metisprompt}

\metisprompttitle{User Payload (JSON Schema).}

\begin{metisprompt}
\{\par
\metispromptindent"instruction": "\{\{judge\_instruction\_above\}\}",\par
\metispromptindent"question": "\{\{question\}\}",\par
\metispromptindent"gold\_answer": "\{\{gold\_answer\}\}",\par
\metispromptindent"model\_output": "\{\{model\_output\}\}",\par
\metispromptindent"raw\_category": "\{\{raw\_category\}\}",\par
\metispromptindent"is\_adversarial": "\{\{is\_adversarial\}\}",\par
\metispromptindent"baseline": "\{\{baseline\}\}"\par
\}
\end{metisprompt}

\subsection{Prompts in General Capability Study}
\label{appendix:prompt_general_capability}

The active stage setting uses the same message for both models.
For Qwen3.5-4B, it is prepended to the complete benchmark prompt.
For Metis-4B, the model is reset for each example, the message is stored in the memory state, and the unchanged benchmark prompt is then provided.

\begin{metisprompt}
This session is for general-purpose evaluation. For any upcoming question, follow the instruction closely, reason carefully when needed, and answer based on the information available.
\end{metisprompt}

MMLU-Pro, IFEval, GSM8K, and MMMLU retain their native per-instance benchmark prompts.

%% file: appendix/Efficiency.tex
This appendix evaluates the inference efficiency of Metis at the 4B scale, characterizing both the benefits of native memory and the costs it introduces. We first measure application-level end-to-end and query latency on LoCoMo (Gold).
Furthermore, we isolate the effect of history through a controlled sweep over 512 to 128K context tokens, with latency decomposed into fine-grained stage-level components.
Beyond latency, we further examine the storage efficiency of maintaining a fixed-size memory state, measuring the storage that each method must persist per session as history grows.
Finally, we prototype multi-user serving with a LoRA-based serving architecture, measuring its isolation and efficiency at serving scale.

\subsection{End-to-End Latency on LoCoMo (Gold)}
\label{app:efficiency_locomo}

\textbf{Evaluation Protocol.} We evaluate all methods on the 1,527 examples in the LoCoMo (Gold) evidence-session setting, following the baseline configurations described in \textbf{Section~\ref{sec:experiments}}.
Each example is evaluated independently under a cold-start protocol: the method-specific state is reset, the evidence session is processed, and the question is then answered without reusing state across examples.
All methods use greedy decoding with at most 96 generated tokens.
Each run includes one unmeasured warm-up example, and CUDA synchronization is applied at every timing boundary.
End-to-end latency includes retrieval, test-time adaptation, and memory writing when required, followed by query generation.
Query latency starts after these preparation operations have completed. Model loading and answer evaluation are excluded.
Because generation uses natural stopping behavior, the results measure application-level latency rather than controlled decode-only latency.

\begin{table}[t]
    \centering
    \small
    \setlength{\tabcolsep}{3.8pt}
    \renewcommand{\arraystretch}{1.18}
    \caption{End-to-end latency, query latency, input length, generation
    length, and effective throughput on the LoCoMo (Gold) evidence-session
    setting. The smallest and second-smallest latency values are
    \textbf{bolded} and \underline{underlined}, respectively.}
    \vspace{-0.2cm}
    \label{tab:locomo_generation_latency}
    \resizebox{\textwidth}{!}{%
    \begin{tabular}{lccccccccc}
        \hline
        \hline
        \multirow{2}{*}{\textbf{Method}}
        & \multicolumn{2}{c}{\makecell[c]{\textbf{E2E}\\\textbf{Latency (s)}}}
        & \multicolumn{2}{c}{\makecell[c]{\textbf{Query}\\\textbf{Latency (s)}}}
        & \multirow{2}{*}{\makecell[c]{\textbf{Prompt Tokens}\\\textbf{(Avg./P95)}}}
        & \multirow{2}{*}{\makecell[c]{\textbf{Committed Tokens}\\\textbf{(Avg./P95)}}}
        & \multicolumn{2}{c}{\makecell[c]{\textbf{Generated}\\\textbf{Tokens}}}
        & \multirow{2}{*}{\makecell[c]{\textbf{Effective}\\\textbf{Throughput}\\\textbf{(tokens/s)}}} \\
        \cmidrule(lr){2-3}\cmidrule(lr){4-5}\cmidrule(lr){8-9}
        & \textbf{Avg.} & \textbf{P95}
        & \textbf{Avg.} & \textbf{P95}
        & &
        & \textbf{Avg.} & \textbf{P95}
        & \\
        \hline
        No Context      & \textbf{0.149} & \textbf{0.145} & \textbf{0.149} & \textbf{0.145} & 50.2 / 58.0     & --              & 3.0 & 3.0  & \underline{20.16} \\
        Full Context    & 0.607 & 3.012 & 0.607 & 3.012 & 1410.6 / 3345.8 & --              & 5.7 & 16.0 & 9.37  \\
        Partial Context & \underline{0.268} & \underline{0.456} & \underline{0.223} & \underline{0.412} & 290.5 / 322.0   & --              & 4.0 & 11.0 & 17.73 \\
        Temp-LoRA       & 1.567 & 3.292 & 0.305 & 0.746 & 56.2 / 64.0     & 1386.0 / 3365.2 & 5.2 & 15.0 & 17.11 \\
        $\delta$-Mem    & 0.884 & 1.600 & 0.656 & 1.365 & 49.2 / 57.0     & 1377.2 / 3353.1 & 8.8 & 21.0 & 13.44 \\
        Metis           & 0.562 & 0.926 & 0.360 & 0.609 & 56.2 / 64.0     & 1449.0 / 3515.4 & 8.3 & 15.0 & \textbf{23.15} \\
        \hline
        \hline
    \end{tabular}%
    }
\end{table}

\textbf{Results.}
Partial Context has the lowest end-to-end latency among methods that use historical information, with an average of 0.268 seconds and a P95 of 0.456 seconds. Metis achieves an average latency of 0.562 seconds, slightly below the 0.607 seconds of Full Context, while reducing P95 latency from 3.012 to 0.926 seconds, a reduction of 69.3\%. Compared with $\delta$-Mem, Metis reduces average and P95 end-to-end latency by 36.4\% and 42.2\%, respectively; compared with Temp-LoRA, the corresponding reductions are 64.1\% and 71.9\%. Metis also reduces the average query latency from 0.656 to 0.360 seconds relative to $\delta$-Mem, with P95 decreasing from 1.365 to 0.609 seconds.

\textbf{Analysis.}
The slightly lower average latency and substantially lower P95 latency relative to Full Context indicate that Metis provides a more stable query path across LoCoMo (Gold) instances with varying history lengths. Full Context processes 1,410.6 prompt tokens on average and 3,345.8 tokens at P95 for every query. Metis commits a similar amount of historical information, at 1,449.0 tokens on average and 3,515.4 tokens at P95, but its subsequent query contains only 56.2 prompt tokens on average and 64 tokens at P95. Thus, Metis shifts historical-information processing into a separate memory-write stage and avoids replaying the original evidence during querying, contributing to its lower query and tail latency.
Its advantage over $\delta$-Mem comes mainly from memory utilization rather than memory construction: its average write latency is moderately lower at 0.199 versus 0.227 seconds, whereas it reduces average and P95 query latency by 45.2\% and 55.4\%, respectively. Temp-LoRA, in contrast, is dominated by its 1.186-second test-time adaptation. These comparisons show that Metis has lower online overhead than the two state-based parametric memory baselines. Partial Context remains the fastest cold-start method because its retrieval stage costs only 0.044 seconds on average and produces a compact 290.5-token query prompt. The token statistics of $\delta$-Mem follow its Qwen3 tokenizer, while the other methods use the Qwen3.5 tokenizer.

\subsection{Latency Scaling with Context Length}
\label{app:efficiency_context_scaling}

\textbf{Evaluation Protocol.}
We evaluate all six methods at the 4B scale on a single NVIDIA A800 GPU using controlled histories from 512 to 128K tokens. Each final query contains 512 content tokens, excluding method-specific templates, and generates either 32 or 128 tokens with batch size 1. For every method, context length, and output length, we report the mean and standard deviation over five measured runs after one warm-up run to remove kernel initialization. Metis processes the history in chunks of at most 2K tokens, so its number of commits grows naturally with context length. Partial Context rebuilds its index over the complete history for every run and retrieves the top five sentence chunks. This is a generation-only experiment without answer scoring. End-to-end latency includes all method-specific history preparation or writing and the subsequent query, whereas query latency is measured after history preparation has completed.

\begin{figure}[t]
    \centering
    \includegraphics[width=\textwidth]{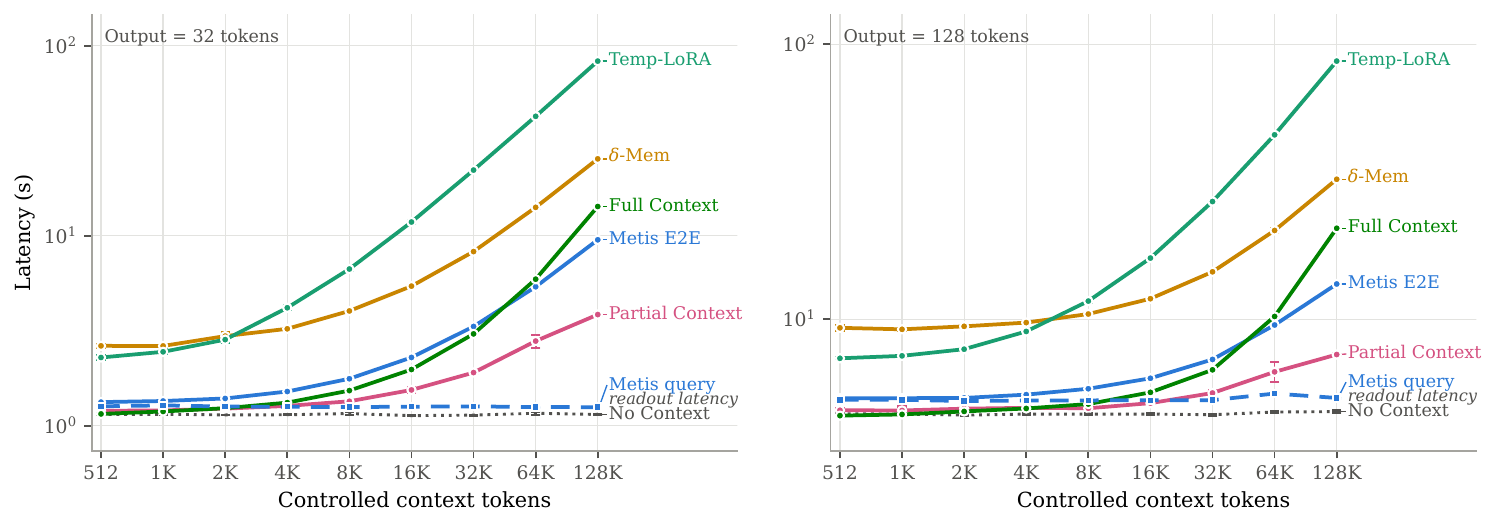}
    \caption{End-to-end latency as controlled context length increases, with 32 generated tokens (left) and 128 generated tokens (right). Metis E2E includes streaming history ingestion and query generation; Metis query excludes ingestion. Readout latency is computed as Metis query latency minus No Context latency. The first measured crossover between Metis E2E and Full Context occurs at 64K for both output lengths.}
    \label{fig:e2e_latency_curves}
    \vspace{-0.3cm}
\end{figure}
\textbf{Results.}
At context lengths up to 32K, Metis has slightly higher end-to-end latency than Full Context. The ordering reverses at the first measured 64K point for both output lengths: for 32 generated tokens, Full Context and Metis take 5.909 and 5.390 seconds, respectively, giving a $1.096\times$ speedup; for 128 generated tokens, they take 10.228 and 9.522 seconds, giving a $1.074\times$ speedup. At 128K, these E2E speedups increase to $1.497\times$ (14.254 versus 9.521 seconds) and $1.595\times$ (21.404 versus 13.423 seconds), respectively, in \textbf{Figure~\ref{fig:e2e_latency_curves}}.

\begin{figure}[t]
    \centering
    \includegraphics[width=\textwidth]{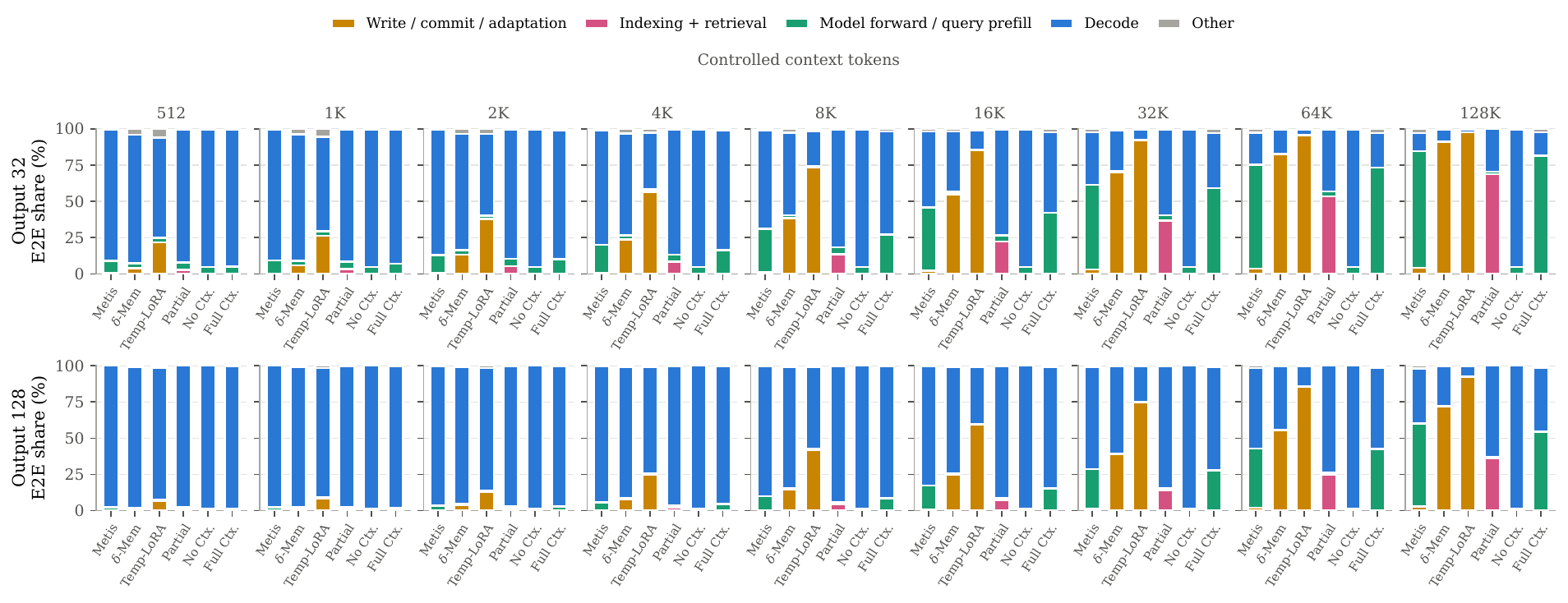}
    \caption{Normalized end-to-end latency breakdown across context lengths for 32-token (top) and 128-token (bottom) outputs. For Metis, model forward/query prefill includes the history-encoding forwards and final query prefill, while write/commit/adaptation contains only the pure memory update. For $\delta$-Mem and Temp-LoRA, the latter category represents state writing and temporary adaptation, respectively; indexing and retrieval is reported separately for Partial Context. Other includes reset, tokenization, prompt construction, synchronization, text decoding, and unseparated host-side work.}
    \label{fig:latency_component_shares}
    \vspace{-0.3cm}
\end{figure}

\textbf{Figure~\ref{fig:latency_component_shares}} further shows that pure memory commit is a small fraction of Metis latency. At 64K and 128K, commit takes approximately 0.203 and 0.406 seconds, respectively, and accounts for only 2.1--4.3\% of Metis E2E latency across the two output lengths; most ingestion time instead comes from the history-encoding backbone forwards, which grow from approximately 3.86 to 7.65 seconds. Metis also has lower E2E latency than both state-based baselines at every measured context and output length. At 128K, it is $2.40$--$2.67\times$ faster than $\delta$-Mem and $6.46$--$8.71\times$ faster than Temp-LoRA. Consistent with this gap, state writing contributes 71.7--90.8\% of $\delta$-Mem E2E latency at 128K, while temporary adaptation contributes 92.2--97.7\% for Temp-LoRA.

\textbf{Analysis.}
The E2E results reveal two complementary efficiency advantages. Relative to Full Context, Metis becomes increasingly advantageous as history grows in the long-context regime. The crossover is first observed only at 64K because Qwen3.5-4B already uses linear attention in 24 of its 32 layers, substantially reducing the context-dependent KV traffic of Full Context and thereby delaying the E2E crossover in favor of Metis. As the KV caches of the remaining 8 full-attention layers continue to grow, the advantage of Metis becomes more pronounced: when context doubles from 64K to 128K, Full Context prefill increases by approximately $2.67\times$ (4.34 to 11.60 seconds) and decode by approximately $1.65\times$ (1.40 to 2.31 seconds for 32-token outputs and 5.74 to 9.45 seconds for 128-token outputs), widening the E2E latency gap. Relative to the two state-based baselines, Metis maintains lower E2E latency throughout the entire context-length sweep because it avoids their increasingly expensive state writing or temporary adaptation. We use query latency only to isolate the cost of memory readout, estimated as Metis query latency minus No Context latency. For 32-token outputs, the estimated readout is 0.091 seconds at 64K and 0.106 seconds at 128K; for 128-token outputs, it is 0.770 and 0.559 seconds, respectively. These values account for only 7.2--14.4\% of Metis query latency and do not increase when the history doubles. \textbf{Figure~\ref{fig:latency_component_shares}} similarly shows that pure memory commit contributes only 2.1--4.3\% of Metis E2E latency at 64K and 128K, with most ingestion time spent in the history-encoding backbone forwards. Thus, neither memory readout nor pure memory commit is a major latency bottleneck. In principle, both stages can be executed concurrently with backbone attention on separate CUDA streams. Such asynchronous overlap, however, requires finer-grained kernel launches and inter-stream synchronization, introducing launch and scheduling overhead that can offset the latency hidden at this scale. We therefore do not enable this optimization in the final implementation.

\subsection{Per-Session Storage across Context Lengths}
\label{app:efficiency_storage}

\textbf{Evaluation Protocol.}
We also measure the persistent per-session storage of the 4B methods over different context lengths from 512 to 32K token. All values are obtained from runtime objects, rather than estimated analytically from parameter counts. For Full Context, we retain the KV cache required to resume a session without replaying its history; the measurement includes the controlled context, a fixed 512-token query prompt, and 32 generated tokens. KV cache and model states are stored in BF16. The RAG store contains FP32 BGE-M3 embeddings and the corresponding chunk text. For state-based methods, we count only state that must be persisted separately for each session.

\begin{figure}[t]
    \centering
    \includegraphics[width=0.75\linewidth]{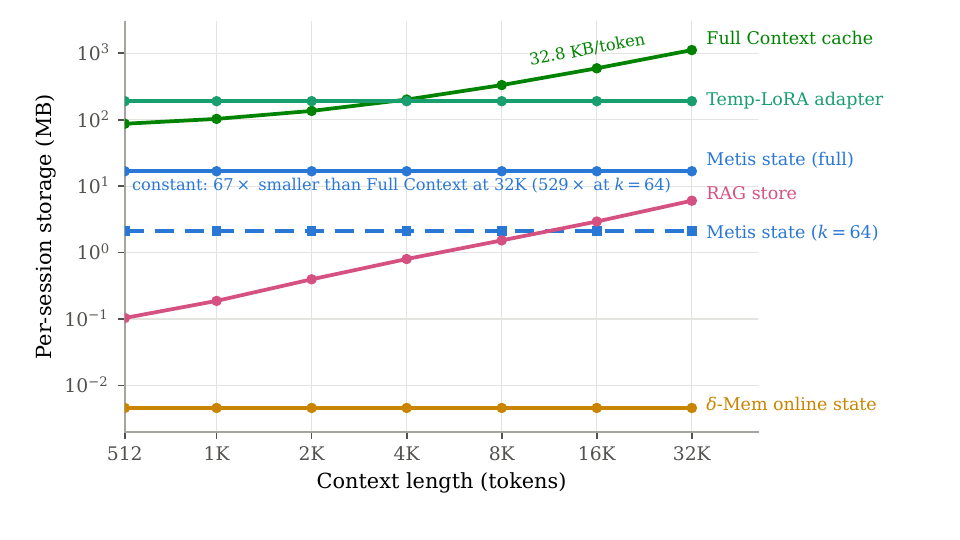}
    \caption{Persistent storage per session as a function of context length.
    }
    \label{fig:storage_vs_context}
    \vspace{-0.3cm}
\end{figure}

\textbf{Results.}
Full Context storage grows from 87.06 MB at 512 context tokens to 1,118.86 MB at 32K. Its 8 full-attention layers add 32.8 KB of KV cache per token, while the 24 linear-attention layers contribute a constant 51.90 MB state. In contrast, Metis Full remains at 16.79 MB and Metis $k=64$ remains at 2.11 MB.
At 32K, Full Context therefore occupies $67\times$ the Metis Full state and $529\times$ the Metis $k=64$ state; extrapolating the measured linear trend to 128K gives approximately $256\times$ and $2{,}000\times$, respectively. The other constant-size states are 4.6 KB for $\delta$-Mem and 190.3 MB for Temp-LoRA, whereas RAG Store grows from 0.104 MB at 512 tokens to 6.051 MB at 32K.

\textbf{Analysis.}
Metis eliminates context-dependent storage growth: the Full Context KV cache reaches 1,066.96 MB at 32K, whereas both Metis states remain constant. Low-rank persistence provides a particularly efficient operating point: Metis $(k=64)$ uses one eighth of the Metis Full state while retaining 99.9\% of full-state performance on average. $\delta$-Mem occupies only 4.6 KB because its official rank-8 configuration stores one $8\times8$ BF16 online matrix in each of 36 attention modules, compressing each session into a low-dimensional state; this limited capacity is consistent with its lower LoCoMo score (Table~\ref{tab:results_memory_based_QA}). Despite the small state, every query token reads it in all 36 attention modules and applies additional $q/k/v/o$ delta projections, introducing per-layer matrix operations and kernel launches on top of the backbone attention. RAG Store also grows with history and depends on a separate retrieval and indexing pipeline; Metis $k=64$ becomes smaller at approximately 11K tokens and remains smaller at the measured 16K and 32K points. Temp-LoRA requires a 190.3 MB per-session adapter, $11.3\times$ the Metis Full state. Overall, Metis provides high storage efficiency for long texts: both Metis Full and $k=64$ remain constant and substantially smaller than Full Context, while $k=64$ also becomes smaller than RAG Store as history grows. As a storage-only upper bound on an 80 GB GPU with approximately 72 GB remaining after 4B model weights, 32K Full Context caches accommodate roughly 64 resident sessions, compared with about 4,000 full-state Metis sessions; the former decreases with history length, while the latter is independent of it.

\subsection{Multi-User Serving}
\label{app:efficiency_serving}

A critical concern for Metis is that maintaining a personalized memory for every user means deploying one model per user, which would be prohibitive at serving scale.
In this section, we introduce a new serving solution for Metis to address this issue, which is inspired by LoRA-based approaches.

\textbf{Serving Architecture.}
All users can share a single copy of the frozen backbone and the static Metis parameters.
Therefore, the only per-user component is the dynamic memory state, which persists at 16.79~MB per user in full precision for Metis-4B.
Based on this property, we refer to the design of multi-tenant LoRA serving systems such as Punica~\cite{chen2024punica} and S-LoRA~\cite{sheng2023s}.
Specifically, we treat each user's dynamic memory state as a user-specific adapter, and design a \textbf{Memory-as-Adapter (MaA)} serving architecture for Metis that batches requests from different users into a single parallel forward pass.
This architecture consists of three components.
First of all, we consider states as adapters.
Each user's memory state is persisted on SSD as swappable per-user data, gathered by user index when a session starts and written back upon eviction.
The second component is stateless model replicas.
A serving instance is bound to no user, so any replica can serve any request, making horizontal scaling and load balancing identical to ordinary LLM serving.
Finally, we use batched state injection.
When requests from different users are batched together, their states are stacked along the batch dimension and injected into the model in one operation.
The key design is parallelism within the adapter layer itself.
Generic multi-adapter inference has an inherent serialization bottleneck at the adapter computation.
Since adapter shapes vary across tenants (\textit{e.g.,} LoRA ranks), naive implementations must partition the batch by adapter and compute group by group, while efficient ones rely on customized gather-style kernels such as Punica.
Metis eliminates this bottleneck by construction.
All users' memory states share an identical shape, so the memory readout reduces to a single standard batched matrix multiplication.
During this process, each request reads its own state within one kernel call, without grouping, sorting, or custom operators. Under MaA, from state injection to the end of decoding, the forward pass contains no per-user serial segment.
The incremental cost of multi-user serving over single-user inference is merely a batch dimension.

\textbf{Evaluation Protocol.}
In this subsection, we evaluate the performance of Metis during serving from three dimensions: isolation, layency scaling under batching, and the overhead of memory path.
Each experiment includes one untimed warm-up, and CUDA synchronization is performed at all timing boundaries.
All results are medians over five runs.
The workload is a synthetic multi-user session.
Each user holds an independent history containing user-specific facts, and each user's state is pre-built via standard write paths.
The measurements of Metis account for the full overhead of loading and injecting state at request.
Each baseline uses the most favorable accounting method.
RAG pre-builds the vector database index, and latency only calculates retrieval and inference latency.
Full Context is considered to have KV Cache and no Cache, while the KV-cache variant assumes that the Context has a pre-built KV Cache residing in HBM.

\textbf{Isolation.}
Because per-request states occupy disjoint slices of the batch dimension, cross-user interference is structurally excluded, and we verify this empirically with 20 concurrently served users holding globally unique persona facts.
In evaluation, across 200 sampled cross-user probes, in which user $i$ is asked about user $j$'s fact, no response ever contained another user's value.
We also test the recall of own-fact across various batch sizes, where we sweep $B \in \{1, 4, 8, 16, 20\}$ yields identical accuracy under a one-fact-per-user load (100.0\% at every $B$). 

\begin{figure}[t]
    \centering
    \includegraphics[width=\textwidth]{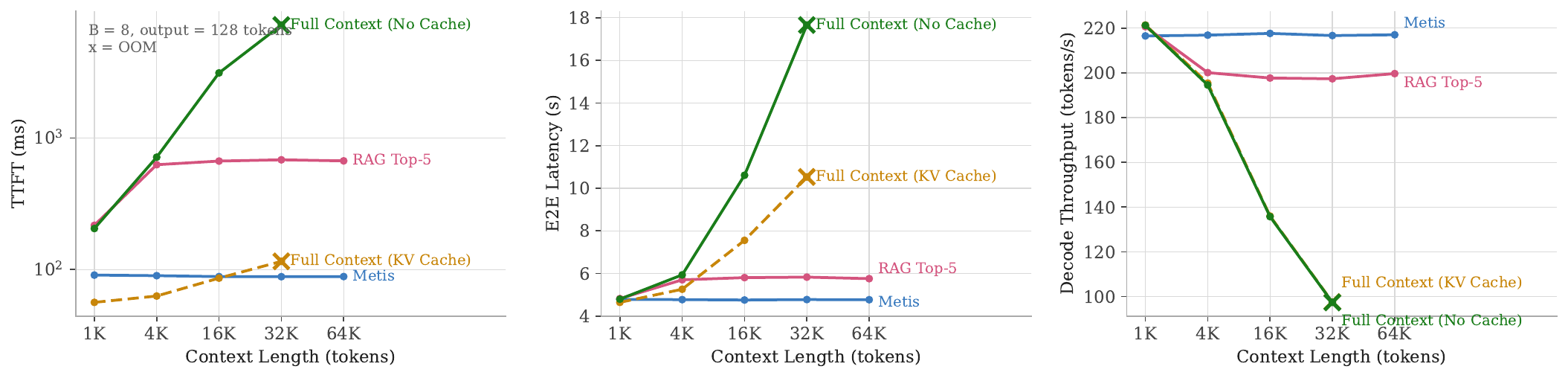}
    \caption{Batched multi-user serving ($B{=}8$, 128 generated tokens per user) as per-user context length grows from 1K to 64K tokens. Left: Time to First Token (TTFT). Middle: end-to-end latency per batched request. Right: decode-phase throughput.}
    \label{fig:serving_context_sweep}
\end{figure}

\textbf{Latency Scaling with Context Length under Batching.}
\textbf{Figure~\ref{fig:serving_context_sweep}} sweeps per-user context length from 1K to 64K tokens under the MaA architecture, with $B{=}8$ and 128 generated tokens per user, comparing Metis against RAG and Full Context with and without a resident KV cache in terms of TTFT, end-to-end latency, and decode throughput.
All three Metis metrics are flat from 1K to 64K. TTFT stays at 88--91~ms (including loading and injecting states from SSD), end-to-end latency at 4.76--4.78~s, and decode throughput at 216--218 tokens/s.

In contrast, the Full Context (KV Cache) variant achieves lower TTFT at short contexts, but its decode throughput halves as context grows (221 to 97 tokens/s from 1K to 32K).
Even with prefill fully amortized by the resident cache, every decoding step must attend over the entire 32K-token cache, whereas the Metis query processes a short prompt and reads a fixed-size state. 
Its end-to-end latency is therefore overtaken by Metis beyond 4K and reaches $2.2\times$ the Metis latency at 32K.
Without a cache, Full Context pays the full prefill on every request, and its TTFT grows from 0.2 to 7.2~s.
At 64K, both Full Context variants run out of memory, as the batched request totals $B \times 64\text{K} \approx 525$K prompt tokens, while Metis and RAG are unaffected. 
RAG plateaus at roughly 0.67~s TTFT and 5.8~s end-to-end, since retrieval bounds its prompt length; it is the closest baseline, but pays a fixed extra second per request and its per-session store grows unboundedly with history.

\begin{figure}[t]
    \centering
    \includegraphics[width=0.98\linewidth]{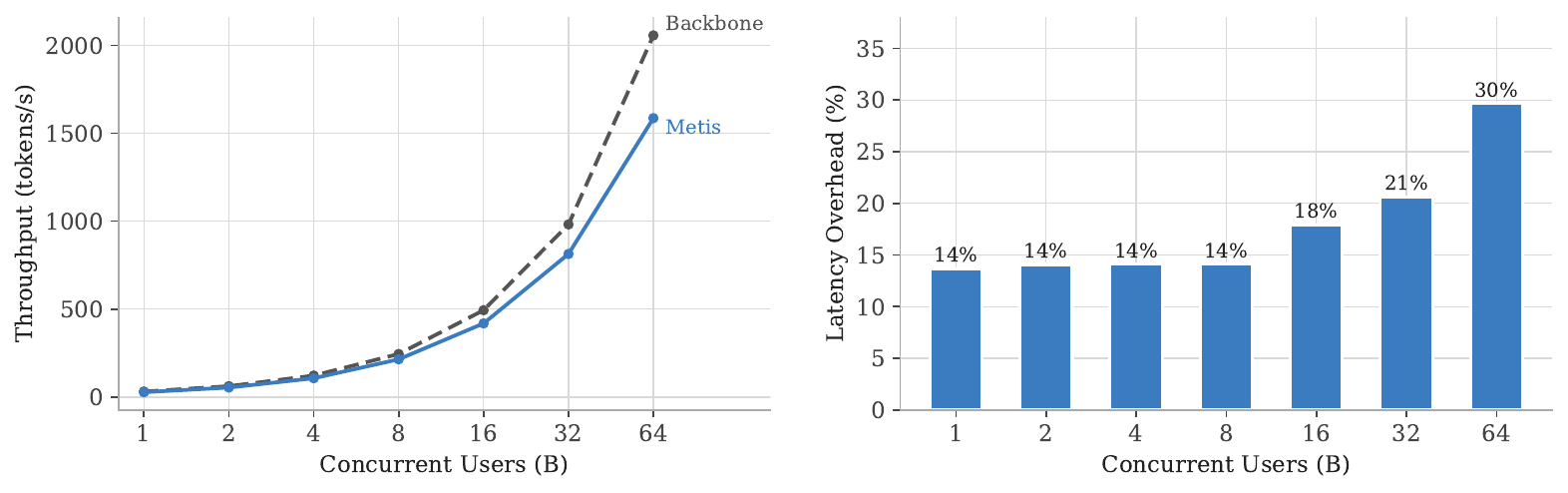}
    \caption{Serving cost of the native-memory path. Left: aggregate decode throughput of batched multi-user Metis. Right: relative latency overhead, which includes re-stacking and re-injecting all per-user states on every request.}
    \label{fig:serving_overhead}
\end{figure}

\textbf{Overhead of the Memory Path.} 
We further compare the latency of Metis-4B against the raw Qwen3.5-4B backbone under batching, to quantify the gap between Metis and the latency lower bound of the serving system.
As shown in \textbf{Figure~\ref{fig:serving_overhead}}, under identical prompts and decoding settings, the full memory path adds only a stable 14\% latency overhead up to $B{=}8$, rising to 30\% at $B{=}64$.
The latter is measured under a deliberately conservative protocol that re-loads, re-stacks, and re-injects all per-user states on every request.
Despite this overhead, scaling remains near-linear.
Aggregate throughput reaches 1{,}585.6 tokens/s at $B{=}64$, a $57.1\times$ speedup over serving the same users serially, with only a 12\% increase in per-user latency.
This finding agrees with the single-request efficiency analysis above.
As shown in \textbf{Section~\ref{app:efficiency_context_scaling}}, at $B{=}1$ memory readout adds only about 0.1~s (7.2--14.4\% of query latency), and pure commit accounts for merely 2.1--4.3\% of end-to-end latency.
In other words, the memory path is not the latency bottleneck, in either the single-request or the batched serving regime, and batching does not alter this property.

\textbf{Analysis.}
In the above part, the MaA architecture provides a foundation for the large-scale deployment that Metis may face in the future.
Under this architecture, the per-user cost of Metis is a fixed-size, swappable state rather than a model replica.
Relative to context-replay baselines, its advantage compounds with context length and with the decode share of the workload.
Compared with the raw backbone, the system overhead it introduces is limited.
Moreover, since per-request states occupy disjoint slices of the batch dimension, the architecture excludes cross-request memory interference by construction.
This simplifies data governance, because deleting a user reduces to deleting one state file.

We note that these results are single-node, static-batch prototype measurements reporting medians rather than tail latency under realistic load.
Several systems challenges remain on the efficiency side.
Integrating MaA with continuous batching requires online allocation and reclamation of state slots as requests dynamically join and leave a batch, 
In addition, memory writes are currently performed as offline commits, production serving will interleave reads (generation) with writes (memory updates), calling for state-consistency and scheduling support at the systems level.
Nevertheless, the fixed size, uniform shape, and read-write separation of memory states provide a solid structural basis for these optimizations, and we believe large-scale deployment of Metis is well within reach.